\documentclass{article} %
\let\standardaddcontentsline\addcontentsline
\usepackage{iclr2027_conference,times}
\let\addcontentsline\standardaddcontentsline
\usepackage{graphicx}
\usepackage{algorithm}
\usepackage{algpseudocode}
\usepackage{longtable}
\usepackage{array}
\usepackage{amssymb}
\usepackage{booktabs}
\usepackage{multirow}
\usepackage{xcolor}
\usepackage{colortbl}
\definecolor{tableblue}{RGB}{232,242,252}
\definecolor{stdorange}{RGB}{90,65,190}
\usepackage{amsmath,amsfonts,bm}

\def\eqref#1{equation~\ref{#1}}

\def\1{\bm{1}}

\DeclareMathAlphabet{\mathsfit}{\encodingdefault}{\sfdefault}{m}{sl}
\SetMathAlphabet{\mathsfit}{bold}{\encodingdefault}{\sfdefault}{bx}{n}

\usepackage{hyperref}
\usepackage{url}
\definecolor{citationblue}{RGB}{44,97,194}
\definecolor{referenceorange}{RGB}{217,116,89}
\definecolor{lossaccent}{RGB}{0,0,255}
\newcommand{\lossdef}[1]{\textcolor{lossaccent}{\boldmath\ensuremath{#1}}}
\hypersetup{
  pdftitle={DepState-JEPA: Learning Depth-Grounded State Representation for Visual Perception and Manipulation},
  pdfauthor={Xiujian Liang, Liuyuan Na, Zehao Du, Jiacheng Xu, Weixi Song, Mingyang Sun, Xingjun Ma, Cewu Lu, Jianhua Sun},
  colorlinks=true,
  linkcolor=referenceorange,
  citecolor=citationblue,
  urlcolor=citationblue,
  linktocpage=true
}
\makeatletter
\AtBeginDocument{%
  \def\Hy@colorlink#1{%
    \begingroup
    \HyColor@UseColor#1%
    \ifx#1\@linkcolor
      \bfseries
    \fi
  }%
}
\makeatother

\title{DepState-JEPA: Learning Depth-Grounded State Representation for Visual Perception and Manipulation}

\author{\textbf{Xiujian Liang}\textsuperscript{\normalfont1,2} \quad \textbf{Liuyuan Na}\textsuperscript{\normalfont2,3} \quad \textbf{Zehao Du}\textsuperscript{\normalfont3} \quad \textbf{Jiacheng Xu}\textsuperscript{\normalfont4} \quad \textbf{Weixi Song}\textsuperscript{\normalfont5,2} \\
\textbf{Mingyang Sun}\textsuperscript{\normalfont5,2} \quad \textbf{Xingjun Ma}\textsuperscript{\normalfont1,2} \quad \textbf{Cewu Lu}\textsuperscript{\normalfont2,3} \quad \textbf{Jianhua Sun}\textsuperscript{\normalfont2,3} \\[2pt]
{\normalfont\small\textsuperscript{\normalfont1}Fudan University \quad \textsuperscript{\normalfont2}Shanghai Innovation Institute (SII)} \\
{\normalfont\small\textsuperscript{\normalfont3}Shanghai Jiao Tong University \quad \textsuperscript{\normalfont4}Shandong University \quad \textsuperscript{\normalfont5}Zhejiang University}
}

\iclrfinalcopy
\begin{document}

\maketitle
\lhead{}
\vspace{-24pt}

\begin{abstract}
Predictive representations for robot manipulation should connect scene geometry with the state changes induced by actions. Our intuition is that understanding what an action will change starts with understanding where things are: complementary RGB and depth observations ground a common scene representation from which to infer those changes. Guided by this intuition, we introduce \textbf{DepState-JEPA}, a dual-stream RGB--Depth joint-embedding predictive architecture that connects spatial understanding with action-conditioned prediction through a shared latent state, \textbf{DepState}.
Built on V-JEPA~2.1, the framework uses depth both as a complementary observation stream and to ground both modalities in metric 3D space, aggregating shared information into DepState while retaining modality-private features for decoding.
Ray-conditioned readouts predict local features without target depth; after an action-conditioned transition, the same decoders predict future RGB and depth features using the predicted state and fixed initial private features.
Robot-state readouts provide recursive proprioceptive inputs; object-trajectory supervision and calibrated forward kinematics shape state learning through motion targets and joint--TCP consistency.
Comparisons show stronger RGB-encoder transfer after RGB-D adaptation, feature completion, conditioned visual forecasting, and manipulation success. Within-model ablations identify contributions from depth content, metric coordinates, and future-state matching, with additional control gains from FK regularization.
The project page is available at \url{https://shawnleung-sii.github.io/depstate-jepa/}.
\end{abstract}
\section{Introduction}
\label{sec:introduction}

Learning visual world models enables anticipation of action outcomes \citep{ha2018worldmodels,maes2026leworldmodel}.
For manipulation in real-world space, this requires relating the current spatial arrangement to
the changes an action may cause. Consider a robot reaching for a cup:
RGB reveals its appearance and context, while depth locates the cup and
gripper relative to the table in three-dimensional space. Anticipating the reach then requires
reasoning about how this arrangement will change, not merely recognizing
the cup. This raises a broader question: \textit{how can the geometry revealed by depth inform a robot’s predictions of how its actions will change the scene?}

\begin{figure*}[t]
    \centering
    \includegraphics[width=\linewidth]{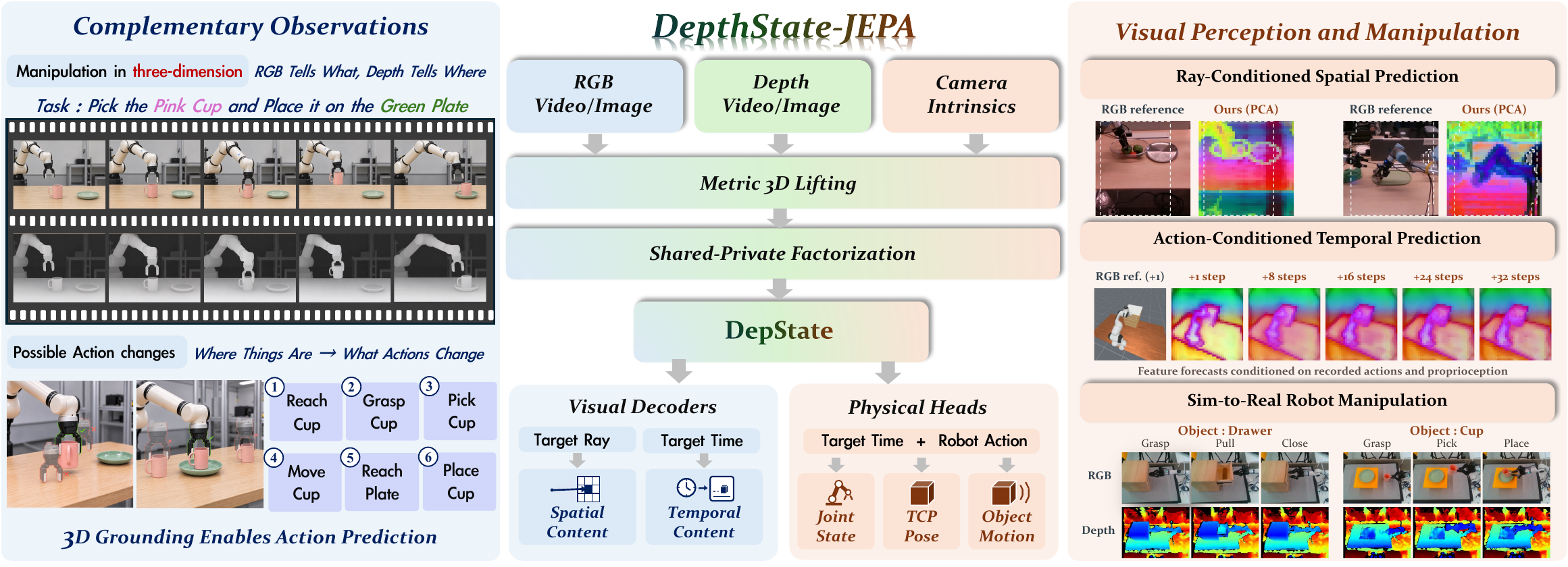}
    \vspace{-10pt}
    \caption{\textbf{From Where Things Are to What Actions Change.}
    DepState-JEPA grounds complementary RGB--depth observations in a shared latent state, DepState, while retaining modality-private features for decoding.
    This depth-grounded state connects spatial queries with action-conditioned prediction of future visual features, robot state, and object motion.}
    \label{fig:method}
    \vspace{-15pt}
\end{figure*}

\textbf{RGB JEPAs} predict representations rather than pixels \citep{assran2023self,bardes2024vjepa}:
V-JEPA~2 supports action-conditioned planning \citep{assran2025vjepa2},
while V-JEPA~2.1 strengthens dense visual features \citep{murlabadia2026vjepa21}.
\textbf{Depth-related JEPAs} offer complementary designs. LeMuMo-JEPA combines RGB
and aligned depth features within a common Transformer for multimodal
perception
\citep{cornelissen2026lemumojepa}.
Locate~3D's 3D-JEPA predicts masked point-cloud features enriched by lifted
2D features, rather than maintaining separate RGB--depth streams
\citep{arnaud2025locate3d}; JEPADepth predicts depth from RGB with auxiliary
latent supervision, rather than encoding measured depth \citep{grigore2026jepadepth}.
These designs target multimodal perception or geometric estimation, but do not establish a common state interface between spatial understanding and action-driven evolution. This motivates our work: 

{\centering
\textit{\textbf{Can complementary observations ground a spatially queryable, temporally predictive shared state representation while retaining modality-private features in separate decoding paths? }}
\par}

To answer this question, we introduce \textbf{\textit{DepState-JEPA}} (Fig.~\ref{fig:method}),
a joint-embedding predictive architecture centered on \textbf{\textit{DepState}},
a depth-grounded shared latent state connecting RGB--depth observations,
spatial readout, and action-conditioned transition. Built on V-JEPA~2.1, an independent
Depth-Video branch provides complementary observations and metric
grounding for both streams. DepState aggregates shared information,
while private paths preserve modality-specific detail for decoding
rather than forming separate evolving states. This common interface
supports reading missing spatial content and forecasting representations
after robot actions.
Representation learning uses images and video; robot prediction
initializes DepState from a single RGB-D frame and evolves it under
actions and proprioception, without accessing future RGB-D images.

Specifically, visible metric depth and camera intrinsics place both streams in metric 3D
coordinates; shared--private projections and learned aggregation form
a compact, geometrically grounded DepState. Ray-conditioned readouts
query local shared features without target depth; private features
supply modality-specific decoding detail. After action-conditioned transitions,
the same decoders predict future RGB and depth features with fixed
initial private memory. Robot-state readouts supply recursive proprioception;
object-motion supervision shapes the state. Calibrated forward kinematics
couples independent joint and end-effector predictions, propagating
consistency gradients into the transition alongside direct physical supervision.

Training proceeds in two stages: \textbf{\textit{1) RGB-D adaptation}} learns
DepState from images and video clips via masked shared-feature completion
and dense modality-feature prediction, exploiting spatial and temporal
context. \textbf{\textit{2) Robot post-training}} learns
action-conditioned transitions and physical readouts using future-state
and feature targets, physical supervision, and kinematic regularization,
while RGB-D replay retains the adaptation objectives. Evaluation
(Sec.~\ref{sec:experiments}) spans
spatial understanding, multi-step forecasting, and manipulation, with
controlled ablations testing the roles of geometry, feature sharing,
and kinematic consistency.
Our contributions are threefold:

\begingroup
\renewcommand{\baselinestretch}{1}\selectfont
\setlength{\leftmargini}{1.5em}
\setlength{\topsep}{0pt}
\setlength{\partopsep}{2pt}
\setlength{\itemsep}{1pt}
\setlength{\parsep}{0pt}
\setlength{\parskip}{0pt}
\vspace{4pt}
\begin{itemize}
\item \textbf{Dual-stream RGB--Depth Video JEPA framework.}
We introduce \textbf{\textit{DepState-JEPA}}, which connects spatial understanding and action-conditioned prediction through a shared latent state, \textbf{\textit{DepState}}. Independent RGB and depth encoders capture complementary observations, while metric depth grounds both streams in a common 3D coordinate frame.
\item \textbf{Unified DepState construction and prediction pipeline.}
We aggregate shared RGB--depth features into DepState, retaining private features for modality-specific decoding. Ray-conditioned readouts predict missing features without target depth and are reused after action-conditioned transitions to predict future features, with initial private memory fixed throughout rollout. Robot-state readouts supply recursive proprioception; object-trajectory supervision and joint--TCP consistency provide complementary training signals for state evolution.
\item \textbf{Empirical gains and controlled component analysis.}
RGB-D adaptation improves RGB-encoder transfer; the full pipeline supports feature prediction and manipulation. Within-model ablations identify contributions from depth and metric coordinates, private decoding for visual accuracy, and future-state matching for control performance.
\end{itemize}
\endgroup

\section{Related Work}
\label{sec:related_work}

\subsection{Joint-Embedding Predictive Architectures}
\label{sec:related-jepa}

Joint-embedding predictive learning uses target representations, rather than
pixels, as the objects of prediction. I-JEPA learns image representations by
predicting masked regions from visible context \citep{assran2023self}, while
V-JEPA extends this approach to spatiotemporal video features
\citep{bardes2024vjepa}. V-JEPA~2 scales video pretraining and connects learned
features to robot prediction and planning \citep{assran2025vjepa2}.
V-JEPA~2.1 further develops dense and multi-level supervision, strengthening
the spatial detail retained by video representations
\citep{murlabadia2026vjepa21}. A complementary direction changes the unit of
prediction: Causal-JEPA masks object-level latents and infers them from
surrounding context to encourage interaction-dependent prediction
\citep{nam2026causaljepa}. These approaches illustrate how prediction targets
and representation structure shape what a model learns. DepState-JEPA
extends the RGB-centric video JEPA framework along the modality axis,
introducing an independent Depth-Video stream. We retain V-JEPA~2.1's dense
feature-learning design while making measured geometry an explicit input
to shared-state construction, complementing spatial cues learned from RGB.

\subsection{Depth and 3D Representation Learning for JEPAs}
\label{sec:related-geometry}

Depth-related JEPAs differ in whether depth is predicted, used for
supervision, or encoded as geometric input.
JEPADepth combines masked latent prediction with photometric supervision
for monocular depth estimation \citep{grigore2026jepadepth}.
Depth-Regularized JEPA aligns RGB features with depth features during
training while retaining RGB-only inference \citep{khan2026depthregularized}.
Le MuMo JEPA fuses RGB and camera-aligned LiDAR depth through learnable
fusion tokens within a shared Transformer \citep{cornelissen2026lemumojepa}.
3D-JEPA in Locate~3D learns masked point-cloud representations from
RGB-D-derived geometry enriched with lifted 2D foundation-model features
\citep{arnaud2025locate3d}.
SR-JEPA predicts missing-entity features in native 3D scenes at supplied
target locations \citep{zhou2026srjepa}.
DepState-JEPA jointly encodes RGB and measured depth in independent video
streams, using visible metric depth to ground both streams in camera
coordinates. Shared--private factorization retains modality-specific
features alongside the shared state, whereas ray-conditioned readout
predicts missing content without target depth or a supplied 3D location.
The same state evolves under actions to predict future RGB-D features,
connecting geometric representation learning with spatial completion and
temporal prediction through a common interface.

\subsection{Action-Conditioned World Models for Robotics}
\label{sec:related-robotics}

Latent world models support robot control by predicting how representations
change under actions. DINO-WM learns transitions over frozen DINOv2 patch
features and optimizes action sequences toward visual goals
\citep{zhou2024dinowm}; PLDM studies JEPA-based planning from reward-free
offline trajectories \citep{sobal2025pldm}. V-JEPA~2 adapts pretrained video
features through action-conditioned robot post-training
\citep{assran2025vjepa2}, whereas LeWorldModel jointly learns its encoder
and dynamics from pixels with a distributional regularizer
\citep{maes2026leworldmodel}. A related question is how physical information
should shape the representation. MCR uses proprioceptive state--action
alignment for robot representation learning \citep{jiang2024mcr};
PSG-JEPA grounds individual states and state pairs in physical quantities
\citep{yan2026psgjepa}, and physically grounded JEPA combines inverse dynamics
with state alignment \citep{liu2026physicaljepa}.
DepState-JEPA studies consistency among predicted physical quantities in
addition to their direct supervision. Calibrated forward kinematics
couples independent joint and end-effector readouts of the predicted state;
object motion is supervised through increment-based pose trajectories.
The kinematic constraint regularizes the depth-grounded state transition without
imposing hard guarantees of physically correct dynamics.

\textbf{Concurrent work.} Closely related recent preprints advance geometric
supervision and spatial prediction \citep{khan2026depthregularized,zhou2026srjepa},
including geometry-conditioned video modeling
\citep{kim2026robotfactored,dreamx2026phi}; physical state grounding and
physics-conditioned latent dynamics
\citep{yan2026psgjepa,liu2026semigroupjepa,liu2026physicaljepa};
and joint world modeling and action generation \citep{lin2026jepawam}.
These directions complement our use of an explicit depth stream,
depth-lifted positions, and selectively shared RGB-D features for spatial,
temporal, and robot prediction.

\section{Method}
\label{sec:method}

\subsection{Problem Setup and JEPA Objective}
\label{sec:method-preliminaries}

\begin{figure}[t]
 \centering
 \includegraphics[width=0.9\linewidth]{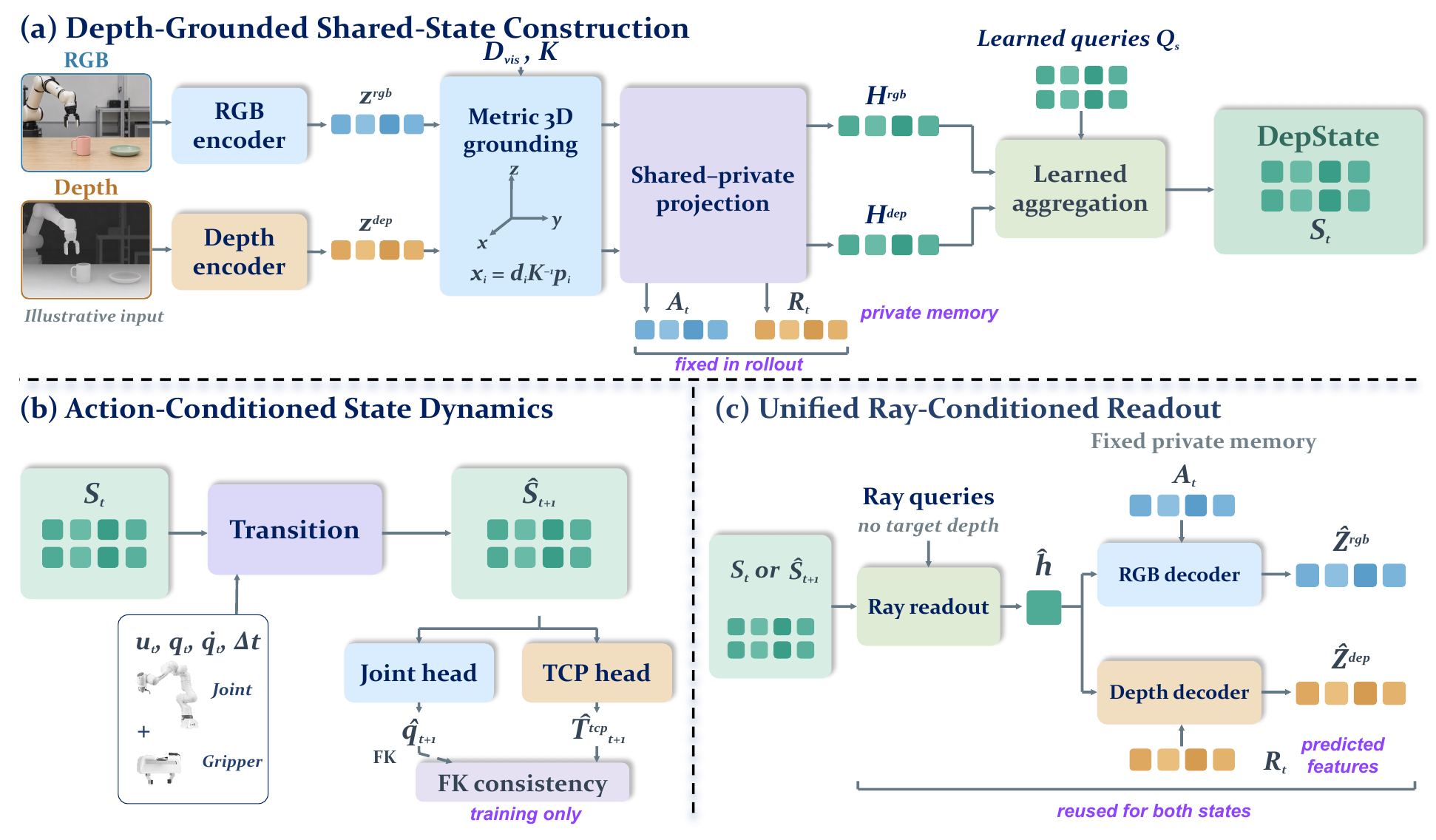}
 \caption{\textbf{Architectural Overview of DepState-JEPA.}
 RGB--Depth features are grounded, factorized, and aggregated into DepState.
 Spatial queries read local content; action-conditioned transitions advance
 the state and reuse the visual decoders with initial private features.
 Physical readouts predict robot state and object motion, with FK
 regularizing independent joint/TCP outputs.
 }
 \label{fig:method-architecture}
\end{figure}

\textbf{Problem Setup.} Given registered RGB $I$, metric depth $D$, and
intrinsics $K$, we construct DepState $S$, a shared latent state that
supports local feature readout and action-conditioned evolution
(Figure~\ref{fig:method-architecture}). Robot-state and object-motion
heads connect predicted states to physical supervision.
RGB-D adaptation uses images or video clips; robot prediction constructs
$S_t$ from the current frame only. Future RGB-D observations supply training
targets, never rollout inputs. $S$ summarizes visible scene information
for the specified readout and transition tasks.

\textbf{JEPA Objective.} JEPAs learn by predicting target representations
\citep{assran2023self}. We adopt V-JEPA~2.1's dense, multi-level objective
\citep{murlabadia2026vjepa21} for both RGB and depth, denoted throughout by
$\mathcal L_{\mathrm{JEPA}}$:
\begin{equation}
 \mathcal L_{\mathrm{JEPA}}
 =\underset{m,\ell}{\operatorname{mean}}\!
 \left[
 \frac{\sum_{i\in\mathcal M_m}e_i^{m,\ell}}{|\mathcal M_m|}
 +\frac{\sum_{i\in\mathcal V_m}w_i e_i^{m,\ell}}{|\mathcal V_m|}
 \right].
 \label{eq:depstate-jepa}
\end{equation}
Here $m$ and $\ell$ index modalities and encoder levels;
$\mathcal M_m,\mathcal V_m$ contain valid masked and visible targets.
The error $e_i^{m,\ell}$ is channel-averaged L1 between layer-normalized
predictions and stop-gradient teacher features; $w_i$ weights visible
targets by distance to masked positions (Appendix~\ref{app:depstate-losses}).
The teacher follows $\bar\theta\leftarrow\tau\bar\theta+(1-\tau)\theta$,
where $\theta$ are online representation parameters and $\tau$ is EMA retention.
This inherited objective supervises visual feature prediction; our state
construction and readout define the interface through which it is learned.

\subsection{Depth-Grounded Shared-State Construction}
\label{sec:method-factorization}

\textbf{Depth-Lifted Spatial Encoding.}
V-JEPA~2.1 RGB and depth encoders with separate weights produce
multi-level features, fused per modality into $z_i^m$,
$m\in\{\mathrm{rgb},\mathrm{dep}\}$. For a visible patch center
$\mathbf p_i=(u_i,v_i,1)^\top$ with valid metric depth $d_i$,
\begin{equation}
 \mathbf x_i=d_iK^{-1}\mathbf p_i,\qquad
 \widetilde z_i^m=z_i^m+\operatorname{PE}_{3D}(\mathbf x_i).
 \label{eq:depstate-lifting}
\end{equation}
Here $\mathbf x_i$ is the patch's metric position in the camera frame,
and $\operatorname{PE}_{3D}$ is a learned projection of Fourier-encoded
coordinates \citep{mildenhall2020nerf}. Adding it to both streams places
corresponding RGB and depth features in the same metric frame, separately
from inherited spatiotemporal RoPE \citep{su2021roformer}.
Masked tokens are removed \emph{before} encoder attention, and only
visible depth supplies online coordinates. Invalid depths receive an
unknown-geometry embedding
(App.~\ref{app:depstate-data}).

\textbf{Shared--Private Feature Projection.}
To aggregate common information without forcing all modality detail into
$S$, independent MLPs project the grounded tokens $\widetilde Z^m$ into
shared features and private decoder inputs:
\begin{equation}
 H^m=F_s^m(\widetilde Z^m),\qquad
 P^m=F_p^m(\widetilde Z^m),\qquad
 P^{\mathrm{rgb}}=A,\quad P^{\mathrm{dep}}=R .
 \label{eq:depstate-factorization}
\end{equation}
Here $F_s^m$ and $F_p^m$ are modality-specific shared and private
projections, with output widths $d_s$ and $d_p$, respectively.
Only $H^m$ enters the recurrent state and control interface; $A$ and $R$
bypass it to supply RGB- and depth-specific content to the visual decoders.
The state and private paths thus have distinct computational roles.

\textbf{Depth-Grounded Shared-State Aggregation.}
Learned queries compress the shared observation features into a
fixed-size state:
\begin{equation}
 S=G_\psi\!\left(Q_s,[H^{\mathrm{rgb}};H^{\mathrm{dep}}]\right).
 \label{eq:depstate-spatial-state}
\end{equation}
The attention aggregator $G_\psi$ uses 64 learned queries $Q_s$ of width
384 to read shared keys/values with modality/time embeddings
\citep{jaegle2021perceiverio}; $[\cdot;\cdot]$ denotes concatenation.
The resulting state varies with the observation. Query indices are generic
slots, not predefined robot joints or object identities.

\textbf{Selective-Sharing Regularization.}
Alignment encourages cross-modal agreement, decorrelation discourages
shared--private redundancy, and a variance floor opposes collapse:
\begin{equation}
 \mathcal L_{\mathrm{fac}}
 =\underset{i\in\mathcal V_{\mathrm{valid}}}{\operatorname{mean}}
 d_c(H_i^{\mathrm{rgb}},H_i^{\mathrm{dep}})
 +\lambda_{\mathrm{decorr}}
 \sum_m\frac{\|C(H^m,P^m)\|_F^2}{d_s d_p}
 +\lambda_{\mathrm{var}}\mathcal L_{\mathrm{var}}.
 \label{eq:depstate-factor-loss}
\end{equation}
Here $d_c(a,b)=1-a^\top b/(\|a\|_2\|b\|_2)$ is cosine distance, and
$\mathcal V_{\mathrm{valid}}$ contains geometrically valid visible
correspondences. $C$ is cross-covariance after channel standardization
over valid batch--position pairs; $\|\cdot\|_F$ is the Frobenius norm,
and $d_s d_p$ normalizes the number of shared--private channel pairs.
Both online branches receive gradients. The variance-floor term
$\mathcal L_{\mathrm{var}}$ \citep{bardes2022vicreg} regularizes features
and across-sample state variation; its definition and the fixed weights
are given in Appendix~\ref{app:depstate-losses}.
Cross-modal alignment acts on shared features, while the private paths
retain separate modality-specific decoding objectives.

\subsection{Ray-Conditioned Spatial Prediction}
\label{sec:method-spatial}

\textbf{Ray-Conditioned State Readout.}
Whereas aggregation writes observations into $S$, the spatial decoder
$h_\eta$ reads local shared content from $S$ alone. A target pixel supplies
a ray direction $r_i$, not its distance, and relative clip time $\nu_i$
(zero for images):
\begin{equation}
 \begin{gathered}
 r_i=K^{-1}\mathbf p_i/\|K^{-1}\mathbf p_i\|_2,\quad
 \hat h_i=h_\eta(S,\operatorname{PE}_{\mathrm{ray}}(r_i,\nu_i)),\\
 \mathcal L_{\mathrm{comp}}
 =\underset{i\in\mathcal M_{\mathrm{valid}}}{\operatorname{mean}}\,
 d_c\!\left(\hat h_i,\operatorname{sg}(h_i^\star)\right).
 \end{gathered}
 \label{eq:depstate-completion}
\end{equation}
Here $\operatorname{PE}_{\mathrm{ray}}$ embeds direction and relative time;
$\mathcal M_{\mathrm{valid}}$ selects valid masked targets for the loss,
without supplying target depth to the online query.
The target
$h_i^\star=\operatorname{norm}(\operatorname{norm}(\bar H_i^{\mathrm{rgb}})
+\operatorname{norm}(\bar H_i^{\mathrm{dep}}))$ uses full-observation
momentum features and L2 normalization; $\operatorname{sg}$ stops gradients.

\textbf{Private-Conditioned Feature Prediction.}
Modality decoders $g_m^\ell$ combine $\hat h_i$ with visible $A$ or $R$ to predict
encoder-level RGB/depth features at both masked and visible positions,
supervised by the inherited dense JEPA objective
in Eq.~(\ref{eq:depstate-jepa}).
Together these define the spatial prediction objective
\lossdef{\mathcal L_{\mathrm{spatial}}=\mathcal L_{\mathrm{JEPA}}+\mathcal L_{\mathrm{comp}}}:
completion supervises local shared features, and dense prediction anchors
modality-specific encoder features. Both operate in representation space.

\subsection{Action-Conditioned State Prediction}
\label{sec:method-transition}

\textbf{Action-Conditioned State Transition.}
A temporal Transformer advances $S_t$ using a joint command $u_t$,
configuration $q_t$, instantaneous velocity $\dot q_t$, and
physical time interval $\Delta t_t$, distinct from clip time $\nu_i$.
The existing visual readouts decode the predicted state:
\begin{equation}
 \begin{gathered}
 \hat S_{t+1}=\mathcal T_\phi(S_t,u_t,q_t,\dot q_t;\Delta t_t),\quad
 \hat Z_{t+1}^{m,\ell}
 =g_m^\ell\!\left(h_\eta(\hat S_{t+1},Q_{\mathrm{ray}}),P_t^m\right).
 \end{gathered}
 \label{eq:depstate-transition}
\end{equation}
Here $Q_{\mathrm{ray}}$ contains the known image-grid ray embeddings,
with $\nu_i=0$ for each predicted frame. $P_t^m$ is extracted from the
initial observation and held fixed across the rollout, serving visual
decoding only. Spatial and future prediction share $h_\eta,g_m^\ell$,
not the temporal Transformer.

Future-feature matching supervises the content read from the predicted
state; state matching anchors the representation reused at the next step:
\begin{equation}
 \begin{gathered}
 \mathcal L_{\mathrm{temporal}}
 =\underset{m,\ell,i}{\operatorname{mean}}\,
 \frac1c\left\|\operatorname{LN}(\hat Z_{t+1,i}^{m,\ell})
 -\operatorname{sg}(\operatorname{LN}(\bar Z_{t+1,i}^{m,\ell}))\right\|_1
 +\underset{k}{\operatorname{mean}}\,
 d_c(\hat S_{t+1,k},\operatorname{sg}(\bar S_{t+1,k})).
 \end{gathered}
 \label{eq:depstate-future-losses}
\end{equation}
Here $\bar Z$ and $\bar S$ are momentum encoder features and aggregated
states; $i$ indexes valid target positions and $k$ corresponding state
slots. LN is non-affine layer normalization and $c$ the feature width.
Both components are averaged over samples and supervised rollout steps
(Appendix~\ref{app:depstate-losses}).
A fixed calibrated external camera supplies future rays without future
depth. Control and ablation rollouts use predicted states and endpoint
proprioception; $A,R$ stay fixed at the initial observation.

\textbf{Robot-State and Object-Motion Readouts.}
Three independent endpoint heads read mean-pooled $\hat S_{t+1}$ to predict
joint configuration $\hat q_{t+1}$, instantaneous velocity
$\widehat{\dot q}_{t+1}$, and tool-center-point (TCP) pose $\hat T_{t+1}^{\mathrm{tcp}}$.
The predicted configuration and velocity condition the next rollout step.
One object-motion head reads the pooled state pair $(S_t,\hat S_{t+1})$
and its difference to predict a finite body-frame pose increment
$\widehat{\Delta\xi}_t^o$, not a velocity, for each annotated object or part $o$:
\begin{equation}
 \hat T_{t+1}^o
 =\tilde T_t^o\operatorname{Exp}([\widehat{\Delta\xi}_t^o]^\wedge).
 \label{eq:depstate-object-update}
\end{equation}
Here $\operatorname{Exp}$ maps the Lie-algebra matrix
$[\widehat{\Delta\xi}_t^o]^\wedge$ to an SE(3) transform acting on the right.
The trajectory starts from a supplied object pose and thereafter uses
predicted poses; this accumulator does not condition the visual state,
transition, or readout heads. The endpoint readouts and integrated object
poses receive valid-label supervision, grouped as \lossdef{\mathcal L_{\mathrm{sup}}}
(Appendix~\ref{app:depstate-losses}).

\textbf{Joint--TCP Kinematic Consistency.}
Independent joint and TCP predictions should describe the same robot
configuration. Direct labels anchor each output to measurements;
fixed differentiable forward kinematics (FK) constrains their agreement:
\begin{equation}
 \mathcal L_{\mathrm{FK}}
 =d_{\mathrm{pose}}(\mathrm{FK}_{\mathrm{tcp}}(\hat q_{t+1}),
                     \hat T_{t+1}^{\mathrm{tcp}}).
 \label{eq:depstate-fk}
\end{equation}
Here $d_{\mathrm{pose}}$ combines unit-normalized translation and SO(3)
geodesic errors. $\mathrm{FK}_{\mathrm{tcp}}$ includes the calibrated chain,
base transform, and tool offset; gradients through \emph{both} heads
reach the latent transition. This consistency term complements direct
joint/TCP supervision by coupling the two output heads through the same
predicted state and known robot geometry.

\subsection{Training and Rollout Protocol}
\label{sec:method-training}

RGB-D adaptation trains state construction and readout on images and
video clips using spatial and temporal context, not causal next-frame
rollouts. Robot post-training adds action-conditioned transitions and
physical heads; image/video replay continues the adaptation objective:
\begin{equation}
 \begin{gathered}
 \mathcal L_{\mathrm{adapt}}=\mathcal L_{\mathrm{spatial}}
               +\mathcal L_{\mathrm{fac}},\\
 \mathcal L_{\mathrm{robot}}=\mathcal L_{\mathrm{temporal}}
               +\mathcal L_{\mathrm{fac}}+\mathcal L_{\mathrm{sup}}
               +\lambda_{\mathrm{FK}}\gamma(n)\mathcal L_{\mathrm{FK}} .
 \end{gathered}
 \label{eq:depstate-total}
\end{equation}
The schedule $\gamma(n)$ gradually enables FK over robot updates $n$;
supervision uses valid labels. The stopped EMA teacher tracks the
representation stack, including factorization and aggregation, but not
predictors or physical heads. Control and ablation rollouts recursively
use predicted states and proprioception with fixed initial private features;
Table~\ref{tab:temporal-prediction} uses recorded proprioception instead.
Appendix~\ref{app:depstate-losses} gives weights and schedules,
Appendix~\ref{app:depstate-architecture} details the modules, and
Algorithms~\ref{alg:depstate-training}--\ref{alg:depstate-rollout}
specify training and rollout.
\label{sec:method-end}

\vspace{-5pt}
\section{Experiments}
\vspace{-5pt}
\label{sec:experiments}
We evaluate four questions:
\textbf{\textit{Q1:}} \textit{Does RGB-D adaptation improve encoder transfer and feature completion?} (Sec.~\ref{sec:exp-spatial})
\textbf{\textit{Q2:}} \textit{How accurate are visual forecasts under recorded physical conditioning?} (Sec.~\ref{sec:exp-dynamics})
\textbf{\textit{Q3:}} \textit{Do learned predictions support planning and control?} (Sec.~\ref{sec:exp-manipulation})
\textbf{\textit{Q4:}} \textit{Which components drive predictive performance?} (Sec.~\ref{sec:exp-ablation})

\subsection{Experimental Setup}
\vspace{-5pt}
\label{sec:exp-setup}
App.~\ref{app:experimental-setup} specifies task protocols and adaptations.
We compare \textbf{\textit{RGB-origin JEPAs}}: LeJEPA \citep{balestriero2025lejepa},
V-JEPA~2 \citep{assran2025vjepa2}, and V-JEPA~2.1 \citep{murlabadia2026vjepa21};
and \textbf{\textit{depth-related JEPAs}}: Locate~3D's 3D-JEPA \citep{arnaud2025locate3d},
LeMuMo-JEPA \citep{cornelissen2026lemumojepa}, and JEPADepth \citep{grigore2026jepadepth}.
These comparisons assess complete systems with their stated pretraining,
capacity, and adaptation. Within-model ablations test the designated
components. Concurrent work is discussed in Sec.~2.
\vspace{-5pt}
\begin{figure}[t]
  \centering
  \includegraphics[width=\linewidth]{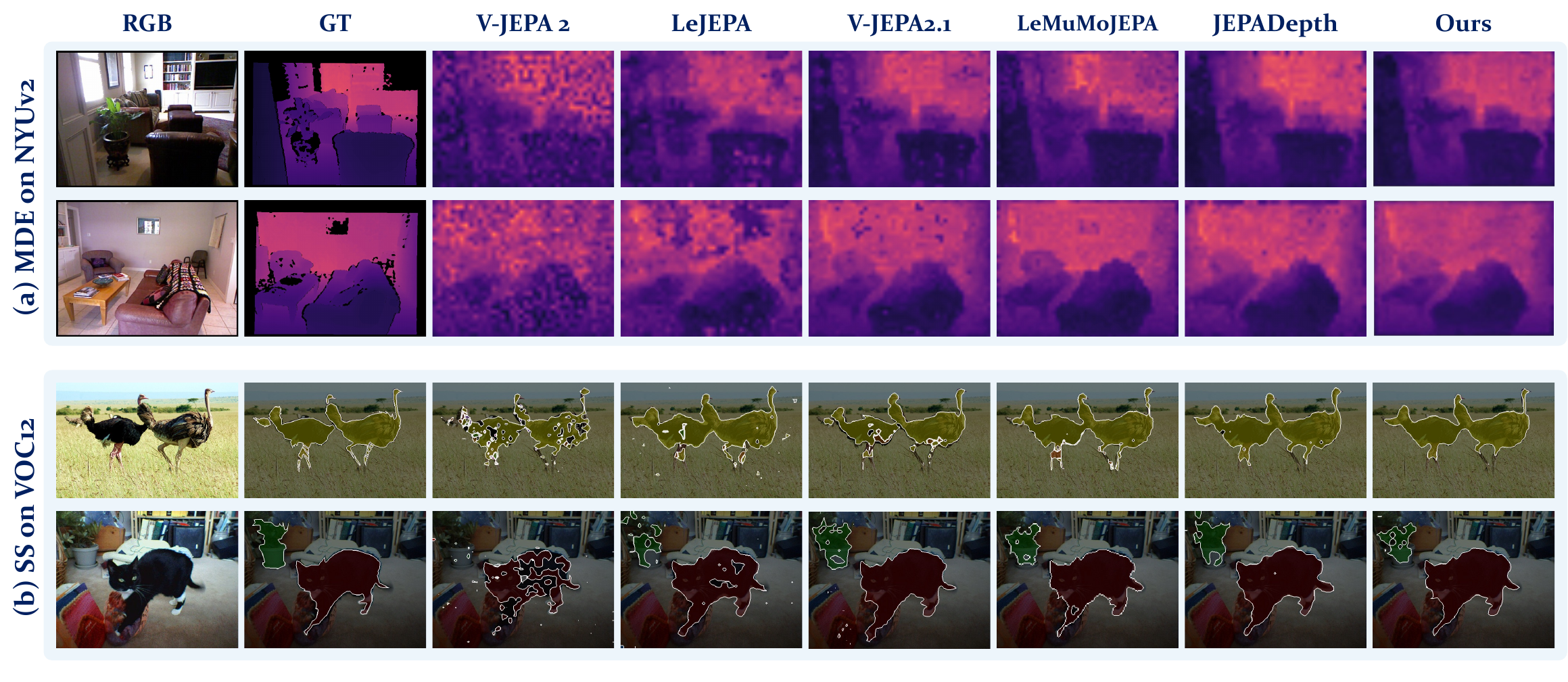}
  \vspace{-20pt}
  \caption{\textbf{Qualitative Evaluation of Frozen Spatial Representations.}
  Task-specific probes trained on frozen encoders produce
  (a) monocular depth estimates on NYUv2 and (b) semantic segmentation maps
  on VOC12. RGB inputs and ground-truth annotations provide references.
  Depth maps share a common metric color scale, and segmentation maps use
  a consistent class palette.}
  \label{fig:spatial-transfer-qualitative}
  \vspace{-15pt}
\end{figure}

\begin{table}[t]
  \centering
  \caption{\textbf{Frozen Spatial Transfer and RGB-D Feature Completion.}
  MDE and SS use frozen-encoder probes on NYUv2 and VOC12.
  SC measures masked-region \textit{L1} and \textit{Cos} against shared
  references on LingBot-Depth, with equal RGB/depth weighting. $\dagger$: method-specific adaptation.}
  \label{tab:spatial-transfer}
  \scriptsize
  \setlength{\tabcolsep}{1.7pt}
  \renewcommand{\arraystretch}{1.22}
  \setlength{\aboverulesep}{0pt}
  \setlength{\belowrulesep}{0pt}
  \resizebox{\linewidth}{!}{%
  \begin{tabular}{lcccccccc}
    \toprule
    \rowcolor{tableblue}
    & \multicolumn{4}{c}{MDE on NYUv2}
    & \multicolumn{2}{c}{SS on VOC12}
    & \multicolumn{2}{c}{SC on LB-Depth} \\
    \noalign{{\color{tableblue}\hrule height \cmidrulewidth}\kern-\cmidrulewidth}
    \cmidrule(lr){2-5}\cmidrule(lr){6-7}\cmidrule(lr){8-9}
    \rowcolor{tableblue}
    \multirow{-2}{*}{Method} & AbsRel$\downarrow$ & RMSE$\downarrow$
    & $\log_{10}\downarrow$ & $\delta_1\uparrow$
    & mIoU$\uparrow$ & mAcc$\uparrow$
    & L1$\downarrow$ & Cos.$\uparrow$ \\
    \midrule
    LeJEPA~\citep{balestriero2025lejepa}
    & 0.137 & 0.490 & 0.059 & 85.7 & 76.0 & 85.0 & 0.352 & 0.891 \\
    V-JEPA 2~\citep{assran2025vjepa2}
    & 0.188 & 0.642 & 0.079 & 77.5 & 63.9 & 76.8 & 0.428 & 0.843 \\
    V-JEPA 2.1~\citep{murlabadia2026vjepa21}
    & 0.098 & 0.350 & 0.042 & 92.5 & 84.7 & 91.8 & 0.298 & 0.931 \\
    3D-JEPA$^\dagger$~\citep{arnaud2025locate3d}
    & \multicolumn{6}{c}{\textit{Not evaluated: point-cloud input}} & 0.283 & 0.938 \\
    LeMuMo-JEPA$^\dagger$~\citep{cornelissen2026lemumojepa}
    & 0.097 & 0.345 & 0.042 & 92.7 & 84.2 & 91.6 & 0.291 & 0.936 \\
    JEPADepth~\citep{grigore2026jepadepth}
    & 0.090 & 0.323 & 0.039 & 93.8 & 84.5 & 91.7 & 0.312 & 0.923 \\
    \midrule
    \rowcolor{tableblue}
    \textbf{DepState-JEPA (Ours)}
    & \textbf{0.088} & \textbf{0.316} & \textbf{0.038} & \textbf{94.2} & \textbf{86.0} & \textbf{92.8} & \textbf{0.247} & \textbf{0.951} \\
    \bottomrule
  \end{tabular}%
  }
  \vspace{-10pt}
\end{table}

\subsection{Spatial Representation Transfer and Prediction}
\vspace{-5pt}
\label{sec:exp-spatial}
\textbf{Spatial Transfer with Frozen Encoders.}
Following V-JEPA~2.1's frozen dense-probing protocol
\citep{murlabadia2026vjepa21}, we evaluate monocular depth estimation (MDE)
on NYUv2 \citep{silberman2012indoor} and semantic segmentation (SS) on
VOC12 \citep{everingham2015pascal}. Linear probes read normalized final-layer
features. Ours uses the Stage-A RGB encoder, measuring transfer after
RGB-D adaptation with the depth branch and DepState bypassed at inference.
Readouts, preprocessing, and probe optimization are shared within each task;
V-JEPA and ours use ViT-g. App.~\ref{app:spatial-readouts} lists representation
training and probing conditions. Tab.~\ref{tab:spatial-transfer} and
Fig.~\ref{fig:spatial-transfer-qualitative} give scores and predictions.

\begin{figure}[t]
  \centering
  \includegraphics[width=\linewidth]{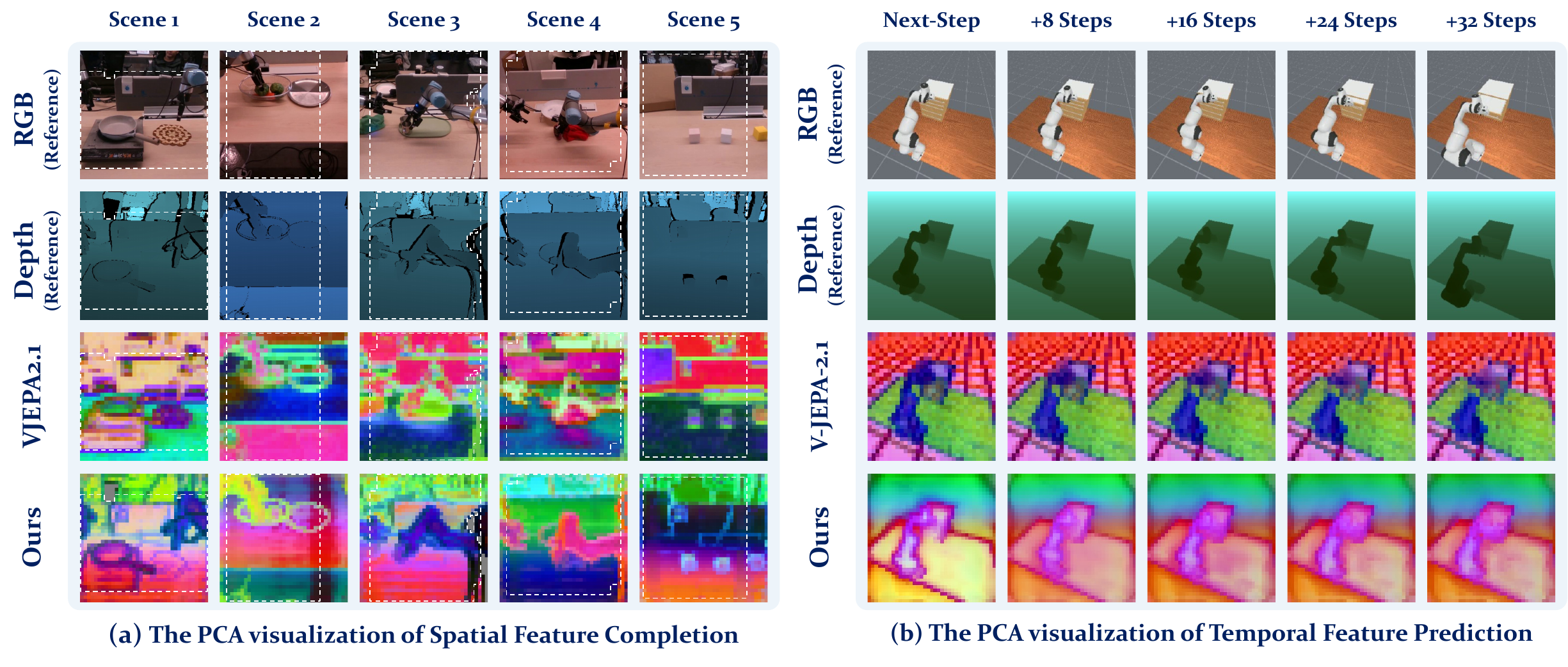}%
  \vspace{-15pt}
  \caption{\textbf{Spatial Completion and Temporal Prediction in Feature Space.}
  PCA of (a) completion from jointly masked RGB-D observations
  and (b) multi-step visual forecasts on robot trajectories.
  }
  \vspace{-10pt}
  \label{fig:spatial-missing}
\end{figure}

\textbf{Masked RGB-D Spatial Feature Completion.}
We extend V-JEPA~2.1's masked feature prediction to RGB-D spatial completion (SC) on
held-out LingBot-Depth scenes \citep{tan2026lingbotdepth}. All methods share
identical RGB/depth masks, visible context, and calibration; masked depth
is excluded from geometric lifting and query construction.
Tab.~\ref{tab:spatial-transfer} reports masked-region L1 and cosine
similarity against shared frozen RGB/depth references, each averaged equally
over modalities. Fig.~\ref{fig:spatial-missing}(a) shows PCA visualizations
of completed features in masked regions. App.~\ref{app:spatial-completion}
specifies synchronized masks, frozen reference targets, and masked-token scoring.

\vspace{-5pt}
\subsection{Temporal Visual Prediction}
\vspace{-5pt}
\label{sec:exp-dynamics}
\begin{table}[t]
  \centering
  \caption{\textbf{Single-Step and Multi-Step Prediction in a Common Feature Space.}
  Frozen readouts map predictions to a shared RGB reference space for \textit{L1}
  and \textit{Cos}. evaluation. $\dagger$: The AC adaptation.
  }
  \label{tab:temporal-prediction}
  \footnotesize
  \setlength{\tabcolsep}{2pt}
  \renewcommand{\arraystretch}{1.22}
  \resizebox{\linewidth}{!}{%
  \begin{tabular}{lcccccccccc}
    \toprule
    \rowcolor{tableblue}
    & \multicolumn{2}{c}{1 step}
    & \multicolumn{2}{c}{8 steps}
    & \multicolumn{2}{c}{16 steps}
    & \multicolumn{2}{c}{24 steps}
    & \multicolumn{2}{c}{32 steps} \\
    \rowcolor{tableblue}
    \multirow{-2}{*}{Method} & L1$\downarrow$ & Cos.$\uparrow$
    & L1$\downarrow$ & Cos.$\uparrow$
    & L1$\downarrow$ & Cos.$\uparrow$
    & L1$\downarrow$ & Cos.$\uparrow$
    & L1$\downarrow$ & Cos.$\uparrow$ \\
    \midrule
    V-JEPA 2-AC~\citep{assran2025vjepa2}
    & 0.242 & 0.916 & 0.421 & 0.792 & 0.523 & 0.701 & 0.580 & 0.646 & 0.658 & 0.598 \\
    V-JEPA 2.1-AC~\citep{murlabadia2026vjepa21}
    & 0.171 & 0.967 & 0.342 & 0.874 & 0.476 & 0.781 & 0.554 & 0.702 & 0.614 & 0.636 \\
    3D-JEPA-AC$^\dagger$~\citep{arnaud2025locate3d}
    & 0.169 & 0.968 & 0.336 & 0.879 & 0.467 & 0.789 & 0.546 & 0.711 & 0.602 & 0.647 \\
    LeMuMo-JEPA-AC$^\dagger$~\citep{cornelissen2026lemumojepa}
    & 0.167 & 0.969 & 0.331 & 0.882 & 0.479 & 0.778 & 0.563 & 0.694 & 0.624 & 0.627 \\
    \midrule
    \rowcolor{tableblue}
    \textbf{DepState-JEPA-AC (Ours)}
    & \textbf{0.145} & \textbf{0.976} & \textbf{0.307} & \textbf{0.903} & \textbf{0.394} & \textbf{0.851} & \textbf{0.451} & \textbf{0.811} & \textbf{0.491} & \textbf{0.779} \\
    \bottomrule
  \end{tabular}%
  }
  \vspace{-16pt}
\end{table}

\textbf{Next-Step Visual Prediction.}
We evaluate next-frame features on held-out robot trajectories.
Frozen linear readouts map predictions to a shared V-JEPA~2.1 RGB
reference space. Fitted on observed training frames, readouts stay frozen
across horizons. L1 and token-wise cosine similarity score the complete
prediction--readout path. App.~\ref{app:temporal-evaluation} specifies
calibration, interfaces, and action-conditioned adaptation.

\textbf{Multi-Step Visual Forecasting.}
We evaluate recursive visual predictions at $8,16,24,32$ steps from matched
starting frames. Table~\ref{tab:temporal-prediction} supplies recorded actions
and proprioception at each step, with no future images. DepState-JEPA
keeps its initial private memory fixed. Control and internal ablation
rollouts instead use predicted proprioception (App.~\ref{app:ablation-evaluation}).
Fig.~\ref{fig:spatial-missing}(b) shows feature forecasts across horizons.

\vspace{-10pt}
\subsection{Embodied Prediction and Manipulation}
\vspace{-5pt}
\label{sec:exp-manipulation}
\textbf{Online Policy Learning through Trajectories.}
Following DreamerV3 \citep{hafner2025dreamerv3}, we train an actor--critic
on imagined trajectories and adapt JEPA dynamics from online replay.
Evaluation directly executes the actor. We use 20 DMControl (DMC) and
50 Meta-World (MW) tasks with matched interaction budgets. Tab.~\ref{tab:embodied-control} reports task-averaged episode
return (the sum of rewards) for DMC and averaged success for MW.
App.~\ref{app:control-actor} specifies replay, imagination, and policy learning.

\textbf{Latent Space Receding-Horizon Planning.}
Following TD-MPC2 \citep{hansen2024tdmpc2}, we use model predictive control
(MPC): optimize action sequences through the learned latent transition
using predicted rewards and terminal values, execute the first action,
and replan at every step. Task sets, interaction budgets, and search budgets
are matched across JEPA models. Tab.~\ref{tab:embodied-control} reports
both metrics; App.~\ref{app:control-planning} specifies value learning
and trajectory optimization.

\begin{figure}[t]
  \centering
  \includegraphics[width=0.95\linewidth]{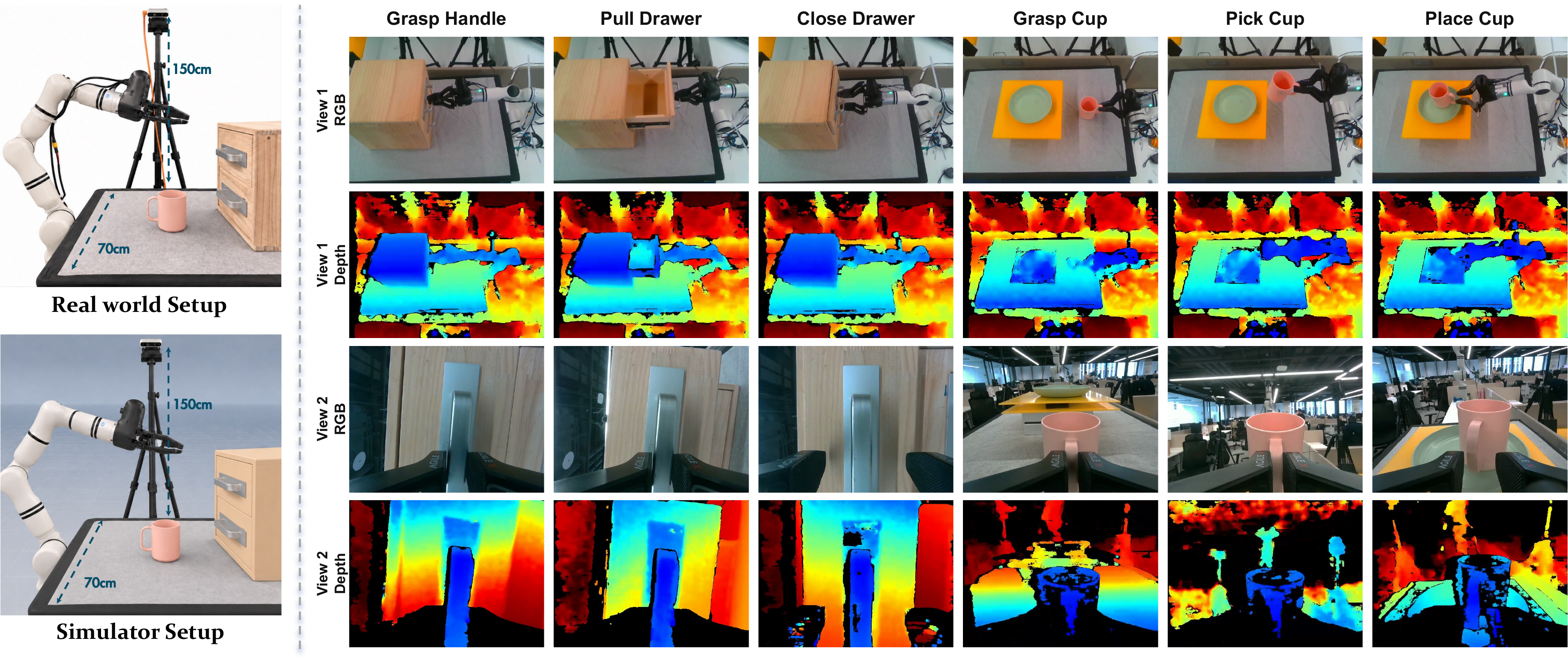}
  \vspace{-15pt}
  \caption{\textbf{Sim-to-Real Setup and Multi-View RGB-D Rollouts.}
  \textbf{Left}: simulation and real-world setups. \textbf{Right}: RGB-D observations from
  two viewpoints during real-robot manipulation tasks.}
  \label{fig:realrobot-qualitative}
  \vspace{-10pt}
\end{figure}

\textbf{Sim2real Manipulation Transfer.}
We construct matched simulation and real environments (Fig.~\ref{fig:realrobot-qualitative})
and augment simulated depth with PRISM \citep{liang2026prism}.
Simulation-trained models remain frozen for drawer and cup visual-goal MPC.
Tab.~\ref{tab:embodied-control} reports full-task success across three seeds,
with 30 trials each. The 60-second robot-execution budget excludes inference
and search. App.~\ref{app:sim2real-evaluation} specifies subgoals,
independent resets, and success aggregation.

\setlength{\floatsep}{6pt}

\begin{table}[!t]
  \centering
  \caption{\textbf{JEPA-Based Policy Learning, Planning, and Sim-to-Real Control.}
  DMC reports task-averaged \textit{episode return}; MW and Sim2Real report \textit{success rate} (\%).
  Entries are \textit{mean} $\pm$ \textit{std}.
  }
  \label{tab:embodied-control}
  \scriptsize
  \setlength{\tabcolsep}{3pt}
  \renewcommand{\arraystretch}{1.0}
  \setlength{\aboverulesep}{0pt}
  \setlength{\belowrulesep}{0pt}
  \resizebox{\linewidth}{!}{%
  \begin{tabular}{lcccccc}
    \toprule
    \rowcolor{tableblue}
    & \multicolumn{2}{c}{Online PL}
    & \multicolumn{2}{c}{Latent MPC}
    & \multicolumn{2}{c}{Sim2Real MPC} \\
    \noalign{{\color{tableblue}\hrule height \cmidrulewidth}\kern-\cmidrulewidth}
    \cmidrule(lr){2-3}\cmidrule(lr){4-5}\cmidrule(lr){6-7}
    \rowcolor{tableblue}
    \multirow{-2}{*}{Method}
    & DMC$\uparrow$ & MW$\uparrow$ & DMC$\uparrow$ & MW$\uparrow$
    & Task 1$\uparrow$ & Task 2$\uparrow$ \\
    \midrule
    V-JEPA 2-AC~\citep{assran2025vjepa2}
    & 682.4\,{\tiny\color{stdorange}\scalebox{0.75}{$\pm$}\,14.7} & 52.8\,{\tiny\color{stdorange}\scalebox{0.75}{$\pm$}\,2.8} & 736.5\,{\tiny\color{stdorange}\scalebox{0.75}{$\pm$}\,15.3} & 66.4\,{\tiny\color{stdorange}\scalebox{0.75}{$\pm$}\,2.4} & 45.6\,{\tiny\color{stdorange}\scalebox{0.75}{$\pm$}\,5.1} & 41.1\,{\tiny\color{stdorange}\scalebox{0.75}{$\pm$}\,5.1} \\
    V-JEPA 2.1-AC~\citep{murlabadia2026vjepa21}
    & 763.8\,{\tiny\color{stdorange}\scalebox{0.75}{$\pm$}\,15.4} & 68.6\,{\tiny\color{stdorange}\scalebox{0.75}{$\pm$}\,2.3} & 807.2\,{\tiny\color{stdorange}\scalebox{0.75}{$\pm$}\,15.0} & 81.2\,{\tiny\color{stdorange}\scalebox{0.75}{$\pm$}\,2.6} & 66.7\,{\tiny\color{stdorange}\scalebox{0.75}{$\pm$}\,6.7} & 64.4\,{\tiny\color{stdorange}\scalebox{0.75}{$\pm$}\,5.1} \\
    3D-JEPA-AC$^\dagger$~\citep{arnaud2025locate3d}
    & 784.4\,{\tiny\color{stdorange}\scalebox{0.75}{$\pm$}\,12.6} & 72.3\,{\tiny\color{stdorange}\scalebox{0.75}{$\pm$}\,2.6} & 826.4\,{\tiny\color{stdorange}\scalebox{0.75}{$\pm$}\,18.2} & 85.6\,{\tiny\color{stdorange}\scalebox{0.75}{$\pm$}\,3.2} & 73.3\,{\tiny\color{stdorange}\scalebox{0.75}{$\pm$}\,5.8} & 71.1\,{\tiny\color{stdorange}\scalebox{0.75}{$\pm$}\,5.1} \\
    LeMuMo-JEPA-AC$^\dagger$~\citep{cornelissen2026lemumojepa}
    & 790.4\,{\tiny\color{stdorange}\scalebox{0.75}{$\pm$}\,16.1} & 73.2\,{\tiny\color{stdorange}\scalebox{0.75}{$\pm$}\,2.5} & 837.5\,{\tiny\color{stdorange}\scalebox{0.75}{$\pm$}\,13.6} & 87.8\,{\tiny\color{stdorange}\scalebox{0.75}{$\pm$}\,2.2} & 75.6\,{\tiny\color{stdorange}\scalebox{0.75}{$\pm$}\,6.9} & 74.4\,{\tiny\color{stdorange}\scalebox{0.75}{$\pm$}\,8.4} \\
    \midrule
    \rowcolor{tableblue}
    \textbf{DepState-JEPA-AC (Ours)}& \textbf{813.2}\,{\tiny\color{stdorange}\scalebox{0.75}{$\pm$}\,14.3} & \textbf{77.1}\,{\tiny\color{stdorange}\scalebox{0.75}{$\pm$}\,2.8} & \textbf{857.8}\,{\tiny\color{stdorange}\scalebox{0.75}{$\pm$}\,17.8} & \textbf{90.1}\,{\tiny\color{stdorange}\scalebox{0.75}{$\pm$}\,2.8} & \textbf{81.1}\,{\tiny\color{stdorange}\scalebox{0.75}{$\pm$}\,5.1} & \textbf{78.9}\,{\tiny\color{stdorange}\scalebox{0.75}{$\pm$}\,3.8} \\
    \bottomrule
  \end{tabular}%
  }
\end{table}

\providecommand{\depstateablationpanel}[2]{%
  \begin{minipage}[t]{0.329\linewidth}
    \vspace{0pt}\centering
    \includegraphics[width=\linewidth]{#1}%
    \par\vspace{-1pt}{\scriptsize\textbf{#2}}%
  \end{minipage}%
}
\begin{figure}[!th]
  \centering
  \depstateablationpanel{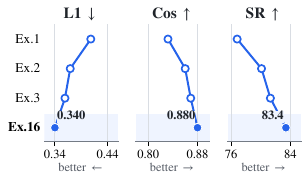}{(a) Geometry-Grounded State}%
  \hspace{0.005\linewidth}%
  \depstateablationpanel{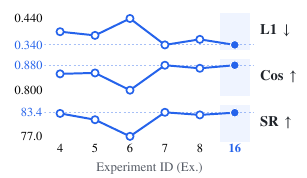}{(b) Sharing and Spatial Readout}%
  \hspace{0.005\linewidth}%
  \depstateablationpanel{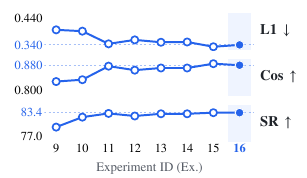}{(c) Temporal Prediction and Grounding}
  \vspace{-5pt}
  \caption{\textbf{Ablation Results.}
  \textit{L1/Cos}: 3 tasks; \textit{SR}: 50 MW + 2 real (\%). Details in App.~\ref{app:ablation-evaluation}.}
  \label{fig:ablation-study}
  \vspace{-10pt}
\end{figure}

\subsection{Ablation Study}
\vspace{-5pt}
\label{sec:exp-ablation}
Ex.1--Ex.15 vary components against the full Ex.16 configuration.
App.~\ref{app:ablation-evaluation} defines interventions and disaggregates completion,
Drawer/Cup prediction, and controller-specific success.

\vspace{-3pt}
\textbf{Effect of Depth Content and Metric Coordinates.}
Ex.1--Ex.3 and Ex.16 isolate depth content and metric coordinates under fixed
backbones and supervision. Combining both reduces L1 from 0.408 to 0.340 and
raises SR from 76.8\% to 83.4\% (Fig.~\ref{fig:ablation-study}(a)). Each factor
improves prediction and control, although their joint gains are subadditive at the reported means.

\vspace{-3pt}
\textbf{Effect of Selective Sharing and Spatial Readout.}
Ex.4--Ex.8 vary private decoding, alignment, aggregation, queries, and completion
(Fig.~\ref{fig:ablation-study}(b)). Removing private decoding raises L1 from 0.340 to 0.390,
with SR at 83.2\% versus 83.4\%. Mean aggregation gives L1 0.440 and SR 77.0\%.
Pixel/ray queries tie in feature scores; private decoding primarily benefits visual readout.

\vspace{-3pt}
\textbf{Effect of Temporal Modeling and Physical Supervision.}
Ex.9--Ex.15 vary temporal targets, physical supervision, memory, and decoders
(Fig.~\ref{fig:ablation-study}(c)). Removing future-state matching lowers SR
from 83.4\% to 79.5\%; removing FK gives 82.5\% with direct labels retained.
Separate decoders reach L1 0.333 and SR 83.4\% with added capacity.
App.~\ref{app:ablation-evaluation} relates these outcomes to each intervention.

\vspace{-10pt}
\section{Conclusion}
\vspace{-10pt}
DepState-JEPA connects RGB--depth geometry, shared-state evolution, and private decoding
for feature completion, conditioned forecasting, and manipulation. RGB-D adaptation improves
encoder transfer; ablations identify contributions from geometry, local state content, and
future-state supervision. App.~\ref{app:limitations} discusses the remaining scope.

\clearpage
\section*{Statement on the Use of AI Tools}
In this work, we used generative AI tools (Codex) solely to assist with drafting parts of the manuscript, editing the text for readability, and adjusting \LaTeX{} formatting and layout.
We did not use generative AI tools to generate synthetic datasets, develop theoretical models or conceptual frameworks, formulate mathematical claims, provide critical ingredients for proofs or assist in writing proofs, propose or refine hypotheses, design or provide feedback on research methodology or experiments, implement the DepState-JEPA method, or interpret experimental results.
All AI-assisted manuscript preparation was conducted under human direction and supervision.
The authors reviewed and revised the AI-assisted text and inspected the compiled manuscript.
We take full responsibility for the final content of this work, including all text, claims, and artifacts produced with the aid of generative AI tools.

\clearpage
\appendix
\begingroup
\hypersetup{linkcolor=citationblue}
\makeatletter
\def\Hy@colorlink#1{\begingroup\HyColor@UseColor#1}
\makeatother
\small
\setlength{\parskip}{0pt}
\tableofcontents
\endgroup
\clearpage
\section{Notation and Conventions}
\label{app:depstate}
\label{app:depstate-notation}

Tables~\ref{tab:notation-preliminaries}--\ref{tab:notation-objectives}
define the notation used in Section~\ref{sec:method}. Architectural sizes
and numerical coefficients are specified in
Appendices~\ref{app:depstate-architecture},
\ref{app:depstate-losses}, and~\ref{app:depstate-optimization}.
Subscript $t$ denotes a native control timestamp; $S_t$ is constructed
from the current frame only.

\subsection{Observations and Predictive Learning}

\begingroup
\small
\renewcommand{\arraystretch}{1.08}
\begin{longtable}{>{\raggedright\arraybackslash}p{\dimexpr0.27\linewidth-\tabcolsep\relax}>{\raggedright\arraybackslash}p{\dimexpr0.69\linewidth-\tabcolsep\relax}}
\caption{Observation, Index, and JEPA Notation.}\label{tab:notation-preliminaries}\\
\toprule
\rowcolor{tableblue}
Symbol & Definition \\ \midrule
\endfirsthead
\multicolumn{2}{l}{Table~\ref{tab:notation-preliminaries} (continued)}\\
\toprule
\rowcolor{tableblue}
Symbol & Definition \\ \midrule
\endhead
\midrule \multicolumn{2}{r}{Continued on the next page}\\ \endfoot
\bottomrule \endlastfoot
$x$, $x_{\mathcal V}$ &
Complete observation and its visible context in the JEPA objective.\\
$I,D,K$; $O_t$ &
Registered RGB, metric depth, and camera intrinsics;
$O_t=(I_t,D_t,K_t)$.\\
$\mathcal V,\mathcal M$ &
Visible and masked token-index sets.\\
$\mathcal V_{\mathrm{valid}}$,
$\mathcal M_{\mathrm{valid}}$ &
Geometrically valid visible correspondences and masked targets.\\
$i,m,\ell,t,o$ &
Token, modality, encoder level, control-time, and object/part-slot indices.\\
$f_\theta,p_\phi,f_{\bar\theta}$ &
Context encoder, predictor, and momentum target encoder.\\
$\theta,\bar\theta,\phi$ &
Online representation parameters, their momentum copy, and predictor
parameters, respectively.\\
$\tau$, $\operatorname{sg}$ &
EMA retention coefficient and stop-gradient operator.\\
$\hat{\cdot}$, $\bar{\cdot}$,
$\tilde{\cdot}$, $\mathrm{gt}$ &
Prediction; momentum target; object-trajectory integration origin;
ground-truth target, respectively.\\
\end{longtable}
\endgroup

\subsection{Spatial Representation and Decoding}

\begingroup
\small
\renewcommand{\arraystretch}{1.08}
\begin{longtable}{>{\raggedright\arraybackslash}p{\dimexpr0.27\linewidth-\tabcolsep\relax}>{\raggedright\arraybackslash}p{\dimexpr0.69\linewidth-\tabcolsep\relax}}
\caption{Spatial-State Variables, Modules, and Similarity Operators.}\label{tab:notation-spatial}\\
\toprule
\rowcolor{tableblue}
Symbol & Definition \\ \midrule
\endfirsthead
\multicolumn{2}{l}{Table~\ref{tab:notation-spatial} (continued)}\\
\toprule
\rowcolor{tableblue}
Symbol & Definition \\ \midrule
\endhead
\midrule \multicolumn{2}{r}{Continued on the next page}\\ \endfoot
\bottomrule \endlastfoot
$\mathbf p_i=(u_i,v_i,1)^\top$ &
Homogeneous coordinates of patch center $i$.\\
$d_i$, $\mathbf x_i$ &
Metric center depth and back-projected point
$\mathbf x_i=d_iK^{-1}\mathbf p_i$.\\
$z_i^m,\widetilde z_i^m$;
$\widetilde Z^m$ &
Fused feature before/after metric encoding; $\widetilde Z^m$ is the
grounded token set.\\
$\operatorname{PE}_{3D}$ &
Learned projection of Fourier-encoded metric coordinates.\\
$F_s^m,F_p^m$; $H^m,P^m$ &
Shared/private projections and their output features for modality $m$.\\
$A,R$ &
RGB/depth private features: $A=P^{\mathrm{rgb}}$ and
$R=P^{\mathrm{dep}}$.\\
$d_s,d_p$ &
Shared/private feature widths.\\
$Q_s,G_\psi,\psi$ &
Learned state queries, shared-state aggregator, and its parameters.\\
$S,S_t,\hat S_{t+1}$ &
Shared state, current shared state, and predicted next shared state.\\
$[\cdot;\cdot]$, $\operatorname{pool}$ &
Concatenation and mean pooling over state tokens.\\
$r_i,\nu_i$ &
Unit camera ray and relative token time ($\nu_i=0$ for images).\\
$\operatorname{PE}_{\mathrm{ray}}$,
$Q_{\mathrm{ray}}$ &
Ray/time encoding and query bank on the target grid.\\
$h_\eta,\eta,\hat h_i$ &
Spatial decoder, its parameters, and predicted local shared feature.\\
$\bar H_i^m,h_i^\star$ &
Momentum shared feature and fused local completion target.\\
$\operatorname{norm}$,
$d_c(a,b)$ &
L2 normalization and cosine distance
$1-a^\top b/(\|a\|_2\|b\|_2)$.\\
$C(H^m,P^m)$,
$\|\cdot\|_F$ &
Standardized shared--private cross-covariance and Frobenius norm.\\
$g_m^\ell$,
$\hat Z_{t+1}^{m,\ell}$ &
Modality readout at level $\ell$ and predicted future features.\\
$P_t^m$ &
Fixed private decoding memory from the initial observation.\\
\end{longtable}
\endgroup

\subsection{Action Conditioning and Physical Quantities}

\begingroup
\small
\renewcommand{\arraystretch}{1.08}
\begin{longtable}{>{\raggedright\arraybackslash}p{\dimexpr0.27\linewidth-\tabcolsep\relax}>{\raggedright\arraybackslash}p{\dimexpr0.69\linewidth-\tabcolsep\relax}}
\caption{Temporal, Physical, and Geometric Notation.}\label{tab:notation-physical}\\
\toprule
\rowcolor{tableblue}
Symbol & Definition \\ \midrule
\endfirsthead
\multicolumn{2}{l}{Table~\ref{tab:notation-physical} (continued)}\\
\toprule
\rowcolor{tableblue}
Symbol & Definition \\ \midrule
\endhead
\midrule \multicolumn{2}{r}{Continued on the next page}\\ \endfoot
\bottomrule \endlastfoot
$\mathcal T_\phi$ &
Action-conditioned temporal Transformer.\\
$u_t,q_t,\dot q_t,\Delta t_t$ &
Joint command, configuration, instantaneous velocity, and control interval.\\
$\hat q_{t+1}$,
$\widehat{\dot q}_{t+1}$,
$\hat T_{t+1}^{\mathrm{tcp}}$ &
Predicted endpoint configuration, velocity, and TCP pose.\\
$D_q,D_{\mathrm{tcp}}$ &
Independent joint-configuration and TCP-pose heads.\\
$\widehat{\Delta\xi}_t^o$, $D_{\mathrm{obj}}$ &
Predicted object/part body increment and its readout head.\\
$\tilde T_t^o$ &
Object/part integration origin.\\
$\hat T_{t+1}^o$ &
Integrated next object/part pose.\\
$q_{t+1}^{\mathrm{gt}}$,
$T_{t+1}^{\mathrm{tcp,gt}}$ &
Ground-truth joint configuration and TCP pose.\\
$\mathrm{SO}(3)$,
$\mathrm{SE}(3)$ &
Groups of 3D rotations and rigid transformations.\\
$\operatorname{Exp}$,
$\operatorname{Log}$,
$\wedge,\vee$ &
Lie exponential/logarithm and vector--algebra maps.\\
$\|\cdot\|_1,\|\cdot\|_2$ &
Elementwise L1 and Euclidean norms.\\
$\|\cdot\|_q^2$ &
Unit-normalized squared joint error.\\
$d_{\mathrm{pose}}$ &
Scaled translation plus SO(3) geodesic error.\\
$\mathrm{FK}_{\mathrm{tcp}}$ &
Fixed differentiable forward kinematics to the TCP.\\
\end{longtable}
\endgroup

\subsection{Objectives and Training Parameters}

\begingroup
\small
\renewcommand{\arraystretch}{1.08}
\begin{longtable}{>{\raggedright\arraybackslash}p{\dimexpr0.27\linewidth-\tabcolsep\relax}>{\raggedright\arraybackslash}p{\dimexpr0.69\linewidth-\tabcolsep\relax}}
\caption{Objectives and Schedule Symbols in the Main Method.}\label{tab:notation-objectives}\\
\toprule
\rowcolor{tableblue}
Symbol & Definition \\ \midrule
\endfirsthead
\multicolumn{2}{l}{Table~\ref{tab:notation-objectives} (continued)}\\
\toprule
\rowcolor{tableblue}
Symbol & Definition \\ \midrule
\endhead
\midrule \multicolumn{2}{r}{Continued on the next page}\\ \endfoot
\bottomrule \endlastfoot
$\mathcal L_{\mathrm{JEPA}}$ &
Dense multi-level RGB/depth latent prediction on masked and visible targets.\\
$\mathcal L_{\mathrm{comp}}$ &
Masked shared-feature completion.\\
$\mathcal L_{\mathrm{spatial}}$ &
Grouped spatial prediction: dense features plus shared completion.\\
$\mathcal L_{\mathrm{shared}}$,
$\mathcal L_{\mathrm{decorr}}$,
$\mathcal L_{\mathrm{var}}$ &
Shared alignment, shared--private decorrelation, and variance floor.\\
$\mathcal L_{\mathrm{fac}}$ &
Grouped factorization objective.\\
$\mathcal L_{\mathrm{future}}$,
$\mathcal L_{\mathrm{state}}$ &
Future feature and future state matching components.\\
$\mathcal L_{\mathrm{temporal}}$ &
Grouped temporal prediction: future features plus future state.\\
$\mathcal L_{\mathrm{jpos}},\mathcal L_{\mathrm{tcp}}$,
$\mathcal L_{\mathrm{obj}},\mathcal L_{\mathrm{vel}}$ &
Direct supervision of joints, TCP poses, object trajectories, and endpoint joint velocities.\\
$\mathcal L_{\mathrm{sup}}$ &
Grouped physical supervision.\\
$\mathcal L_{\mathrm{FK}},\lambda_{\mathrm{FK}}$ &
Joint--TCP kinematic consistency and its fixed coefficient.\\
$\mathcal L_{\mathrm{adapt}},\mathcal L_{\mathrm{robot}}$ &
RGB-D adaptation objective and robot-trajectory objective.\\
$n,\gamma(n)$ &
Robot-update count and FK warm-up coefficient.\\
\end{longtable}
\endgroup

\section{Representation Architecture and Spatial Learning}
\label{app:depstate-representation}

This section specifies the observation interface, online representation,
and momentum targets used by the spatial and temporal objectives.

\subsection{Data Interfaces, Geometry, and Masking}
\label{app:depstate-data}

\textbf{Observation and control streams.}
Representation adaptation uses registered RGB-D images and video from
LingBot-Depth (RobbySim training scenes; \citealp{tan2026lingbotdepth}),
with 16-frame clips sampled at 4\,fps. Image and video batches alternate. Depth is
metric and intrinsics are associated with each image. Clips do not cross
trajectory boundaries. Robot post-training instead uses adjacent native
control records, with their actual positive time difference
$\Delta t_t$. Each robot state is encoded from a \emph{single} RGB-D image
at the corresponding control timestamp using the image tokenizer.
It is not encoded from a bidirectional video window containing future
frames. Missing control intervals and intervals crossing resets are
excluded rather than interpolated into artificial transition targets.

The primary robot protocol uses one calibrated, fixed external RGB-D
camera, and trains a separate post-training instance for each embodiment.
Consequently, the camera-ray grid can be specified for future predictions
without knowing future camera motion. Wrist-camera-only trajectories
are outside this protocol. Visual lifting remains in camera coordinates;
TCP and object/part poses are expressed in the robot base frame.
Camera extrinsics convert annotations where necessary but do not
constitute an input of future geometry.

\textbf{Actions and proprioception.}
We preserve the dataset's joint-command semantics: absolute joint
position, relative joint position, or joint velocity. A three-component
one-hot vector identifies the control mode. A command is never substituted
for the measured next joint state. A recorded gripper command contributes
one additional channel, mapped to the data source's declared $[0,1]$
opening range, together with an availability bit. When absent, both are
zero. The action-token input therefore has dimension $n_q+5$.
Joint positions and relative commands are divided by $s_{q,j}$; velocity
commands and proprioceptive velocities by $s_{q,j}/(1\,\mathrm{s})$;
$\Delta t$ is expressed in seconds. The scales are defined in
Appendix~\ref{app:depstate-physics}.

Current proprioception comprises measured $q_t$ and a synchronized
instantaneous velocity when available. Otherwise, we use the backward
difference of consecutive measured configurations. This fallback uses
only past and current records, not future labels. Continuous revolute
coordinates are temporally unwrapped; bounded joints retain their
physical coordinates. Fixed and mimic joints are expanded according to the robot
description; all independent joints on the TCP chain must be available.

\textbf{Image transforms and depth encoding.}
RGB and depth share a random resized crop with area fraction in
$[0.7,1]$ and aspect ratio in $[0.9,1.1]$. RGB is resized bilinearly;
depth and its validity map use nearest-neighbor interpolation.
The pixel-coordinate transform $A_{\mathrm{img}}$ updates intrinsics as
$K'=A_{\mathrm{img}}K$. A clip or robot rollout uses one common geometric
augmentation across time and across online/target branches.
We do not apply horizontal flips, time reversal, metric-scale changes,
or additional RGB-only photometric augmentation.
RGB uses channel means $(0.485,0.456,0.406)$ and standard deviations
$(0.229,0.224,0.225)$ for intensities in $[0,1]$.

The depth encoder receives three identical channels containing
\begin{equation}
 D_{\mathrm{enc}}
 =2\frac{\log(\operatorname{clip}(D,0.05,100))-\log(0.05)}
          {\log(100)-\log(0.05)}-1 ,
 \label{eq:app-depth-input}
\end{equation}
where depths are expressed in meters. Invalid inputs are set to zero
after this transformation and retain a separate validity map.
Unclipped metric depth is preserved for lifting. Values outside the
declared range are not treated as valid geometric targets; clipped values
are never back-projected as if they were measured distances.

\textbf{Patch geometry and missing depth.}
A patch uses the nearest pixel to its fixed center; an invalid center
has unknown geometry. A video tubelet is anchored using the center
depth and intrinsics of its second frame.
Invalid centers receive a learned unknown-geometry embedding instead
of metric PE. Their RGB features can still contribute to the state.
Depth patches with fewer than 50\% valid pixels are excluded from
depth aggregation. Shared alignment and completion use only positions
with valid center depth and valid depth patches. Depth latent targets
require at least 50\% valid pixels; RGB latent targets do not require depth.

\textbf{Synchronized masking.}
We mask 75\% of the spatial patch grid in both modalities. Rectangles
have target area 20\% of the grid and uniformly sampled aspect ratio
in $[0.75,1.5]$, with integer dimensions fitted to the grid and uniformly
sampled valid origins. We accumulate their union until the target count
is reached, randomly subsampling the last rectangle's new cells if
necessary. After 100 attempts, any shortfall is sampled uniformly from
remaining cells. Video tubelets share the same spatial mask.
Masked embeddings are removed immediately after non-overlapping
patch/tubelet tokenization and before encoder attention, while preserving
their original indices for positional encoding.
The online branch reads metric depth only at retained positions.

The teacher sees the full augmented observation. Its target-validity
mask selects losses only; it does not select online queries or expose
target depth. Spatial queries are defined on the predetermined image grid.
Robot transition updates use unmasked current images; masked spatial
learning continues through separate representation-replay updates.

\subsection{Backbones, State Construction, and Prediction Heads}
\label{app:depstate-architecture}

\textbf{Initialization and inherited structure.}
Both encoders start from the same native V-JEPA~2.1 ViT-g/16 checkpoint
at 384 resolution \citep{murlabadia2026vjepa21}, then evolve independently.
We retain the image stem ($16\times16$ patches), video stem
($2\times16\times16$ tubelets), image/video embeddings, and
spatiotemporal RoPE. The encoder has 40 blocks, width 1408, and 22
attention heads. We use normalized outputs from blocks 10, 20, 30, and 40.
RGB-D factorization, metric PE, state aggregation, and prediction
heads are added to this backbone.

\begin{table}[htbp]
 \caption{State-Construction and Prediction Architecture. Arrows specify
 MLP widths; all new hidden MLP activations are GELU.}
 \label{tab:depstate-architecture}
 \centering
 \small
 \vspace{0.4em}
 \renewcommand{\arraystretch}{1.1}
 \begin{tabular}{p{0.27\linewidth}p{0.65\linewidth}}
 \hline
 \rowcolor{tableblue}
 Component & Configuration \\ \hline
 Multi-level fusion & Per modality: $5632\to1024\to512$, output LN. \\
 Shared / private heads & Independent per modality:
 $512\to512\to384$ / $512\to256\to128$, output LN. \\
 Metric PE & Fourier XYZ: $39\to128\to512$; shared across modalities;
 final linear layer initialized to zero. \\
 State aggregator & 64 learned queries, width 384, 2 blocks;
 query self-attention, cross-attention, and FFN. \\
 Ray-query embedding & Fourier ray plus time: $52\to384\to384$. \\
 Spatial decoder & 6 blocks, width 384; query self-attention,
 cross-attention to $S$, and FFN. \\
 Modality decoders & Two separate 2-block cross-attention/FFN decoders;
 four $384\to1408$ output maps per modality. \\
 Temporal predictor & 6 Transformer blocks, width 384;
 64 state tokens and 4 condition tokens; residual state update. \\
 Endpoint heads & Three independent $384\to256\to\text{output}$ MLPs:
 $q$, endpoint $\dot q$, and 9 TCP parameters. \\
 Object-motion head & One $1152\to512\to256\to6n_o$ MLP:
 body-frame increments for $n_o$ object/part slots. \\
 \hline
 \end{tabular}
\end{table}

\textbf{Position encoding and aggregation.}
For a vector $a\in\mathbb R^d$, define
\begin{equation}
 \Gamma(a)=\operatorname{Concat}
 \left(a,\{\sin(2^k\pi a),\cos(2^k\pi a)\}_{k=0}^{5}\right)
 \in\mathbb R^{13d}.
 \label{eq:app-fourier}
\end{equation}
Metric PE uses $\Gamma(\mathbf x/(1\,\mathrm m))$ and is added after
four-level fusion but before factorization.
A spatial query uses the concatenation of $\Gamma(r)$ and
$\Gamma(\nu)$, where $r$ is a unit ray and $\nu\in[0,1]$ is relative
token time ($\nu=0$ for images).
An independent $13\to384\to384$ temporal MLP and a modality embedding
are added to the aggregator's shared-feature keys/values.
Queries are generic learned vectors: their indices do not encode a
robot joint, object part, or contact category.

All new attention modules use six heads, pre-LN, FFN expansion ratio
four, and zero attention dropout and stochastic depth.
LN uses $\epsilon=10^{-6}$; final shared, private, and state
normalizations have no affine parameters. Other than specified
initializations, new linear weights use truncated normal with standard
deviation 0.02 and zero biases. We do not use BatchNorm.

\textbf{Private-conditioned decoding.}
The spatial decoder accesses $S$ only. Its output queries feed the
modality decoders, whose cross-attention memory comprises corresponding
private features projected from 128 to 384 dimensions, with ray/time PE.
During training, the entire private cross-attention residual of a modality
is disabled with probability 0.5, without inverse-probability rescaling;
the query residual remains. Each trajectory uses the same dropout draw
across rollout steps. Both private paths are enabled at inference.
The same spatial and modality decoders serve spatial denoising and future
prediction. Future prediction never reads future private features.

\textbf{Temporal conditioning and readouts.}
Each condition uses an input-width$\to384\to384$ MLP and a type
embedding. Input widths are $n_q+5$, $n_q$, $n_q$, and 1 for action,
configuration, instantaneous velocity, and time interval.
State tokens receive learned slot embeddings. The first 64 Transformer
outputs are linearly mapped to a residual:
\begin{equation}
 \hat S_{t+1}=\operatorname{LN}
 \left(S_t+W_{\mathrm{out}}
 [\operatorname{Transformer}_\phi(S_t,\mathrm{conditions})]_{1:64}
 +b_{\mathrm{out}}\right).
 \label{eq:app-state-update}
\end{equation}
$W_{\mathrm{out}}$ and $b_{\mathrm{out}}$ are zero-initialized.
The spatial and temporal predictors do not share parameters.
Endpoint heads receive mean-pooled $\hat S_{t+1}$. The object-motion input is
$[\operatorname{pool}(S_t);
\operatorname{pool}(\hat S_{t+1});
\operatorname{pool}(\hat S_{t+1}-S_t)]$, of width 1152.
The object-motion head outputs $6n_o$ components, where $n_o$ is the
number of available object/part label slots for that training instance.
These are readout slots, not typed tokens in the visual state, and do
not require edges in a part-topology graph.

\subsection{Momentum Targets and Gradient Boundaries}
\label{app:depstate-teacher}

We initialize the target branch as a copy of the online representation
branch and update it after each optimizer step:
$\bar\theta\leftarrow0.99925\bar\theta+0.00075\theta$.
The copied parameters comprise both encoders, multi-level fusion,
metric and aggregation-time PE, unknown-geometry embeddings,
shared/private projections, generic queries, and the aggregator.
Predictors, ray-query decoder embeddings, and physical heads have no
target counterparts. FK parameters are fixed.
The teacher uses evaluation mode and no gradient computation.
EMA updates occur once per optimizer step, not once per accumulated
micro-batch.

For completion, the target at a valid location is the L2-normalized sum
of the individually normalized target shared features.
For dense latent matching, targets are the four encoder-level outputs
\emph{before} metric PE and factorization.
For temporal matching, the teacher independently processes each full
future image and aggregates
$\bar S_{t+h}=G_{\bar\psi}(Q_{\bar s},
[\bar H_{t+h}^{\mathrm{rgb}};\bar H_{t+h}^{\mathrm{dep}}])$.
The state target loss is
\begin{equation}
 \mathcal L_{\mathrm{state}}=\frac1{64}\sum_{k=1}^{64}
 d_c(\hat S_{t+h,k},\operatorname{sg}(\bar S_{t+h,k})).
 \label{eq:app-state-target}
\end{equation}
Online and teacher slots have persistent corresponding query indices;
they are not arbitrarily rematched at each timestep.
All target features and states are stopped. In contrast, online
cross-modal alignment and both prediction sides of each physical
consistency retain gradients. Inference uses the online encoder, state constructor, predictors, and heads.

\section{Training Losses and Physical Consistency}
\label{app:depstate-objectives}

The following definitions make the objective reductions and physical
conventions explicit. Missing annotations mask their own loss groups;
they do not change the online prediction path.

\subsection{Complete Loss Definitions}
\label{app:depstate-losses}

Losses are averaged over valid samples, target positions, modalities,
levels, and rollout steps as appropriate. Each group is normalized before
its coefficient is applied. A group with no valid entries is skipped.

\textbf{Cross-covariance and variance floor.}
For $N$ sampled valid context positions, we center and channel-standardize
the rows of $H^m$ and $P^m$:
\begin{equation}
 \widehat F_{id}
 =\frac{F_{id}-\mu_d(F)}{\sqrt{\operatorname{Var}(F_{\cdot d})+10^{-4}}},
 \qquad
 C_m=\frac{(\widehat H^m)^\top\widehat P^m}{N-1}.
 \label{eq:app-covariance}
\end{equation}
At most 4096 positions are sampled uniformly per statistical batch;
the covariance is skipped if fewer than two are valid.
Following the variance-floor principle of \citet{bardes2022vicreg}, define
\begin{equation}
 v(F)=\frac1d\sum_{j=1}^d
 \max\left(0,1-\sqrt{\operatorname{Var}(F_{\cdot j})+10^{-4}}\right).
 \label{eq:app-variance}
\end{equation}
We average $v$ over the four shared/private feature sets and a state term.
The state term is computed \emph{across samples for each fixed slot},
then averaged over slots. Computing it only across slots would allow
different input-independent query constants to pass this regularizer.
During robot updates, the state term averages the current-state value
and the predicted-state values at all supervised horizons.
Define
$\mathcal L_{\mathrm{shared}}=
\operatorname{mean}_{i\in\mathcal V_{\mathrm{valid}}}
d_c(H_i^{\mathrm{rgb}},H_i^{\mathrm{dep}})$ and
$\mathcal L_{\mathrm{decorr}}=\sum_m\|C_m\|_F^2/(d_sd_p)$.
The combined objective in Eq.~(\ref{eq:depstate-factor-loss}) is
\lossdef{\mathcal L_{\mathrm{fac}}=
\mathcal L_{\mathrm{shared}}+
\lambda_{\mathrm{decorr}}\mathcal L_{\mathrm{decorr}}+
\lambda_{\mathrm{var}}\mathcal L_{\mathrm{var}}},
with $\lambda_{\mathrm{decorr}}=0.005$ and $\lambda_{\mathrm{var}}=0.1$.
The two-modality averaging factor is absorbed into
$\lambda_{\mathrm{decorr}}$.
Distributed statistics use differentiable all-gather over the global
micro-batch of 16 samples; eight accumulated steps form the effective
optimization batch of 128. These two batch sizes are not interchangeable.

\textbf{Dense spatial prediction.}
Let $c=1408$, $\ell\in\{10,20,30,40\}$, and
\begin{equation}
 e_i^{m,\ell}
 =\frac1c\left\|
 \operatorname{LN}(\hat z_i^{m,\ell})
 -\operatorname{sg}(\operatorname{LN}(\bar z_i^{m,\ell}))
 \right\|_1 ,
 \label{eq:app-latent-error}
\end{equation}
where LN has no affine parameters. Substituting these errors into
Eq.~(\ref{eq:depstate-jepa}) gives the dense multi-level JEPA objective
used in RGB-D adaptation and representation replay. The visible weights are
\begin{equation}
 w_i=\frac{\lambda_{\mathrm{vis}}(n)}
 {\sqrt{\max(1,d_{\min}(i,\mathcal M))}} .
 \label{eq:app-dense-loss}
\end{equation}
$\mathcal M_m,\mathcal V_m$ include modality-valid target positions;
$d_{\min}$ is Euclidean distance on the spatial patch grid to the closest
masked patch. The mask is shared across video tubelets.
During the first 5000 RGB-D adaptation updates, $\lambda_{\mathrm{vis}}$ increases
linearly from zero to 0.7 for images or 0.5 for videos, then stays fixed.
The completion objective in Eq.~(\ref{eq:depstate-completion}) is averaged
only over valid masked geometric targets and has coefficient one.
The spatial objective is
\lossdef{\mathcal L_{\mathrm{spatial}}=\mathcal L_{\mathrm{JEPA}}+\mathcal L_{\mathrm{comp}}};
its two components retain their distinct targets and reductions.

\textbf{Future prediction and supervision.}
$\mathcal L_{\mathrm{future}}$ averages Eq.~(\ref{eq:app-latent-error})
over all valid future target positions, both modalities, all four levels,
and the supervised horizons. Future grids are not partitioned into
visible and masked positions. Together with the state matching defined
in Eq.~(\ref{eq:app-state-target}), this gives
\lossdef{\mathcal L_{\mathrm{temporal}}=\mathcal L_{\mathrm{future}}+\mathcal L_{\mathrm{state}}},
with both coefficients equal to one.

Direct physical supervision uses measured endpoint configurations and TCP
poses, integrated object poses, and instantaneous endpoint joint velocities:
\begin{equation}
 \begin{gathered}
 \mathcal L_{\mathrm{jpos}}
 =\operatorname{mean}_{h}\|\hat q_{t+h}-q_{t+h}^{\mathrm{gt}}\|_q^2,\\
 \mathcal L_{\mathrm{tcp}}
 =\operatorname{mean}_{h}d_{\mathrm{pose}}
 (\hat T_{t+h}^{\mathrm{tcp}},T_{t+h}^{\mathrm{tcp,gt}}),\\
 \mathcal L_{\mathrm{obj}}
 =\operatorname{mean}_{o,h}d_{\mathrm{pose}}
 (\hat T_{t+h}^o,T_{t+h}^{o,\mathrm{gt}}),\\
 \mathcal L_{\mathrm{vel}}
 =\operatorname{mean}_{h,j}
 \left[\frac{\widehat{\dot q}_{t+h,j}-\dot q_{t+h,j}^{\mathrm{gt}}}
 {s_{q,j}/(1\,\mathrm s)}\right]^2 .
 \end{gathered}
 \label{eq:app-direct-supervision}
\end{equation}
Each mean includes only valid annotations and also averages over samples.
An object trajectory is supervised only where a valid initial pose and
corresponding endpoint labels are available. Endpoint velocity targets
and physical scales are specified in Appendix~\ref{app:depstate-physics}.
The supervision group and robot objective are
\begin{equation}
 \begin{gathered}
 \mathcal L_{\mathrm{sup}}
 =\mathcal L_{\mathrm{jpos}}+\mathcal L_{\mathrm{tcp}}
    +\mathcal L_{\mathrm{obj}}+\lambda_{\mathrm{vel}}\mathcal L_{\mathrm{vel}},\\
 \mathcal L_{\mathrm{robot}}
 =\mathcal L_{\mathrm{temporal}}+\mathcal L_{\mathrm{fac}}
   +\mathcal L_{\mathrm{sup}}+\lambda_{\mathrm{FK}}\gamma(n)\mathcal L_{\mathrm{FK}},\\
 \gamma(n)=\operatorname{clip}((n-2000)/3000,0,1).
 \end{gathered}
 \label{eq:app-physical-grouping}
\end{equation}
We set $\lambda_{\mathrm{vel}}=0.25$ for the endpoint velocity objective
and $\lambda_{\mathrm{FK}}=0.1$; the other supervision coefficients are one.
The FK loss is averaged over valid samples and supervised rollout steps,
using the calibrated robot model and valid joint/TCP outputs.
$n$ counts robot optimizer updates, excluding replay. Supervision and
latent matching are enabled from the first robot update; the FK term
has zero weight for the first 2000, reaches its full weight at 5000,
and remains fixed thereafter.

\subsection{Physical Readouts, Units, and Kinematic Consistency}
\label{app:depstate-physics}

\textbf{Norms and physical units.}
For an $n_q$-dimensional joint residual $a$,
$\|a\|_q^2=n_q^{-1}\sum_j(a_j/s_{q,j})^2$, with
$s_{q,j}=1\,\mathrm{rad}$ for revolute joints and $0.1\,\mathrm m$
for prismatic joints. Pose distances use
$s_p=0.1\,\mathrm m$ and $s_R=1\,\mathrm{rad}$.
Joint velocity scales are $s_{q,j}/(1\,\mathrm s)$.
These fixed scales render the joint and velocity errors dimensionless.

For $T_a=(R_a,p_a)$ and $T_b=(R_b,p_b)$, the pose distance used throughout
the main text and algorithms is
\begin{equation}
 d_{\mathrm{pose}}(T_a,T_b)
 =\frac{\|p_a-p_b\|_2^2}{s_p^2}
 +\frac{\|\operatorname{Log}(R_a^\top R_b)^\vee\|_2^2}{s_R^2}.
 \label{eq:depstate-pose-distance}
\end{equation}

\textbf{Endpoint velocity targets.}
Instantaneous endpoint velocity is supervised by aligned measurements
when available and otherwise by a backward difference using observations
up to that endpoint. Revolute trajectories are unwrapped before
finite differences are computed. This readout supplies the next
transition's proprioceptive velocity input. The object-motion readout is
trained through the accumulated object pose objective in
Eq.~(\ref{eq:app-direct-supervision}).

\textbf{SE(3) convention and initialization.}
$T^o={}^bT_o$ maps object/part coordinates into the robot base frame.
Body increments use angular-first order $\Delta\xi=[\omega^\top,v^\top]^\top$ and
$[\Delta\xi]^\wedge=
\left[\begin{smallmatrix}[\omega]_\times&v\\0&0\end{smallmatrix}\right]$.
Body increments act on the right. At the first supervised step,
$\tilde T_t^o=T_t^{o,\mathrm{gt}}$; subsequent integration uses predicted
poses, not refreshed labels. Joint configurations are predicted directly
by the endpoint head and passed to subsequent transitions.
The object pose initializer does not condition the visual state,
temporal predictor, or object-motion head. Without an initial object pose,
the model can output relative increments but not an absolute object
trajectory. Part identity is given by consistent annotation slots;
missing slots are masked and do not introduce a topology requirement.

Object trajectories use the SE(3) update in
Eq.~(\ref{eq:depstate-object-update}), supervised through accumulated poses.
Rotation logarithms use the principal branch; Exp/Log use FP32 with
stable small-angle branches.

\textbf{Independent TCP prediction.}
The TCP head outputs translation and two rotation vectors
$a_1,a_2\in\mathbb R^3$. We use the continuous 6D rotation representation
of \citet{zhou2019continuity}:
\begin{equation}
 b_1=\frac{a_1}{\|a_1\|_2},\quad
 \tilde b_2=a_2-(b_1^\top a_2)b_1,\quad
 b_2=\frac{\tilde b_2}{\|\tilde b_2\|_2},\quad
 R=[b_1,b_2,b_1\times b_2].
 \label{eq:app-tcp-rotation}
\end{equation}
The rotation-output bias starts at $(1,0,0,0,1,0)$.
Normalization uses $\epsilon=10^{-6}$.
The independent joint and TCP heads are anchored to the actual dataset
labels, not to commands or TCP labels regenerated by our FK model.

\textbf{FK model and calibration.}
With fixed joint origins/axes, tool transform, and a declared reference
frame, differentiable FK composes the complete robot chain:
\begin{equation}
 \mathrm{FK}_{\mathrm{tcp}}(q)
 ={}^{r}T_b
  \left(\prod_{j=1}^{J}{}^{j-1}T_j(q)\right)
  {}^{J}T_{\mathrm{tcp}} .
 \label{eq:app-fk-chain}
\end{equation}
$J$ includes the chain's fixed and dependent joints, whose transforms
are determined by the robot description; $r=b$ for base-frame outputs.
The TCP tool offset is included even when the final actuated link is a
flange. No FK parameter is optimized.

FK consistency requires compatible base/tool frames, joint order,
units, zero offsets, and timestamps. Independent TCP annotations need not be generated
by FK, but must describe the same tool and time. During optimization, neither predicted side of
$\mathcal L_{\mathrm{FK}}$ is detached. One gradient path passes through
the fixed differentiable chain into $D_q$; the other passes into
$D_{\mathrm{tcp}}$. Both reach $\mathcal T_\phi$ and the online
representation, coupling joint and TCP predictions, not TCP contact
with the object.

\section{Optimization, Algorithms, and Inference}
\label{app:depstate-procedures}
\raggedbottom

\subsection{Optimization Protocol}
\label{app:depstate-optimization}

\begin{table}[htbp]
 \caption{Fixed Optimization Protocol. Counts denote optimizer updates,
 not micro-steps.}
 \label{tab:depstate-training}
 \centering
 \small
 \vspace{0.4em}
 \renewcommand{\arraystretch}{1.1}
 \begin{tabular}{p{0.23\linewidth}p{0.335\linewidth}p{0.335\linewidth}}
 \hline
 \rowcolor{tableblue}
 Setting & RGB-D Adaptation & Robot Post-training \\ \hline
 Initialization & Pretrained encoders; new state and decoding modules &
 Adapted online/teacher; new temporal predictor and physical heads \\
 Total updates & 35,000 & 20,000; every fourth update is replay \\
 Resolution & 256 for 30,000 updates; 384 for the last 5000 &
 384 for robot data and replay \\
 Temporal input & Images / 16-frame videos at 4\,fps &
 Single image per robot state; 16-frame replay videos \\
 Effective / global micro-batch & 128 / 16 & 128 / 16 \\
 Peak encoder LR & $2\cdot10^{-5}$ & $2\cdot10^{-6}$ \\
 Peak new-module LR & $2\cdot10^{-4}$ & $5\cdot10^{-5}$ \\
 LR warm-up & 2000 updates & 1000 total updates \\
 LR decay & Constant to 30,000, then cosine to $0.1\times$ peak &
 Cosine after warm-up to $0.1\times$ peak \\
 Robot horizon & Not applicable & 1 for 2000 robot updates, then 3 \\
 EMA coefficient & 0.99925 & 0.99925 \\
 \hline
 \end{tabular}
\end{table}

\textbf{RGB-D Adaptation.}
We initialize the state modules and modality decoders and optimize
$\mathcal L_{\mathrm{adapt}}$. During the first 2000 updates, all pretrained encoder
parameters are frozen except the depth image/video stems. Those stems
use the new-module learning rate during this interval, while all new
modules are trainable. We then unfreeze both encoders and assign their
parameters the encoder learning rate. The last 5000 updates increase
spatial resolution but retain 16-frame video input.

\textbf{Robot Post-training and Replay.}
Robot post-training retains the adapted online and target representation branches,
adds the temporal predictor and physical readouts. Every fourth optimizer update
uses $\mathcal L_{\mathrm{adapt}}$ on training-split representation data; the remaining
15,000 updates use $\mathcal L_{\mathrm{robot}}$ on robot trajectories.
Image and video replay alternate by \emph{replay index}. Visible-target weights are already
at their full values. The new-module learning-rate group for robot post-training
includes the state and decoding modules learned during RGB-D adaptation.
The FK consistency weight uses the robot-update ramp in
Eq.~(\ref{eq:app-physical-grouping}).

\textbf{Optimizer and arithmetic.}
Both stages use AdamW with $\beta=(0.9,0.95)$, $\epsilon=10^{-8}$,
weight decay 0.04, and global gradient-norm clipping at 1.0.
Biases, normalization parameters, and learned embeddings have no weight
decay. Eight global micro-batches of 16 are accumulated per optimizer
update. Encoder and attention arithmetic use BF16; geometry,
orthogonalization, FK, Exp/Log, and physical losses use FP32.
Target updates follow optimizer updates.

\textbf{Rollout curriculum.}
Algorithm~\ref{alg:depstate-rollout} defines the open-loop training path.
The horizon is one for the first 2000 robot updates and three thereafter.
Horizon losses have equal weight; recorded future commands condition
the rollout but future measurements never replace recursive predictions.

\textbf{Run selection and result scope.}
Control comparisons and Ex.1--Ex.16 use seeds 0, 1, and 2 and the final
checkpoint of each fixed-length run. The task-specific probes
and common-reference readouts use seed 0 under the separate settings in
Appendix~\ref{app:experimental-setup}. The reported tables distinguish
single-run probe scores from means and sample standard deviations across
three training seeds.

\subsection{Two-Stage Training Algorithm}
\label{app:depstate-algorithms}

\textbf{Notation and spatial objective.}
Write $O_t=(I_t,D_t,K_t)$ and let $\mathcal R_\theta(O;\mathcal V)$
return $(S,A,R)$ and intermediate features through encoding, lifting,
factorization, and aggregation. $\Omega$ contains all initialized
online parameters; only $\theta$ has a teacher copy $\bar\theta$.
$\operatorname{SpatialLoss}$ computes $\mathcal L_{\mathrm{adapt}}$ using synchronized
masking before attention, visible-only metric lifting, ray querying,
and private-conditioned decoding. Full-observation stopped targets
supervise local shared and pre-fusion encoder features.
Augmentation, validity reductions, and private dropout follow
Appendices~\ref{app:depstate-data}, \ref{app:depstate-architecture},
and~\ref{app:depstate-losses}.

\begin{algorithm}[H]
\caption{Two-Stage Training of DepState-JEPA}
\label{alg:depstate-training}
\small
\begin{algorithmic}[1]
\Require RGB-D training data $\mathcal D_{\mathrm{rgbd}}$; valid robot trajectories $\mathcal D_{\mathrm{robot}}$;
V-JEPA~2.1 initialization; fixed calibrated FK
\Ensure Online representation, shared decoders, temporal predictor, and physical heads
\State Initialize independent RGB/depth encoders, state modules, and shared decoders
\State $\bar\theta\gets\theta$; set teacher to evaluation mode with gradients disabled
\For{$s\in\{\mathrm{adapt},\mathrm{robot}\}$}
  \If{$s=\mathrm{robot}$}
    \State Initialize $\mathcal T_\phi$ and physical heads; $n\gets0$; $r\gets0$
  \EndIf
  \For{$k=1,\ldots,N_s$} \Comment{$N_{\mathrm{adapt}}=35000$, $N_{\mathrm{robot}}=20000$}
    \State Set resolution, learning rates, and freezing using Table~\ref{tab:depstate-training}
    \State $\mathrm{isRobot}\gets(s=\mathrm{robot}\ \mathrm{and}\ k\bmod4\ne0)$
    \If{$\mathrm{isRobot}$}
      \State $n\gets n+1$; $H\gets1$ if $n\le2000$, else $3$
    \Else
      \If{$s=\mathrm{robot}$}
        \State $r\gets r+1$; $j_{\mathrm{rep}}\gets r$
      \Else
        \State $j_{\mathrm{rep}}\gets k$
      \EndIf
      \State Select image/video batches by alternating $j_{\mathrm{rep}}$
    \EndIf
    \State Zero online gradients
    \For{$b=1,\ldots,8$} \Comment{Global micro-batch size $16$}
      \If{$\mathrm{isRobot}$}
        \State Sample an $H$-step trajectory from $\mathcal D_{\mathrm{robot}}$; apply one common geometric transform
        \State $(\mathcal P,\mathcal L)\gets\operatorname{Rollout}(H,n,\mathrm{train})$
          \Comment{Algorithm~\ref{alg:depstate-rollout}}
      \Else
        \State Sample the selected observation type $O\sim\mathcal D_{\mathrm{rgbd}}$
        \State $\mathcal L\gets\operatorname{SpatialLoss}(O)$
      \EndIf
      \State Backpropagate $\mathcal L/8$ into online parameters
    \EndFor
    \State Clip global gradient norm at $1$; take one AdamW step on active $\Omega$
    \State $\bar\theta\gets0.99925\bar\theta+0.00075\theta$ \Comment{Once per optimizer update}
  \EndFor
\EndFor
\State \Return the online stack; discard teacher from deployment
\end{algorithmic}
\end{algorithm}

\subsection{Open-Loop Prediction and Training Algorithm}
\label{app:depstate-open-loop}

\textbf{Recursive rollout and optional supervision.}
Algorithm~\ref{alg:depstate-training} passes the sampled trajectory's
initial inputs and action/time sequence to the rollout below.
$D_{\mathrm{end}}$ groups independent heads on pooled next state;
$D_{\mathrm{obj}}$ predicts object increments from the pooled state pair
and difference. $\mathcal P_h$ stores neural predictions and, when
initialized, integrated object poses. Only the training branch reads
future observations and labels.

\begin{algorithm}[H]
\caption{Action-Conditioned Recursive Prediction with Optional Training Losses}
\label{alg:depstate-rollout}
\small
\begin{algorithmic}[1]
\Require Initial $O_t,q_t,\dot q_t$; commands/times
$\{u_{t+h},\Delta t_{t+h}\}_{h=0}^{H-1}$; online stack; mode; robot-update index $n$
\Require Optional initial object poses $T_t^o$; teacher, future $O_{t+1:t+H}$ and labels
\emph{only in training mode}
\Ensure Predictions $\{\mathcal P_h\}_{h=1}^H$; training loss $\mathcal L_{\mathrm{robot}}$ or $0$ at inference
\State $(S^0,A^0,R^0)\gets\mathcal R_\theta(O_t;\mathrm{all})$ \Comment{Current frame only}
\State $q^0\gets q_t$; $v^0\gets\dot q_t$; $P_0^{\mathrm{rgb}}\gets A^0$;
$P_0^{\mathrm{dep}}\gets R^0$
\State Build $Q_{\mathrm{ray}}$ from the fixed calibrated image grid \Comment{No future depth}
\State If available, initialize object trajectory accumulator $T^{o,0}\gets T_t^o$
\State Sample one private-path dropout draw per modality in training; enable both paths at inference
\For{$h=0,\ldots,H-1$}
  \State $S^{h+1}\gets\mathcal T_\phi(S^h,u_{t+h},q^h,v^h;\Delta t_{t+h})$
  \State $(q^{h+1},v^{h+1},T^{\mathrm{tcp},h+1})\gets D_{\mathrm{end}}(S^{h+1})$
  \State $\Delta\xi^{o,h}\gets D_{\mathrm{obj}}(S^h,S^{h+1})$
  \State $U^{h+1}\gets h_\eta(S^{h+1},Q_{\mathrm{ray}})$
  \State $\hat Z_{t+h+1}^{m,\ell}\gets g_m^\ell(U^{h+1},P_0^m)$ for all $m,\ell$
    \Comment{Fixed initial private memory}
  \If{initial object pose is available}
    \State $T^{o,h+1}\gets T^{o,h}\operatorname{Exp}([\Delta\xi^{o,h}]^\wedge)$
      \Comment{No pose feedback into $\mathcal T_\phi$}
  \EndIf
  \State Store $S^{h+1}$, $\hat Z_{t+h+1}$, endpoint readouts, object increments, and available poses in $\mathcal P_{h+1}$
  \If{training mode}
    \State Compute stopped encoder/state targets from full $O_{t+h+1}$ with the teacher
    \State Record future-feature and state-matching errors
    \State Record valid direct-label losses and FK residuals
      \Comment{Eqs.~(\ref{eq:app-direct-supervision}) and~(\ref{eq:depstate-fk})}
  \EndIf
  \State Reuse $S^{h+1},q^{h+1},v^{h+1}$ at the next step \Comment{No future measurements}
\EndFor
\If{training mode}
  \State Form $\mathcal L_{\mathrm{fac}}$ from initial features and current/predicted-state variances
  \State Reduce valid losses using Appendix~\ref{app:depstate-losses}, averaging supervised horizons
  \State $\mathcal L_{\mathrm{robot}}\gets\mathcal L_{\mathrm{temporal}}+
    \mathcal L_{\mathrm{fac}}+\mathcal L_{\mathrm{sup}}+\lambda_{\mathrm{FK}}\gamma(n)\mathcal L_{\mathrm{FK}}$
\Else
  \State $\mathcal L_{\mathrm{robot}}\gets0$
\EndIf
\State \Return $\{\mathcal P_h\}_{h=1}^H,\mathcal L_{\mathrm{robot}}$
\end{algorithmic}
\end{algorithm}

\noindent\textbf{Gradient and information boundaries.}
No recursive state, endpoint proprioception, or integrated pose is
detached during training. Both sides of the FK consistency
backpropagate through online heads and transition; only teacher targets
are stopped. Physical operations use FP32. Empty loss groups are skipped
and factorization is added once. Without an initial object pose, the model
returns relative object increments but no absolute object trajectory.

\subsection{Inference Interfaces}
\label{app:depstate-inference}

RGB-only perception uses the RGB-D-adapted RGB encoder without accessing
depth, metric PE, or the RGB-D state. The primary frozen probe reads the
final layer. These probes evaluate transfer of the adapted RGB encoder,
not a direct readout of $S$. RGB-D completion instead uses the spatial
readout and private decoding paths defined in Section~\ref{sec:method-spatial}.
Robot policies read state tokens and current proprioception, and
predictive planners can query the same action-conditioned transition.
Joint/TCP heads remain available at inference without target labels.
Relative object increments do not require object-pose input,
but reconstructing absolute trajectories requires an initial pose.
The teacher and losses are not deployed.

\section{Experimental Settings, Evaluation Protocols, and Additional Results}
\label{app:experimental-setup}
We describe the datasets, metrics, readouts, and method-specific
adaptations used in our evaluation.

\subsection{Datasets for Frozen Spatial Transfer}
\label{app:spatial-datasets}

\textbf{NYUv2: monocular depth estimation.}
NYU Depth V2 \citep{silberman2012indoor} is used through two separate
training views. Supervised depth probes use the BTS-extracted RGB--depth
training split of the DINOv3 evaluation pipeline, which is distinct from
the 795-image labeled training subset. LeJEPA uses only RGB from this split for
in-domain self-supervision. JEPADepth instead uses raw RGB sequences from
training scenes, with explicit target/previous/next timestamps and camera
calibration. Training excludes all test scenes and their adjacent frames.
Probe training uses a fixed task-specific configuration and the training
split shared across methods. All methods are evaluated on the
same 654 labeled test images. Depth units, registration, and label-validity
rules are fixed before training. Test depth is used only for scoring, never for scale alignment or model selection.

\textbf{VOC12: semantic segmentation.}
PASCAL VOC 2012 \citep{everingham2015pascal} provides 1,464 segmentation
training images and 1,449 validation images, with 20 foreground categories
and one background class. The main protocol trains probes on the
official training split and reports results on the official validation
split, denoted \emph{VOC12 val}, rather than the hidden test set.
Probe hyperparameters are fixed across methods, with one configuration
per method--task pair. Augmented
segmentation labels from external datasets are not included in this
track. Label 255 denotes void and is excluded from both training
and evaluation. VOC12 does not provide registered metric depth; sensor
depth or depth reconstructed from test labels cannot be supplied to a
baseline.

\subsection{Metrics for Frozen Spatial Transfer}
\label{app:spatial-metrics}

\textbf{Depth errors and threshold accuracy.}
Let $d_i$ and $\hat d_i$ be ground-truth and predicted depth in meters at
pixel $i$. Predictions are bilinearly restored to the original label grid and scored
with the shared DINOv3 \texttt{NYU\_EIGEN} mask and depth range
$[0.001,10]$\,m. Ground-truth maps are not resized for scoring.
The depth columns of Table~\ref{tab:spatial-transfer} follow a single
supervised frozen-probe protocol based on V-JEPA~2.1
\citep{murlabadia2026vjepa21}. All participating methods, including
JEPADepth, fit the same depth probe using training GT and predict metric
depth from RGB alone. Predictions are clipped to the declared range after
resolution restoration; no per-image median scaling or test-label scale
calibration is applied. Nonfinite predictions and empty valid masks trigger
explicit evaluation failures rather than silently omitted observations.
For the $N$ valid pixels of one image, define
\begin{align}
 \mathrm{AbsRel}
 &=\frac1N\sum_{i=1}^{N}\frac{|\hat d_i-d_i|}{d_i},
 \label{eq:spatial-absrel}\\
 \mathrm{RMSE}
 &=\sqrt{\frac1N\sum_{i=1}^{N}(\hat d_i-d_i)^2},
 \label{eq:spatial-rmse}\\
 \mathrm{log}_{10}
 &=\frac1N\sum_{i=1}^{N}|\log_{10}\hat d_i-\log_{10}d_i|,
 \label{eq:spatial-log10}\\
 \delta_1
 &=\frac{100}{N}\sum_{i=1}^{N}
 \mathbf{1}\!\left[
 \max\!\left(\frac{\hat d_i}{d_i},\frac{d_i}{\hat d_i}\right)<1.25
 \right].
 \label{eq:spatial-delta}
\end{align}
Metrics are computed per image and then averaged equally over the
654 test images. RMSE is reported in meters; AbsRel and $\mathrm{log}_{10}$
are dimensionless; $\delta_1$ is reported as a percentage. Lower errors
and higher threshold accuracies indicate better performance. RMSE is
the primary depth endpoint, with AbsRel and the other metrics describing
relative errors and their distribution. These definitions follow common
metric-depth evaluation conventions \citep{bhat2023zoedepth}.

\textbf{Semantic segmentation metrics.}
Let $C_{ab}$ count non-void pixels with ground-truth class $a$ and
predicted class $b$, accumulated over the entire VOC12 validation split.
With $K=21$ classes, including background,
\begin{equation}
 \mathrm{mIoU}
 =\frac{100}{K}\sum_{c=0}^{K-1}
 \frac{C_{cc}}
 {\sum_{b=0}^{K-1}C_{cb}+\sum_{a=0}^{K-1}C_{ac}-C_{cc}}.
 \label{eq:spatial-miou}
\end{equation}
All 21 classes have ground-truth support in the full validation set.
In addition to mIoU, we report mean class accuracy (mAcc):
\begin{equation}
 \mathrm{mAcc}
 =\frac{100}{K}\sum_{c=0}^{K-1}
 \frac{C_{cc}}{\sum_{b=0}^{K-1}C_{cb}}.
 \label{eq:spatial-macc}
\end{equation}
Both scores are percentages and higher is better. mIoU is the primary
segmentation endpoint: it measures overlap and penalizes both false
positives and false negatives. mAcc measures per-class recall with equal
class weights and complements mIoU. Both exclude void pixels and include
background.
We aggregate the confusion matrix over the full validation set before
computing metrics, rather than averaging per-image scores. Probe logits are bilinearly resized to the original
image grid before taking the class argmax; label-map transformations use
nearest-neighbor interpolation. Single-scale evaluation without horizontal
flip averaging or a conditional random field is used for all rows.

\subsection{Frozen Spatial Transfer: Readouts and Training}
\label{app:spatial-readouts}

\textbf{Comparison scope and shared adaptation.}
Each RGB encoder receives separately fitted depth and segmentation probes;
3D-JEPA participates only in RGB-D completion. Within each task, methods
share splits, masks, evaluators, output restoration, probe architecture,
augmentation, sample budgets, and optimizer settings; input feature widths
follow the native encoders. Pretraining, initialization, capacity, and
supervision differ across methods. In Table~\ref{tab:spatial-transfer},
$\dagger$ marks Le MuMo JEPA's RGB-only retraining and 3D-JEPA's ray-query
adaptation (Appendix~\ref{app:spatial-completion}).

\textbf{LeJEPA representation training.}
Separate task-specific encoders use the native invariance and SIGReg
objectives. Training uses two global $224^2$ and six local $96^2$ crops,
AdamW with learning rate $5\times10^{-4}$, weight decay 0.05,
SIGReg weight 0.05, projection width 128, statistical batch 64,
and 100 epochs. Initialization and data for all methods are summarized below.

\begin{samepage}
\textbf{Representation conditions for frozen transfer.}
The following summary separates representation learning from the common
supervised probes in Table~\ref{tab:spatial-transfer}.
\begin{center}
\footnotesize
\setlength{\tabcolsep}{3pt}
\renewcommand{\arraystretch}{1.12}
\begin{tabular}{p{\dimexpr0.19\linewidth-2\tabcolsep\relax}p{\dimexpr0.24\linewidth-2\tabcolsep\relax}p{\dimexpr0.38\linewidth-2\tabcolsep\relax}p{\dimexpr0.19\linewidth-2\tabcolsep\relax}}
\toprule
Method & Initialization & Representation learning before probing & Probe input \\
\midrule
V-JEPA 2 / 2.1 & Released ViT-g & Released checkpoint; no extra in-domain SSL & RGB features \\
LeJEPA & Random ViT-S/16 & Task-specific training RGB; 100 SSL epochs & RGB features \\
Le MuMo RGB-only & Random ViT-S/16 & Task-specific training RGB; 100 SSL epochs; second input zero & Fusion tokens \\
JEPADepth & DINOv3-Small & NYUv2 raw RGB; native self-supervision; 20 epochs & Context features \\
DepState-JEPA & V-JEPA 2.1 ViT-g & 35,000 Stage-A RGB-D updates & RGB branch only \\
\bottomrule
\end{tabular}
\end{center}
\end{samepage}
All encoders are frozen during depth/segmentation probe fitting; task
labels update only the probes. Thus the MDE/SS columns describe transfer
of the resulting encoders, whereas SC evaluates a state-and-readout
pipeline. The depth-content and coordinate contrasts in
Table~\ref{tab:ablation-geometry} hold the dual-backbone setting and
supervision fixed within DepState-JEPA.

\textbf{Frozen features and RGB-only inputs.}
For every frozen probe, the encoder remains in evaluation mode with all
parameters excluded from the optimizer. We extract normalized final-layer
spatial features as $B\times C\times H_p\times W_p$, removing nonspatial
tokens; intermediate layers, global pooling, and SSL projectors are excluded.
V-JEPA~2.1 uses its single-frame image branch. V-JEPA~2 uses repeated copies
of the same image when required by its temporal stem, without accessing
other frames. All image-to-token wrappers preserve the true rectangular
patch grid. Le MuMo JEPA supplies final fusion tokens. Neither RGB-only
task receives measured depth or intrinsics at inference, and no pseudo-depth
is generated. NYUv2 GT depth and VOC labels supervise only their respective
probe heads; they do not update the frozen representations.

\textbf{Common depth probe and post-training.}
We reuse the DINOv3 linear-depth pipeline: final normalized patch
features, learnable head BatchNorm, a 256-channel linear output,
the native depth conversion, and a scale-invariant logarithmic loss.
Probe training uses $416\times544$ crops and AdamW for 4,800 updates
with a global batch size of 128. The learning rate is $3\times10^{-4}$ and weight decay is
$10^{-4}$, with one fixed hyperparameter configuration and random seed 0.
Depth PNG values are converted to meters using 1,000
stored units per meter. Evaluation uses the \texttt{NYU\_EIGEN} mask,
without test-time augmentation or median scaling.
The quantitative table and qualitative depth panel use the same probe
checkpoints and inference settings. Depth and segmentation probes are
trained separately.

\textbf{Common segmentation probe and post-training.}
All segmentation methods use a $1\times1$ linear classifier on normalized
last-layer spatial features, dropout 0.1, no extra head BatchNorm, and
bilinear logit restoration. Training uses unweighted cross-entropy over
21 classes, including background 0, with ignore index 255 and no label
reduction. Labels use nearest-neighbor resizing and padding value 255.
The common recipe uses $512^2$ crops, random scale $[0.5,2]$, horizontal
flips, and AdamW for 5,000 updates with a global batch size of 128. The learning rate is
$5\times10^{-4}$ and weight decay is 0.01. Each method uses this single
configuration with random seed 0.
Evaluation preserves aspect ratio, resizes the shorter side to 512, uses
$512^2$ sliding windows with stride 341, restores the original label grid,
and applies no test-time augmentation. All 1,449 validation images are
evaluated; the hidden official test split is not used. These task-specific
settings adapt the frozen-probing protocol of V-JEPA~2.1 to VOC12.

\textbf{JEPADepth: native self-supervision followed by frozen probing.}
JEPADepth retains DINOv3-Small initialization, its FPN depth decoder,
PoseNet, masked predictor, EMA target, and photometric, smoothness, and JEPA
losses. It is retrained on NYUv2 raw RGB videos, without GT depth, true
poses, or stereo supervision during representation training.
After this stage, we retain and freeze the context encoder, remove its
native depth decoder, PoseNet, predictor, and EMA target from the evaluation
path, and train the same supervised depth probe used for the other methods.
Thus, the table evaluates JEPADepth features under the V-JEPA~2.1 probing
protocol, rather than the complete native self-supervised depth estimator.
GT depth is absent from representation-training losses but is used for
subsequent probe training, as for all other NYUv2 rows. The learned probe
provides metric predictions without test-time scale correction.
NYUv2 intrinsics follow image preprocessing; training triplets contain
distinct frames from the same sequence.
The task-specific configuration uses $384\times512$ images, batch 12, AdamW
with learning rate $10^{-4}$ and weight decay 0.01, 20 epochs with a
0.1 decay at epoch 15, JEPA weight 1, smoothness weight 0.001, and EMA
momentum increasing from 0.996 to 1. The output range is $[0.1,10]$\,m.
Photometric inputs retain original colors;
network-input color augmentation is shared across each triplet.

For VOC12 transfer, the same NYUv2-self-supervised JEPADepth context
encoder is frozen and receives the common segmentation probe trained on
VOC12 train. VOC12 still images do not supply its photometric video
objective; no VOC depth or pseudo-depth is introduced. This evaluates
cross-dataset feature transfer, not a native JEPADepth segmentation model.

\textbf{Le MuMo JEPA$^\dagger$: RGB-only adaptation.}
For each task, we train a separate randomly initialized ViT-S/16 fusion
encoder on its training RGB (the NYUv2 training split or VOC12 train),
retaining the native modality stems, fusion tokens, token routing,
joint-view consistency, and SIGReg.
The second-modality tensor is exactly zero throughout SSL, probe fitting,
and inference, constructed after RGB preprocessing without depth
normalization. Biases and modality/position embeddings remain trainable;
zero input does not imply zero intermediate tokens. No depth supervision,
pseudo-depth, or probabilistic depth dropout is used. The task-specific recipe
uses two $224^2$ global and eight $96^2$ local crops, projection width 16,
SIGReg weight 0.1, learning rate $10^{-4}$, batch 64, and 100 SSL epochs;
modality-specific auxiliary SIGReg is disabled. After SSL, the frozen
spatial fusion tokens feed the common depth or segmentation probe.
The table marks this RGB-only adaptation with $\dagger$.

\textbf{Qualitative comparison.}
Figure~\ref{fig:spatial-transfer-qualitative} presents RGB inputs,
ground-truth annotations, and task-specific frozen-probe predictions.
Panel (a) uses a common metric color range for depth estimates.
No depth map is independently rescaled for visualization.
Panel (b) uses fixed sample IDs/crops and a common class palette for
semantic segmentation. Both panels use the same task-specific probe
checkpoints and inference settings as quantitative evaluation.
Figures~\ref{fig:spatial-transfer-mde-additional}
and~\ref{fig:spatial-transfer-ss-additional} extend this comparison with six
NYUv2 scenes and six VOC12 examples, respectively. Rows align RGB inputs, ground truth, and each method's predictions.

\begin{figure}[p]
  \centering
  \includegraphics[width=\linewidth]{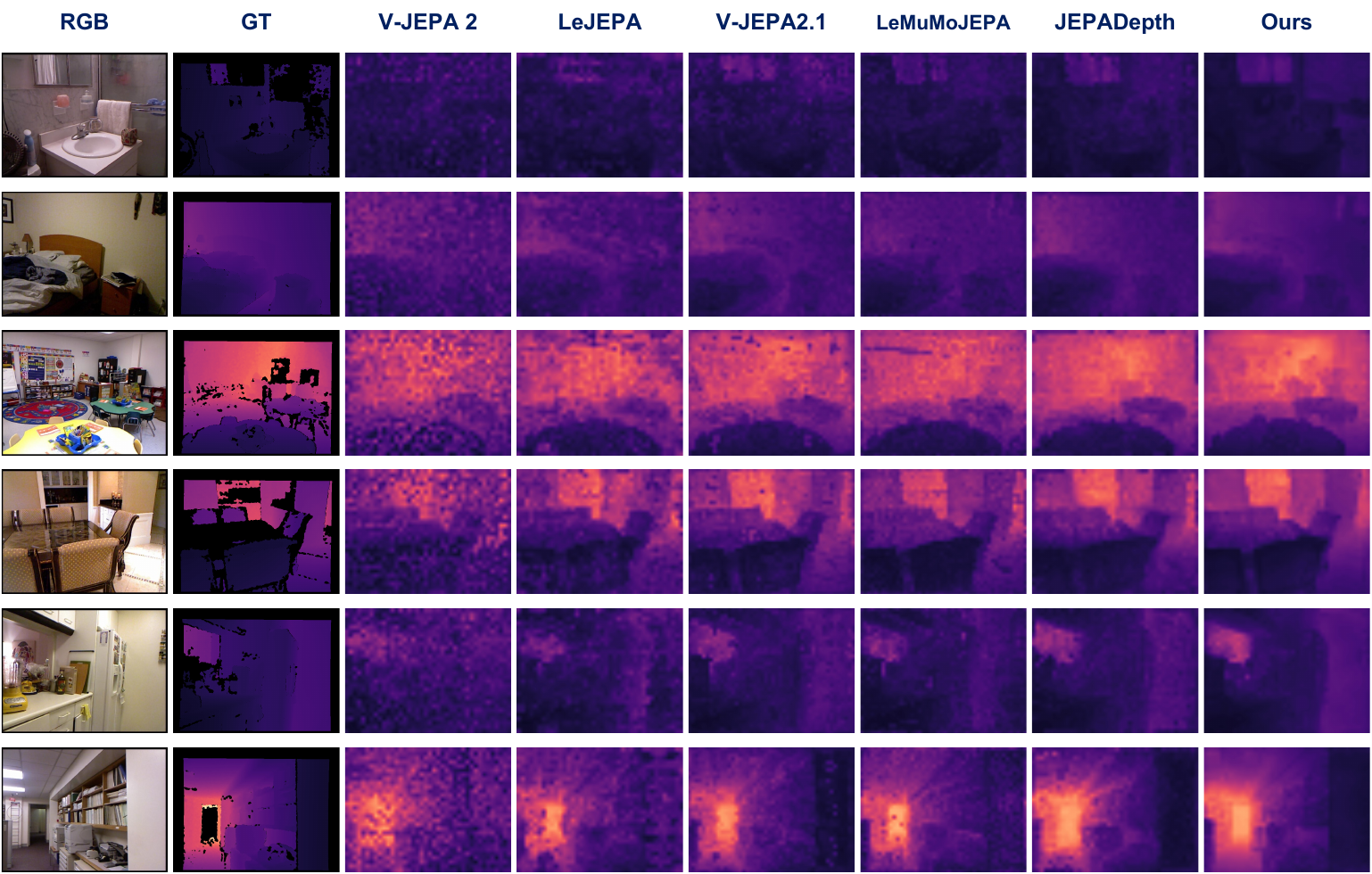}
  \caption{\textbf{Additional Monocular Depth Estimation Results on NYUv2.}
  RGB inputs, ground-truth depth (GT), and frozen-probe predictions for six indoor scenes.}
  \label{fig:spatial-transfer-mde-additional}
\end{figure}

\begin{figure}[p]
  \centering
  \includegraphics[width=\linewidth]{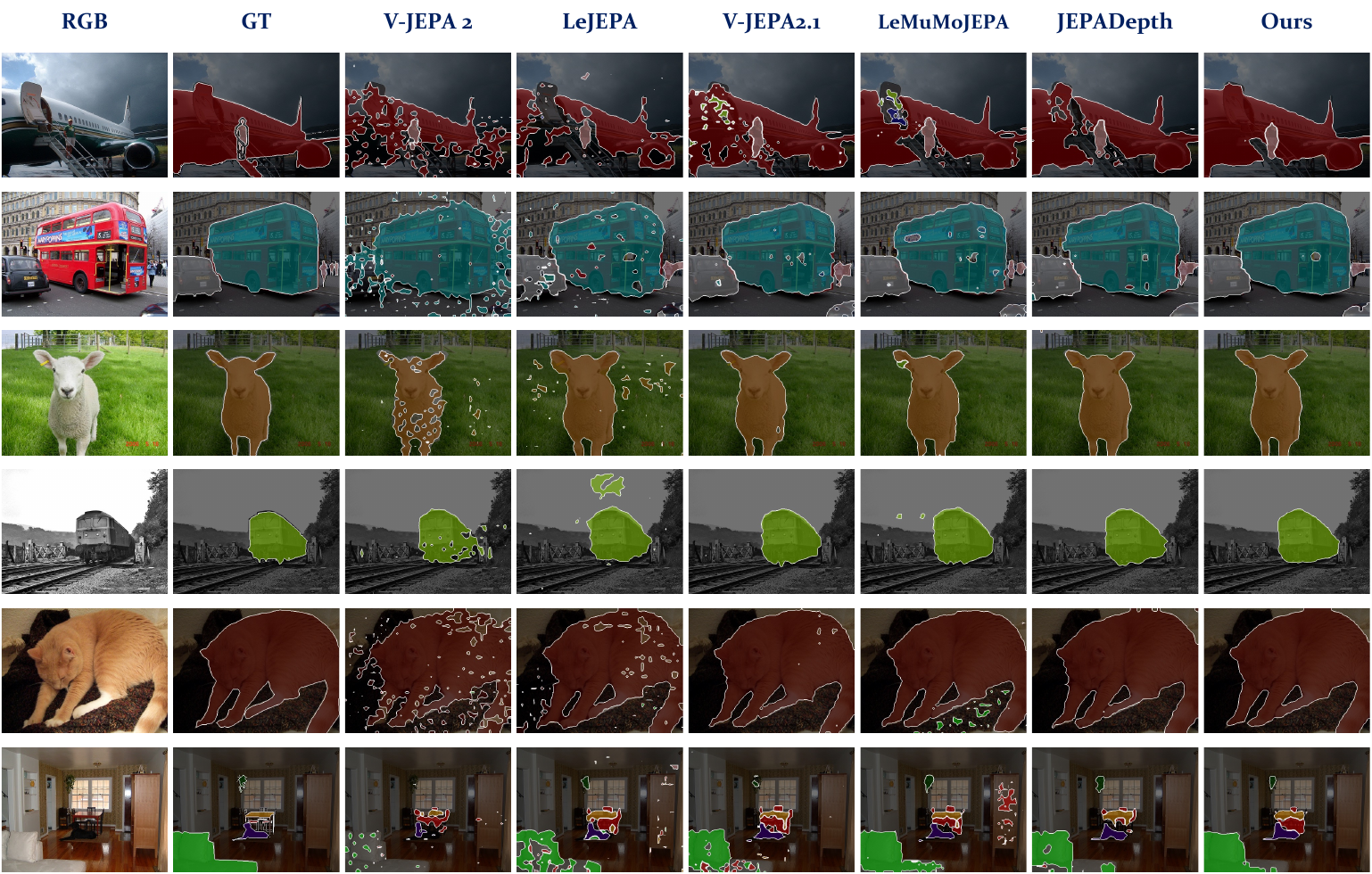}
  \caption{\textbf{Additional Semantic Segmentation Results on VOC12.}
  Ground-truth and predicted class regions are overlaid on RGB images for six examples.}
  \label{fig:spatial-transfer-ss-additional}
\end{figure}

\clearpage

\subsection{RGB-D Spatial Completion}
\label{app:spatial-completion}

\textbf{LingBot-Depth completion data and split.}
We evaluate synchronized RGB-D masked-feature completion using
LingBot-Depth (\texttt{robbyant/mdm\_depth}) \citep{tan2026lingbotdepth}.
RobbySim supplies training data and RobbySimVal supplies held-out evaluation;
shared scene identifiers, alternate views, and adjacent frames are excluded
from training. An inner scene-disjoint training split is used for development.
All representation adaptation, reference-encoder preparation, and completion
training exclude evaluation scenes. RGB, reference depth, and intrinsics
are registered; resizing and cropping update intrinsics consistently.
Reference depth provides the controlled visible context and target signal,
so this protocol measures missing-content inference rather than sensor-noise
removal. The completion adaptation below is separate from each backbone's native pretraining.

\textbf{Synchronized masks and permitted context.}
A union of patch-aligned rectangular blocks covers approximately 50\% of
the image grid, with aspect ratios sampled in $[1/2,2]$ and five fixed mask
seeds per evaluation image. The same spatial mask is applied to RGB and
depth, and is shared across all methods. The image geometry and common
mask grid are fixed before evaluation; online tokens whose receptive fields
intersect a masked block cannot read original pixels from that block.
Hidden content is removed before online feature extraction, cross-modal
attention, metric lifting, state aggregation, or private-memory construction.
Zero-filling is accompanied by an explicit visibility mask; it is not
interpreted as a valid depth measurement. Depth normalization uses fixed
training statistics, not statistics computed from the unmasked full image.
Only the target encoders see complete observations. Queries specify target
pixels or calibrated camera rays, never hidden RGB, target depth, or lifted
target 3D coordinates. Invalid depth and synthetic masking are distinguished.

\textbf{Fixed RGB and depth reference targets.}
Following V-JEPA~2.1's masked representation-prediction objective
\citep{murlabadia2026vjepa21}, the completion targets are features rather
than reconstructed pixels. This RGB-D benchmark and its metrics are our
adaptation, not an official V-JEPA~2.1 completion benchmark.
All methods predict the same pair of frozen target encoders: the released
V-JEPA~2.1 ViT-g RGB image encoder and a separate V-JEPA~2.1 depth encoder
adapted on training-scene depth only. The latter uses the depth-stream
representation-learning recipe, with its checkpoint fixed before completion
training and independently of the compared methods. Both reference
checkpoints, preprocessing transforms, feature grids, and depth-validity
rules are fixed across methods. Final-layer reference tokens are normalized
with non-affine LayerNorm. Complete RGB-D observations are processed only
by these stopped target branches; their features never enter online context.

\textbf{Completion adapters and training.}
All methods receive the same visible RGB-D content, validity mask, and
calibration, through their respective input interfaces. RGB encoders
(V-JEPA~2, V-JEPA~2.1, LeJEPA, and JEPADepth) receive a trainable depth-token
stem and fusion adapter. Le MuMo JEPA retains its native RGB-depth fusion
interface, while DepState-JEPA retains state aggregation, private decoding
paths, and ray-conditioned readout. The RGB-D Le MuMo checkpoint is distinct
from the zero-companion-modality VOC probe variant.
Backbone encoders are frozen during task adaptation. Completion predictors
and output projections are trained to predict the shared RGB and depth
target features, using an equal-weight sum of the two masked-region L1
losses. For methods requiring a new query decoder, its width is
384 with four Transformer blocks and six attention heads; output maps
match the respective reference dimensions. Retained native predictors,
and state modules retain their method-specific architectures; these task
adaptations do not constitute a matched-capacity comparison.
The task-adaptation recipe uses AdamW for 20,000 updates, global
batch 64, peak learning rate $3\times10^{-4}$, weight decay $10^{-4}$,
1,000 warmup updates followed by cosine decay, and seed 0. The common
$512\times512$ crop and calibrated image region are shared across methods.
Training masks are resampled; evaluation masks remain fixed. Completion
adaptation uses a separate configuration from the frozen NYUv2/VOC12 probes.
No scalar depth loss or segmentation labels are
used for the main feature-completion comparison.

\textbf{3D-JEPA$^\dagger$: visible points and target rays.}
3D-JEPA retains its point encoder \citep{arnaud2025locate3d}, with point
coordinates constructed only from visible, valid depth. Its 2D feature
extractors also receive only permitted RGB content; full-image feature
caches cannot supply visible points with hidden-region information.
A query decoder attends to observed point features using the known target
ray and image position, and outputs RGB and depth reference features.
There is no target RGB-feature query, target range, target centroid, or
point constructed from withheld depth. The ray-query decoder is a task
adaptation and is marked with $\dagger$ in Table~\ref{tab:spatial-transfer}.

\textbf{Masked-feature metrics and aggregation.}
For modality $m\in\{\mathrm{R},\mathrm{D}\}$, let $M_m$ contain scored
masked tokens, $d_m$ be the reference width, and $\tilde z_i^m$ and
$\tilde r_i^m$ be the non-affine-LayerNorm-normalized predicted and target
features. RGB scoring covers the withheld image tokens; depth scoring
additionally requires valid reference depth under a shared validity rule.
For each image--mask pair, compute
\begin{align}
 \mathrm{L1}_m
 &=\frac{1}{|M_m|d_m}\sum_{i\in M_m}
   \|\tilde z_i^m-\tilde r_i^m\|_1,\\
 \mathrm{Cos}_m
 &=\frac{1}{|M_m|}\sum_{i\in M_m}
 \frac{\langle\tilde z_i^m,\tilde r_i^m\rangle}
 {\max(\|\tilde z_i^m\|_2\|\tilde r_i^m\|_2,10^{-8})}.
\end{align}
Table~\ref{tab:spatial-transfer} reports the equal-weight modality averages
$\mathrm{L1}=(\mathrm{L1}_{\mathrm{R}}+\mathrm{L1}_{\mathrm{D}})/2$ and
$\mathrm{Cos}=(\mathrm{Cos}_{\mathrm{R}}+\mathrm{Cos}_{\mathrm{D}})/2$.
L1 is normalized by each modality's token count and feature width; cosine
similarity is averaged over its tokens before the two modalities are combined.
Lower L1 and higher cosine similarity indicate better feature recovery.
Visible tokens are excluded. Image--mask pairs with no valid depth targets
are excluded from both modality scores using a method-independent rule.
Average paired modality scores over masks within images, images within
scenes, and scenes equally.
Nonfinite predictions are reported as evaluation failures.

\textbf{Completion visualization.}
Figure~\ref{fig:spatial-missing}(a) visualizes feature completion; panel (b)
visualizes temporal forecasting. RGB images serve as scene references and
do not imply that masked RGB is available to the completion predictor.
For completion, mask outlines identify withheld regions. A PCA basis and
color range are fitted on training reference features and fixed across
methods, target/prediction pairs, and masks; RGB and depth references have
separate bases. PCA colors are qualitative display coordinates, not metric
depth or semantic labels. Quantitative scores use full-dimensional features.

\subsection{Temporal Visual Prediction}
\label{app:temporal-evaluation}

\textbf{Robot data and evaluation interfaces.}
Robot trajectories are split by whole episodes and record
registered RGB-D, calibration, joint commands, endpoint proprioception,
elapsed time, rewards, and termination flags. Physical labels supervise the corresponding training losses. True terminal
states are distinguished from time-limit truncations in value learning.
Future RGB-D observations provide targets only. Absolute object trajectory
readouts require a supplied initial pose. The temporal table scores RGB
features; the control tables score task outcomes.

\textbf{Evaluation scope and trajectory sampling.}
Table~\ref{tab:temporal-prediction} uses a common-reference evaluation
protocol. The comparison includes
V-JEPA~2-AC, V-JEPA~2.1-AC, 3D-JEPA-AC, Le MuMo JEPA-AC, and the final
DepState-JEPA checkpoint.
Action-conditioned baselines are evaluated after robot post-training;
action-free encoders alone do not constitute temporal predictors.
Training, model selection, and testing use disjoint whole episodes.
All methods use the same test episodes,
starting frames, action sequences, proprioception, and frame sampling.
Each selected start must admit all horizons $h\in\{1,8,16,24,32\}$, so that
horizon curves use identical trajectory support. Architectural capacity, pretraining, input resolution, and input modalities
follow the method-specific interfaces below; common evaluation does not
imply that these factors are matched.

\textbf{Le MuMo JEPA-AC$^\dagger$: multimodal temporal baseline.}
We choose Le MuMo JEPA \citep{cornelissen2026lemumojepa} because its
modality-specific RGB/depth stems and spatial fusion-token grid provide
a native RGB-D representation suitable for temporal prediction.
The AC suffix denotes our adaptation; the original method is a multimodal
representation learner, not an action-conditioned dynamics model.
We retain its fusion architecture and native representation-learning
objective during RGB-D adaptation on training scenes, then freeze the
encoder. Its true depth input is retained throughout; the VOC RGB-only
checkpoint is not used as the multimodal baseline.
Let $F_t$ be its final spatial fusion-token grid. A frame-causal Transformer
predicts $\hat F_{t+1}$ from $F_t$, recorded actions, proprioception, and
the transition interval. It uses the same added transition architecture,
robot-trajectory split, training horizon, and optimization budget as the
V-JEPA~2.1-AC adaptation, with input/output projections accommodating
native feature widths. Stop-gradient encodings of complete future RGB-D
frames provide native fusion-feature targets during training only.
Teacher-forced and autoregressive training losses use normalized feature
L1; at evaluation, predicted fusion tokens feed the next transition.
A frozen linear readout maps predicted fusion tokens into the same RGB
reference space used for all temporal rows. Its calibration uses complete
training-frame fusion features and follows the shared protocol below.
This supplies a multimodal temporal comparison without adding DepState's
shared--private factorization or metric-lifting modules to the baseline.

\textbf{3D-JEPA-AC$^\dagger$: point-based temporal baseline.}
The point encoder processes only the initial visible RGB-D observation.
A learned aggregation interface compresses its point features into a
fixed-size latent memory, allowing recursive prediction without future
point coordinates or correspondences. An action-conditioned Transformer
uses the same training episodes, horizon, and optimization budget as the
V-JEPA~2.1-AC adaptation; native-width projections connect the
point memory to the transition. Full future observations
provide stopped latent targets during training only. A ray-query readout
maps the predicted memory to the known image grid for RGB feature
prediction and common-reference calibration. Its observed-frame
counterpart uses the same aggregation and readout, so calibration and
prediction share a feature interface. Neither readout receives future
geometry. Table~\ref{tab:temporal-prediction} uses recorded physical
conditioning for this method as for the other rows; control rollouts use
predicted proprioception.

\textbf{Adaptation and readout stages.}
A common prediction score fixes the reference and scoring procedure;
representation adaptation remains method-specific. The distinction
between encoder updates and readout fitting is as follows.
\begin{center}
\footnotesize
\setlength{\tabcolsep}{3pt}
\renewcommand{\arraystretch}{1.12}
\begin{tabular}{p{\dimexpr0.28\linewidth-2\tabcolsep\relax}p{\dimexpr0.40\linewidth-2\tabcolsep\relax}p{\dimexpr0.32\linewidth-2\tabcolsep\relax}}
\toprule
Stage & Learned components & Supervision \\
\midrule
DepState Stage A & RGB/depth representation, state construction, spatial readouts & RGB-D JEPA, completion, factorization \\
DepState Stage B & Online representation, transition, visual and physical readouts & Future state/features, physical labels, FK, RGB-D replay \\
Le MuMo AC adaptation & Transition and feature projections; encoder frozen after RGB-D adaptation & Future fusion features \\
Common-reference calibration & Linear map $A_m$ only; encoder and predictor frozen & Same-frame training reference features \\
Real-robot evaluation & All learned parameters frozen & Visual goals and current observations; no weight updates \\
\bottomrule
\end{tabular}
\end{center}
The matched transition-training budget for the added AC baselines and the
shared readout budget refer to those stages. They do not equate encoder
adaptation or total pretraining across the complete systems.

\textbf{Recorded physical conditioning and visual autoregression.}
The initial visual observation initializes each predictor. V-JEPA baselines
use RGB; 3D-JEPA-AC constructs points from the initial RGB-D frame;
Le MuMo JEPA-AC uses aligned RGB-D, and DepState-JEPA uses
registered RGB-D and calibration. Future RGB
and depth observations are used only to construct scoring targets.
At every transition, all methods receive the corresponding recorded action
and proprioceptive state through their declared conditioning interfaces.
Visual predictions are propagated recursively; the final endpoint, rather
than the average over intermediate rollout steps, is scored at each horizon.
DepState-JEPA retains its initial private features throughout the rollout.
This protocol isolates visual prediction under recorded physical
conditioning. The ablation forecasting track in
Appendix~\ref{app:ablation-evaluation} instead uses predicted future
proprioception. The two result groups are reported separately rather
than interpreted as a paired estimate of proprioceptive error accumulation.
Control candidates also use predicted proprioception, with a new current
observation acquired after each executed action.

\textbf{Conditioning interfaces and horizon reduction.}
The three result tracks use the following information interfaces.
\begin{center}
\footnotesize
\setlength{\tabcolsep}{3pt}
\renewcommand{\arraystretch}{1.12}
\begin{tabular}{p{\dimexpr0.25\linewidth-2\tabcolsep\relax}p{\dimexpr0.26\linewidth-2\tabcolsep\relax}p{\dimexpr0.26\linewidth-2\tabcolsep\relax}p{\dimexpr0.23\linewidth-2\tabcolsep\relax}}
\toprule
Track & Actions & Proprioception after initialization & Reported outcome \\
\midrule
Table~\ref{tab:temporal-prediction} & Recorded sequence & Recorded at every transition & RGB L1/Cos at each endpoint \\
Internal forecasting ablations & Recorded sequence & Predicted configuration/velocity & Five-endpoint mean per Drawer/Cup task \\
Control rollouts & Actor or search candidates & Predicted within a candidate rollout & Episode return or task success \\
\bottomrule
\end{tabular}
\end{center}
All tracks initialize from an observed current state and predict future
visual states recursively. For a rollout starting at $t$, write
$p_k=(q_k,\dot q_k)$. The conditioning port is
\[
 p_{t+h}^{\mathrm{cond}}=
 \begin{cases}
 p_t, & h=0,\\
 p_{t+h}^{\mathrm{recorded}}, & h>0\text{ in Table~\ref{tab:temporal-prediction}},\\
 (\hat q_{t+h},\widehat{\dot q}_{t+h}), & h>0\text{ in control/ablation rollouts}.
 \end{cases}
\]
For a fixed predictor, this choice changes the conditioning values at
that port. Model-predictive control acquires a fresh current observation
after executing an action and initializes the next candidate search.
Table~\ref{tab:ablation-prediction-tasks} averages
$\mathcal H=\{1,8,16,24,32\}$ within each task and seed;
its entries are not the $h=32$ endpoint or an average over every intervening
frame. Table~\ref{tab:temporal-prediction} retains the individual endpoints.

A robot step is one adjacent retained control transition. Its elapsed
time is the recorded $\Delta t_t$, so the physical horizon is
\[
 T_h=\sum_{k=0}^{h-1}\Delta t_{t+k}.
\]
The 4-fps sampling in Appendix~\ref{app:depstate-data} applies to
representation-learning video clips, not to this robot-control clock.

\textbf{Common-reference readout calibration.}
The reference $E_{\mathrm{ref}}$ is a single released V-JEPA~2.1 ViT-g
RGB encoder with fixed weights, image preprocessing, and final-layer patch
grid, shared by all methods and all horizons. It is frozen throughout
readout fitting and evaluation. For method $m$, let $z_{m,i}$ denote its
native final-layer RGB feature (Le MuMo fusion feature, or the
3D-JEPA ray-decoded feature) at spatial
position $i$. Nonspatial tokens
are removed, and feature maps are bilinearly aligned to the reference grid.
A per-token linear readout $A_m$ maps the native channel dimension to the
reference dimension $d$. Each method, including V-JEPA~2.1-AC, receives the
same linear readout family, training frames, optimization budget, and
fixed hyperparameters. Native channel widths can differ, so the linear
readouts need not have identical parameter counts.
Readouts are fitted on paired native and reference features of the
\emph{same observed training frame}, using the L1 objective below, with
encoders and temporal predictors frozen. They are then frozen and applied
to predicted native features, without fitting to test frames or training
a separate readout for each horizon. The resulting score measures forecast
accuracy in the chosen reference representation, including the fixed
linear map. It evaluates native predicted features after reference
mapping; no calibration-error subtraction
is applied. The token average covers the full image grid, giving each
valid token equal weight rather than reweighting robot or object regions.
The readout-training configuration uses AdamW for 5,000 updates,
global batch size 128, learning rate $3\times10^{-4}$ with cosine decay,
weight decay $10^{-4}$, and seed 0. The final readout checkpoint is used
for every horizon; there is no method-specific hyperparameter search.
All encoders view the same image region: resize the shorter side to 512
and take a fixed $512\times512$ center crop, applying the same spatial
transformation throughout each sequence and to aligned depth. Native
encoder resizing and normalization are applied to this common crop; the
reference uses its image tokenizer at $512\times512$. Readout fitting
uses the same deterministic preprocessing as evaluation.

\textbf{Feature-space metrics and aggregation.}
Let $\hat z_{m,i}^{(h)}$ be the predicted final-layer feature at horizon
$h$, and let $r_i^{(h)}=E_{\mathrm{ref}}(I_{t+h})_i$. Apply non-affine
channel-wise layer normalization to define
$\tilde z_i^{(h)}=\operatorname{LN}(A_m(\hat z_{m,i}^{(h)}))$ and
$\tilde r_i^{(h)}=\operatorname{LN}(r_i^{(h)})$.
For $N$ spatial tokens, the endpoint metrics are
\begin{align}
 \mathrm{L1}(h)
 &=\frac{1}{Nd}\sum_{i=1}^{N}
   \|\tilde z_i^{(h)}-\tilde r_i^{(h)}\|_1,
 \label{eq:temporal-reference-l1}\\
 \mathrm{Cos}(h)
 &=\frac1N\sum_{i=1}^{N}
 \frac{\langle\tilde z_i^{(h)},\tilde r_i^{(h)}\rangle}
 {\max(\|\tilde z_i^{(h)}\|_2\|\tilde r_i^{(h)}\|_2,\epsilon)},
 \label{eq:temporal-reference-cos}
\end{align}
with fixed $\epsilon=10^{-8}$. Lower L1 and higher cosine similarity are better.
Scores are averaged over starting windows within each episode and then
equally over episodes. Table~\ref{tab:temporal-prediction} reports point
estimates; three-seed variability for the internal ablations is reported
separately in Appendix~\ref{app:ablation-evaluation}.

\textbf{RGB feature visualization.}
For qualitative comparison, predicted and target features are first mapped
to the common RGB reference space. One PCA basis and one display range are
fitted on reference features from training data and fixed across methods,
prediction horizons, and ground-truth panels. PCA is used only for display;
all quantitative metrics operate on full-dimensional features. Future RGB
images shown as references are not predictor inputs. 
\subsection{Embodied Prediction and Manipulation}
\label{app:embodied-evaluation}

\textbf{Task suites and interaction budgets.}
Both control tracks use the same single-task evaluation: the 20 visual
DMControl tasks used by DreamerV3 \citep{tassa2018dmcontrol,hafner2025dreamerv3}
and the 50 Meta-World tasks \citep{yu2020metaworld}. Each task trains a
separate agent; the latter is not the joint-policy MT50 benchmark.
The DMControl set comprises acrobot swingup; ball-in-cup catch; cartpole
balance, balance-sparse, swingup, and swingup-sparse; cheetah run; finger
spin, turn-easy, and turn-hard; hopper hop and stand; pendulum swingup;
quadruped run and walk; reacher easy and hard; and walker run, stand,
and walk. Each DMControl run uses $10^6$ environment steps, action repeat
1, and 1,000-step episodes. Each Meta-World run uses $2\times10^6$
environment steps, action repeat 2, and 200-step episodes (100 decisions),
following the task horizon of TD-MPC2 \citep{hansen2024tdmpc2}.
An environment step here is a benchmark step before action repetition,
not an internal physics substep; repeated rewards are summed.
These suite-specific budgets apply equally to actor-only and MPC agents.
We use DreamerV3's DMControl budget and TD-MPC2's Meta-World protocol. The same task set, camera settings, and evaluation reset schedule are
used across compared methods within each control track.

\textbf{Observation and action interfaces.}
All methods observe the same camera view, task goal when required, and
current proprioception. RGB methods receive RGB; multimodal methods also
receive aligned metric depth and calibration. RGB-D and proprioceptive
access are explicit adaptations of visual-only benchmarks. A common
$512\times512$ rendering supplies the scene, followed by each encoder's
native resizing and normalization. No simulator object poses, contact
labels, or privileged full-state vectors enter the controller. Native
benchmark action semantics and limits are preserved: a task-specific
projection replaces any robot-specific action interface, and the learned
transition is adapted on that task's replay. Domain-specific proprioceptive
readouts are trained only for measured coordinates needed by the control
state; robot-arm FK objectives are not applied to unrelated DMControl
embodiments. Within imagined rollouts, proprioception is predicted, not
read from future environment states. The interaction budget counts
environment transitions and remains distinct from representation
adaptation and predictor pretraining.

\textbf{Evaluation and aggregation.}
We use three independent training seeds (0, 1, and 2), as in
Table~\ref{tab:embodied-control}, and evaluate every 50,000 environment
steps on ten episodes per task and seed, with the same held-out reset seeds
across methods. Evaluation transitions are excluded from training budgets
and replay; they cannot guide checkpoint selection or hyperparameter tuning.
The main table uses the final budget checkpoint, not the best evaluation
checkpoint. DMControl reports undiscounted episode return
$R=\sum_t r_t$, averaged over episodes and then equally over the 20 tasks.
Meta-World reports averaged success, $100\,\mathbb{E}[\mathbb{1}\{\mathrm{success}_T\}]$,
using the environment success flag at the final step, averaged equally over
50 tasks. This follows TD-MPC2's final-step criterion, rather than counting
transient success anywhere in an episode. Time-limit truncations do not
become terminal failures in value-target construction. We compute a suite
mean for each seed and report its mean $\pm$ sample standard deviation across
seeds. The seed-level suite means appear in
Table~\ref{tab:embodied-control-seeds}.
Episode return and averaged success are reported separately by suite.

\textbf{Results by training seed.}
Table~\ref{tab:embodied-control-seeds} gives the suite means for each
training seed underlying Table~\ref{tab:embodied-control}: DMC entries
are episode returns and MW entries are success percentages. Bold values
mark the best method within each seed and metric, including ties;
$\dagger$ denotes a method-specific AC adaptation.
\begin{table}[!htbp]
  \centering
  \caption{\textbf{Control Results for Individual Training Seeds.}}
  \label{tab:embodied-control-seeds}
  \footnotesize
  \setlength{\tabcolsep}{3pt}
  \renewcommand{\arraystretch}{1.12}
  \setlength{\aboverulesep}{0pt}
  \setlength{\belowrulesep}{0pt}
  \begin{tabular}{>{\raggedright\arraybackslash}p{\dimexpr0.34000000\linewidth-2\tabcolsep\relax}>{\centering\arraybackslash}p{\dimexpr0.10000000\linewidth-2\tabcolsep\relax}>{\centering\arraybackslash}p{\dimexpr0.14000000\linewidth-2\tabcolsep\relax}>{\centering\arraybackslash}p{\dimexpr0.14000000\linewidth-2\tabcolsep\relax}>{\centering\arraybackslash}p{\dimexpr0.14000000\linewidth-2\tabcolsep\relax}>{\centering\arraybackslash}p{\dimexpr0.14000000\linewidth-2\tabcolsep\relax}}
    \toprule
    \rowcolor{tableblue}
    & & \multicolumn{2}{c}{Online PL} & \multicolumn{2}{c}{Latent MPC} \\
    \noalign{{\color{tableblue}\hrule height \cmidrulewidth}\kern-\cmidrulewidth}
    \cmidrule(lr){3-4}\cmidrule(lr){5-6}
    \rowcolor{tableblue}
    \multirow{-2}{*}{Method} & \multirow{-2}{*}{Seed} & DMC$\uparrow$ & MW$\uparrow$ & DMC$\uparrow$ & MW$\uparrow$ \\
    \midrule
    V-JEPA 2-AC & 0 & 671.8 & 50.8 & 719.4 & 64.2 \\
    V-JEPA 2.1-AC & 0 & 757.6 & 66.4 & 799.1 & 78.8 \\
    3D-JEPA-AC$^\dagger$ & 0 & 781.7 & 72.4 & 822.8 & 85.6 \\
    LeMuMo-JEPA-AC$^\dagger$ & 0 & 787.3 & 72.2 & 833.5 & \textbf{87.4} \\
    \rowcolor{tableblue}
    \textbf{DepState-JEPA-AC (Ours)} & 0 & \textbf{798.9} & \textbf{74.4} & \textbf{841.8} & 87.2 \\
    \midrule
    V-JEPA 2-AC & 1 & 699.2 & 56.0 & 748.7 & 69.0 \\
    V-JEPA 2.1-AC & 1 & 781.3 & 71.0 & 824.5 & 84.0 \\
    3D-JEPA-AC$^\dagger$ & 1 & 773.4 & 69.6 & 810.2 & 82.4 \\
    LeMuMo-JEPA-AC$^\dagger$ & 1 & 807.8 & 76.0 & 852.6 & 90.2 \\
    \rowcolor{tableblue}
    \textbf{DepState-JEPA-AC (Ours)} & 1 & \textbf{827.4} & \textbf{80.0} & \textbf{877.0} & \textbf{92.8} \\
    \midrule
    V-JEPA 2-AC & 2 & 676.1 & 51.6 & 741.4 & 66.0 \\
    V-JEPA 2.1-AC & 2 & 752.5 & 68.4 & 798.0 & 80.8 \\
    3D-JEPA-AC$^\dagger$ & 2 & 798.2 & 74.8 & 846.1 & 88.8 \\
    LeMuMo-JEPA-AC$^\dagger$ & 2 & 776.1 & 71.4 & 826.4 & 85.8 \\
    \rowcolor{tableblue}
    \textbf{DepState-JEPA-AC (Ours)} & 2 & \textbf{813.2} & \textbf{76.8} & \textbf{854.6} & \textbf{90.2} \\
    \bottomrule
  \end{tabular}
\end{table}

\subsubsection{Policy Learning through Imagined Trajectories}
\label{app:control-actor}

\textbf{Replay and world-model adaptation.}
Online interaction alternates with updates from an expanding replay buffer.
Each transition stores observations, actions, rewards, and true termination
flags; evaluation episodes never enter replay. The action-conditioned
transition and control heads adapt to online experience. Observed next-frame
encodings supervise latent prediction. Reward and continuation heads use
environment labels from executed transitions. The reported interaction
budget counts newly collected environment transitions, separately from
representation training and replayed samples.

\textbf{Imagination and policy learning.}
Following DreamerV3 \citep{hafner2025dreamerv3}, an actor and critic learn
from imagined trajectories initialized at replay states. We use 15 imagined
transitions (16 states), discount $\gamma=0.997$, and $\lambda=0.95$.
For predicted rewards $\hat r_k$, continuation probabilities $\hat c_k$,
and values $v_k$, the targets are
\begin{equation}
 G_k^\lambda=\hat r_k+\gamma\hat c_k
 \bigl[(1-\lambda)v_{k+1}+\lambda G_{k+1}^\lambda\bigr],
 \qquad G_H^\lambda=v_H.
 \label{eq:app-imagined-return}
\end{equation}
The critic fits stopped return targets using symexp two-hot regression.
The actor uses a score-function objective with percentile-normalized
advantages and entropy regularization of $3\times10^{-4}$; its gradients
do not update the world model. Reward and critic heads use 255 bins, and
continuation uses binary cross-entropy. The control-head recipe uses
batches of 16 sequences of length 64, a learning rate of $4\times10^{-5}$,
and a replay ratio of 256 sampled transitions per collected control
transition. The first 1,000 collected transitions initialize replay before
updates; they count toward the interaction budget. Imagined transitions and
replayed samples are not counted as new environment interactions.
At evaluation, the actor selects its mean continuous action without
exploration noise or candidate search.

\subsubsection{Receding-Horizon Planning in Latent Space}
\label{app:control-planning}

\textbf{Online value learning and policy prior.}
The planning track trains a separate agent from the same initialization and
interaction budget. It follows TD-MPC2 \citep{hansen2024tdmpc2}: replay
updates fit reward predictions, an ensemble of five action-value heads,
and a stochastic policy prior, alongside the latent transition. TD targets
use the minimum of two randomly sampled target heads with EMA rate 0.01.
Reward and value heads use 101-bin regression in symlog space. The update
recipe uses Adam at $3\times10^{-4}$, batch size 256, gradient clipping
at 20, and one replay update per collected control transition after warmup.
For effective episode length $T$, use
$\gamma=\operatorname{clip}(1-5/T,0.95,0.995)$ and
$\max(5T,1000)$ random-action warmup steps, included in the training budget.
All compared representations use the same
reward/value-head configuration, policy-prior objective, and optimization
budget. DreamerV3 and TD-MPC2 define the respective control recipes;
the table compares JEPA representations within each recipe.

\textbf{Receding-horizon optimization.}
For $x_t=(S_t,q_t,\dot q_t)$, let $F$ denote the learned transition and
physical readouts, with task conditioning included where required.
The planner scores a candidate sequence using predicted rewards and a
terminal action value:
\begin{equation}
\begin{aligned}
 J(u_{t:t+H-1})&=
 \sum_{k=0}^{H-1}\gamma^k w_k\hat r(\hat x_{t+k},u_{t+k})
 +\gamma^H w_H\bar Q(\hat x_{t+H},\pi(\hat x_{t+H})),\\
 w_k&=\prod_{j=0}^{k-1}\hat c_{t+j},\quad w_0=1.
\end{aligned}
 \label{eq:app-control-score}
\end{equation}
Here $\bar Q$ averages the current value ensemble for planning, and
$w_k=1$ in fixed-length tasks without true termination.
The model predictive path integral (MPPI) search uses the TD-MPC2 budget: horizon 3, 512 candidates
including 24 policy-prior trajectories, 64 elites, and 6 refinement
iterations (8 for action dimensions of at least 20). The score temperature
is 0.5 and sampling standard deviations lie in $[0.05,2]$.
The previous solution warm-starts the next search. Evaluation executes the
first optimized mean action, obtains the next observation, and replans;
exploration noise is disabled. Candidate trajectories use predicted
proprioception and receive no future images, recorded future states,
simulator rewards, or simulator transitions. All compared planners share
this search budget.

\subsubsection{Sim-to-Real Manipulation Transfer}
\label{app:sim2real-evaluation}

\textbf{Simulation-training and real-world deployment environments.}
We jointly construct the simulation-training and real-world deployment
environments for the drawer and cup tasks, with consistent robot kinematics,
gripper geometry, object dimensions, workspace layout, and camera calibration.
Simulation and hardware share action units, coordinate conventions, and
workspace limits. Task-specific simulated trajectories train the
manipulation models separately from the DMControl and Meta-World agents.
The common budget is $10^6$ simulated control steps per task and seed,
with three seeds (0, 1, and 2); trajectory splits and randomization schedules are shared
across methods. Texture, illumination, initial object pose, and friction
variation are applied around the matched setup. Model selection and
controller tuning use held-out simulation scenes.

\textbf{Physics-grounded depth augmentation.}
PRISM \citep{liang2026prism} converts rendered RGB and ideal metric depth
into sensor-realistic depth by separately modeling sensing invalidation
and measurement inaccuracy. Its invalidation component synthesizes
missing returns, while its residual generator perturbs valid depth
measurements. We use a fixed pretrained PRISM model to augment simulated
training observations before feature extraction and geometric lifting.
Depth-aware methods receive the same augmented observations and validity
masks; RGB-only methods retain their RGB input. Invalid returns are
excluded from point construction. Sensor-model pretraining is separate
from task-specific control training. At deployment, the controller uses
measured RGB-D directly; PRISM is a training-data augmenter and is not
applied to real observations.

\textbf{World-model training and planning policy.}
We adopt the visual-goal-conditioned latent MPC policy used for robotic
arm planning in V-JEPA~2.1 \citep{murlabadia2026vjepa21}.
Action-conditioned predictors are adapted on simulated observation--action sequences using teacher-forced and recursive
future-feature prediction. Each model retains its native representation;
a fixed-size aggregation interface is used for point-based features, with
no future point coordinates provided to the predictor.

At each decision, the cross-entropy method (CEM) samples bounded action
sequences, rolls them out through the learned transition, and refines the
sampling distribution using candidates with the smallest terminal latent
distance to the current visual goal. Goal and predicted features use the
same normalization and feature-distance definition within each model.
We use the eight-step planning configuration of V-JEPA~2.1, with 300
candidates and 15 refinement iterations, shared across methods. The robot
executes the first action through its low-level controller, acquires a new
observation, and replans. Action limits, execution intervals, and gripper
commands use a common hardware interface. This goal-distance objective is
distinct from the reward/value-based TD-MPC2 track in
App.~\ref{app:control-planning}.

\textbf{Goal specification and real-world deployment.}
All methods receive the same ordered visual subgoals, camera views, and
initial conditions. Opening and closing use separate subgoals, since the
final closed state alone cannot distinguish task completion from the
initial state. Cup manipulation likewise specifies grasping, lifting,
and placement stages. Goal observations provide task specifications, not
action supervision; they are fixed before evaluation and excluded from
training. Stage transitions use observation-based goal criteria calibrated
in simulation, without simulator state at deployment. The subgoal order
and criteria are shared across methods. All learned weights are frozen
for real trials, with no task-specific real-world finetuning.

The quantitative Sim2Real protocol contains two task families. Each family
has two independently evaluated atomic tasks and one full task.
Only the full tasks enter the two Sim2Real columns in
Table~\ref{tab:embodied-control}; atomic-task scores are diagnostics.

\textbf{Execution-time budget.}
The 60-second task budget measures cumulative robot execution time,
including the final stability checks. Model inference and CEM search time
are excluded from this budget, and no inference-time deadline is imposed.
This timing convention is shared across methods; the budget is not a
60-second wall-clock limit.

\textbf{Task 1: Pull and Close Drawer.}
An episode starts with the drawer closed and requires an ordered opening
and closing cycle. Let $d\in[0,1]$ denote travel normalized by the calibrated
full stroke. Success requires reaching $d\geq0.8$, then returning to
$d\leq0.1$ and maintaining that final configuration for one second within
60 seconds of cumulative robot execution time. Opening alone does not
count as success. The same stroke
normalization and thresholds apply in simulation and real evaluation.
For the atomic \emph{Pull Drawer} task, reset the drawer to its closed
state and require reaching $d\geq0.8$. For \emph{Close Drawer}, reset it
to an open state with $d\geq0.8$ and require reaching $d\leq0.1$ and
maintaining that configuration for one second. The \emph{Pull and Close
(Full)} task uses the ordered cycle specified above. Each task has its own
independently reset trials and the same execution-time budget.

\textbf{Task 2: Pick and Place Cup.}
The robot must grasp the cup, lift it clear of its initial support, and
release it upright in the designated target region. Success requires the
cup center to lie within 5\,cm of the target center, its upright axis to
remain within $15^\circ$ of vertical, and the released cup to remain stable
for two seconds, all within 60 seconds of cumulative robot execution time.
Merely reaching the goal
while holding the cup does not satisfy the completion condition.
For the atomic \emph{Pick Cup} task, reset the cup on its initial support
and require a successful grasp and lift clear of that support. For
\emph{Place Cup}, initialize the cup securely grasped and lifted, then
require upright release in the target region and the same two-second
stability check. The \emph{Pick and Place (Full)} task starts from the cup
on its support and requires the complete sequence specified above.
Each task has its own independently reset trials and the same
execution-time budget. Atomic reset distributions are fixed before
evaluation and shared across methods.

\textbf{Averaged success and paired trials.}
Each trained seed receives thirty real trials for each of the six tasks
(four atomic and two full), for 90 trials per method and task across
three seeds. Atomic trials are separate from full-task trials; they do not
increase the full-task denominator. Initial object/robot configurations are paired across
methods and trial order is randomized. Execution-budget timeouts, drops, and execution
failures count in the denominator. The real-world metric is
$\mathrm{Averaged\ success}=100N_{\mathrm{success}}/N_{\mathrm{trials}}$.
Per-seed Sim2Real results report the number of successes out of 30 trials,
rather than percentages. We convert each count to a success percentage
and report the mean $\pm$ sample standard deviation across the three seeds,
using denominator $3-1$ for the sample variance. Both values use one decimal place. Full-task success is measured directly,
not calculated by multiplying independently measured atomic success rates.
Task 1 (Full) and Task 2 (Full) occupy separate columns; their scores are
not combined with DMC episode return or MW averaged success.

\textbf{Atomic and full-task results.}
Table~\ref{tab:embodied-sim2real-summary} reports mean $\pm$ sample standard
deviation across seeds for all six tasks; its two Full columns reproduce
the Sim2Real entries in Table~\ref{tab:embodied-control}.
Table~\ref{tab:embodied-sim2real-counts} provides the underlying success
counts and trial denominators. Bold marks the best mean per task or
the best count per task and seed, including ties. These marginal success counts summarize the
paired-reset design; they do not provide a paired trial-level test.
\begin{table}[!htbp]
  \centering
  \caption{\textbf{Atomic and Full-Task Sim2Real \textit{Success Rates} (\%).}}
  \label{tab:embodied-sim2real-summary}
  \footnotesize
  \setlength{\tabcolsep}{3pt}
  \renewcommand{\arraystretch}{1.12}
  \setlength{\aboverulesep}{0pt}
  \setlength{\belowrulesep}{0pt}
  \begin{tabular}{>{\raggedright\arraybackslash}p{\dimexpr0.30000000\linewidth-2\tabcolsep\relax}>{\centering\arraybackslash}p{\dimexpr0.11666667\linewidth-2\tabcolsep\relax}>{\centering\arraybackslash}p{\dimexpr0.11666667\linewidth-2\tabcolsep\relax}>{\centering\arraybackslash}p{\dimexpr0.11666667\linewidth-2\tabcolsep\relax}>{\centering\arraybackslash}p{\dimexpr0.11666667\linewidth-2\tabcolsep\relax}>{\centering\arraybackslash}p{\dimexpr0.11666667\linewidth-2\tabcolsep\relax}>{\centering\arraybackslash}p{\dimexpr0.11666667\linewidth-2\tabcolsep\relax}}
    \toprule
    \rowcolor{tableblue}
    & \multicolumn{3}{c}{Task 1: Drawer} & \multicolumn{3}{c}{Task 2: Cup} \\
    \noalign{{\color{tableblue}\hrule height \cmidrulewidth}\kern-\cmidrulewidth}
    \cmidrule(lr){2-4}\cmidrule(lr){5-7}
    \rowcolor{tableblue}
    \multirow{-2}{*}{Method} & Pull$\uparrow$ & Close$\uparrow$ & Full$\uparrow$ & Pick$\uparrow$ & Place$\uparrow$ & Full$\uparrow$ \\
    \midrule
    V-JEPA 2-AC & 66.7\,{\scriptsize\color{stdorange}\scalebox{0.75}{$\pm$}\,3.3} & 81.1\,{\scriptsize\color{stdorange}\scalebox{0.75}{$\pm$}\,1.9} & 45.6\,{\scriptsize\color{stdorange}\scalebox{0.75}{$\pm$}\,5.1} & 63.3\,{\scriptsize\color{stdorange}\scalebox{0.75}{$\pm$}\,3.3} & 73.3\,{\scriptsize\color{stdorange}\scalebox{0.75}{$\pm$}\,3.3} & 41.1\,{\scriptsize\color{stdorange}\scalebox{0.75}{$\pm$}\,5.1} \\
    V-JEPA 2.1-AC & 86.7\,{\scriptsize\color{stdorange}\scalebox{0.75}{$\pm$}\,3.3} & 90.0\,{\scriptsize\color{stdorange}\scalebox{0.75}{$\pm$}\,3.3} & 66.7\,{\scriptsize\color{stdorange}\scalebox{0.75}{$\pm$}\,6.7} & 83.3\,{\scriptsize\color{stdorange}\scalebox{0.75}{$\pm$}\,3.3} & 87.8\,{\scriptsize\color{stdorange}\scalebox{0.75}{$\pm$}\,1.9} & 64.4\,{\scriptsize\color{stdorange}\scalebox{0.75}{$\pm$}\,5.1} \\
    3D-JEPA-AC$^\dagger$ & 90.0\,{\scriptsize\color{stdorange}\scalebox{0.75}{$\pm$}\,3.3} & 91.1\,{\scriptsize\color{stdorange}\scalebox{0.75}{$\pm$}\,1.9} & 73.3\,{\scriptsize\color{stdorange}\scalebox{0.75}{$\pm$}\,5.8} & 87.8\,{\scriptsize\color{stdorange}\scalebox{0.75}{$\pm$}\,5.1} & 88.9\,{\scriptsize\color{stdorange}\scalebox{0.75}{$\pm$}\,1.9} & 71.1\,{\scriptsize\color{stdorange}\scalebox{0.75}{$\pm$}\,5.1} \\
    LeMuMo-JEPA-AC$^\dagger$ & 90.0\,{\scriptsize\color{stdorange}\scalebox{0.75}{$\pm$}\,3.3} & 91.1\,{\scriptsize\color{stdorange}\scalebox{0.75}{$\pm$}\,1.9} & 75.6\,{\scriptsize\color{stdorange}\scalebox{0.75}{$\pm$}\,6.9} & 88.9\,{\scriptsize\color{stdorange}\scalebox{0.75}{$\pm$}\,5.1} & 90.0\,{\scriptsize\color{stdorange}\scalebox{0.75}{$\pm$}\,3.3} & 74.4\,{\scriptsize\color{stdorange}\scalebox{0.75}{$\pm$}\,8.4} \\
    \midrule
    \rowcolor{tableblue}
    \textbf{DepState-JEPA-AC (Ours)} & \textbf{95.6}\,{\scriptsize\color{stdorange}\scalebox{0.75}{$\pm$}\,1.9} & \textbf{95.6}\,{\scriptsize\color{stdorange}\scalebox{0.75}{$\pm$}\,1.9} & \textbf{81.1}\,{\scriptsize\color{stdorange}\scalebox{0.75}{$\pm$}\,5.1} & \textbf{94.4}\,{\scriptsize\color{stdorange}\scalebox{0.75}{$\pm$}\,1.9} & \textbf{93.3}\,{\scriptsize\color{stdorange}\scalebox{0.75}{$\pm$}\,3.3} & \textbf{78.9}\,{\scriptsize\color{stdorange}\scalebox{0.75}{$\pm$}\,3.8} \\
    \bottomrule
  \end{tabular}
\end{table}

\begin{table}[!htbp]
  \centering
  \caption{\textbf{Sim2Real \textit{Successes/Trials} for Individual Training Seeds.}}
  \label{tab:embodied-sim2real-counts}
  \footnotesize
  \setlength{\tabcolsep}{3pt}
  \renewcommand{\arraystretch}{1.12}
  \setlength{\aboverulesep}{0pt}
  \setlength{\belowrulesep}{0pt}
  \begin{tabular}{>{\raggedright\arraybackslash}p{\dimexpr0.30000000\linewidth-2\tabcolsep\relax}>{\centering\arraybackslash}p{\dimexpr0.06000000\linewidth-2\tabcolsep\relax}>{\centering\arraybackslash}p{\dimexpr0.10666667\linewidth-2\tabcolsep\relax}>{\centering\arraybackslash}p{\dimexpr0.10666667\linewidth-2\tabcolsep\relax}>{\centering\arraybackslash}p{\dimexpr0.10666667\linewidth-2\tabcolsep\relax}>{\centering\arraybackslash}p{\dimexpr0.10666667\linewidth-2\tabcolsep\relax}>{\centering\arraybackslash}p{\dimexpr0.10666667\linewidth-2\tabcolsep\relax}>{\centering\arraybackslash}p{\dimexpr0.10666667\linewidth-2\tabcolsep\relax}}
    \toprule
    \rowcolor{tableblue}
    & & \multicolumn{3}{c}{Task 1: Drawer} & \multicolumn{3}{c}{Task 2: Cup} \\
    \noalign{{\color{tableblue}\hrule height \cmidrulewidth}\kern-\cmidrulewidth}
    \cmidrule(lr){3-5}\cmidrule(lr){6-8}
    \rowcolor{tableblue}
    \multirow{-2}{*}{Method} & \multirow{-2}{*}{Seed} & Pull$\uparrow$ & Close$\uparrow$ & Full$\uparrow$ & Pick$\uparrow$ & Place$\uparrow$ & Full$\uparrow$ \\
    \midrule
    V-JEPA 2-AC & 0 & 20/30 & 24/30 & 14/30 & 18/30 & 21/30 & 11/30 \\
    V-JEPA 2.1-AC & 0 & 26/30 & 27/30 & 20/30 & 24/30 & 26/30 & 18/30 \\
    3D-JEPA-AC$^\dagger$ & 0 & 27/30 & 28/30 & \textbf{23/30} & 26/30 & 27/30 & 21/30 \\
    LeMuMo-JEPA-AC$^\dagger$ & 0 & 27/30 & 27/30 & 22/30 & 27/30 & 27/30 & 22/30 \\
    \rowcolor{tableblue}
    \textbf{DepState-JEPA-AC (Ours)} & 0 & \textbf{28/30} & \textbf{29/30} & \textbf{23/30} & \textbf{28/30} & \textbf{28/30} & \textbf{23/30} \\
    \midrule
    V-JEPA 2-AC & 1 & 21/30 & 25/30 & 15/30 & 20/30 & 23/30 & 14/30 \\
    V-JEPA 2.1-AC & 1 & 27/30 & 28/30 & 22/30 & 26/30 & 27/30 & 21/30 \\
    3D-JEPA-AC$^\dagger$ & 1 & 26/30 & 27/30 & 20/30 & 28/30 & 27/30 & 23/30 \\
    LeMuMo-JEPA-AC$^\dagger$ & 1 & 28/30 & 28/30 & 25/30 & 25/30 & 26/30 & 20/30 \\
    \rowcolor{tableblue}
    \textbf{DepState-JEPA-AC (Ours)} & 1 & \textbf{29/30} & \textbf{29/30} & \textbf{26/30} & \textbf{29/30} & \textbf{29/30} & \textbf{25/30} \\
    \midrule
    V-JEPA 2-AC & 2 & 19/30 & 24/30 & 12/30 & 19/30 & 22/30 & 12/30 \\
    V-JEPA 2.1-AC & 2 & 25/30 & 26/30 & 18/30 & 25/30 & 26/30 & 19/30 \\
    3D-JEPA-AC$^\dagger$ & 2 & 28/30 & 27/30 & 23/30 & 25/30 & 26/30 & 20/30 \\
    LeMuMo-JEPA-AC$^\dagger$ & 2 & 26/30 & 27/30 & 21/30 & \textbf{28/30} & \textbf{28/30} & \textbf{25/30} \\
    \rowcolor{tableblue}
    \textbf{DepState-JEPA-AC (Ours)} & 2 & \textbf{29/30} & \textbf{28/30} & \textbf{24/30} & \textbf{28/30} & 27/30 & 23/30 \\
    \bottomrule
  \end{tabular}
\end{table}

\textbf{Interpretation of physical supervision.}
Configuration and velocity readouts supply recursive proprioception.
TCP supervision, FK consistency, and accumulated object-pose losses shape
the predicted state; object poses do not feed the transition or other readouts.
Ex.11--Ex.13 assess these training paths through the feature and control
outcomes in Tables~\ref{tab:ablation-temporal} and~\ref{tab:ablation-control-tasks}.

\textbf{Multi-view real-robot rollouts.}
Figures~\ref{fig:real-rollout-task1} and~\ref{fig:real-rollout-task2}
expand the qualitative examples in Fig.~\ref{fig:realrobot-qualitative} with
RGB-D sequences for the drawer and cup tasks, respectively. Each task is
shown from two recorded viewpoints, exposing the manipulation sequence
and changes in visible geometry over time.
\begin{figure}[p]
  \centering
  \includegraphics[width=\linewidth]{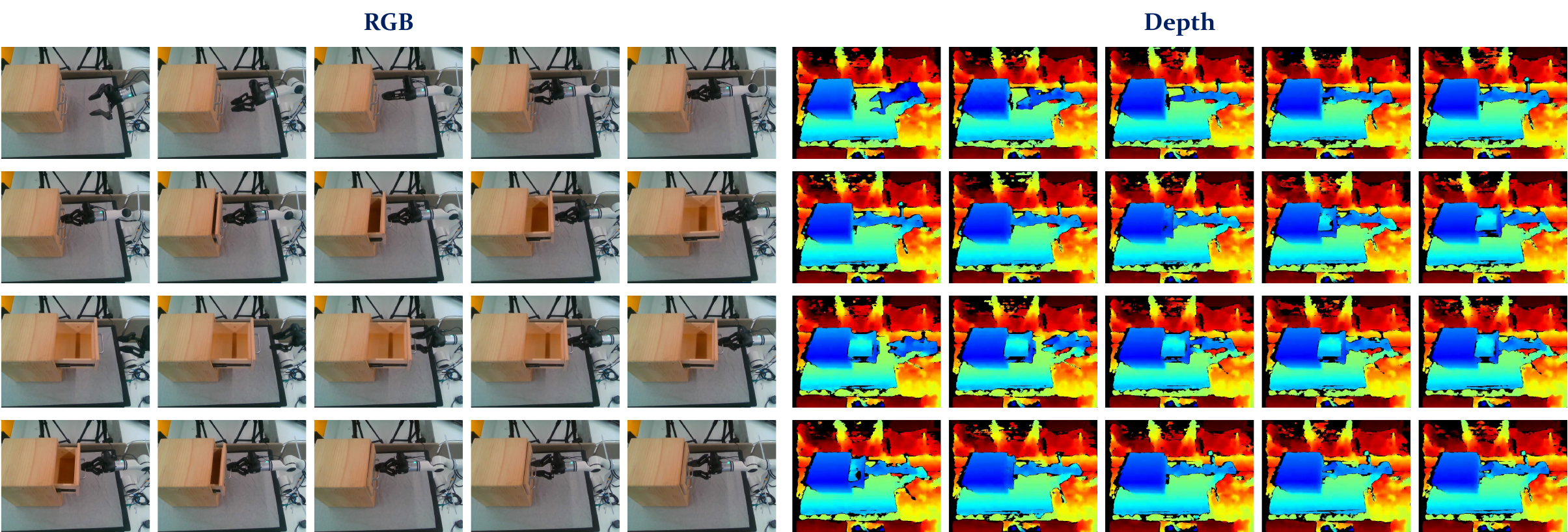}
  \par\vspace{3pt}{\small\textbf{(a) View 1}}\par\vspace{10pt}
  \includegraphics[width=\linewidth]{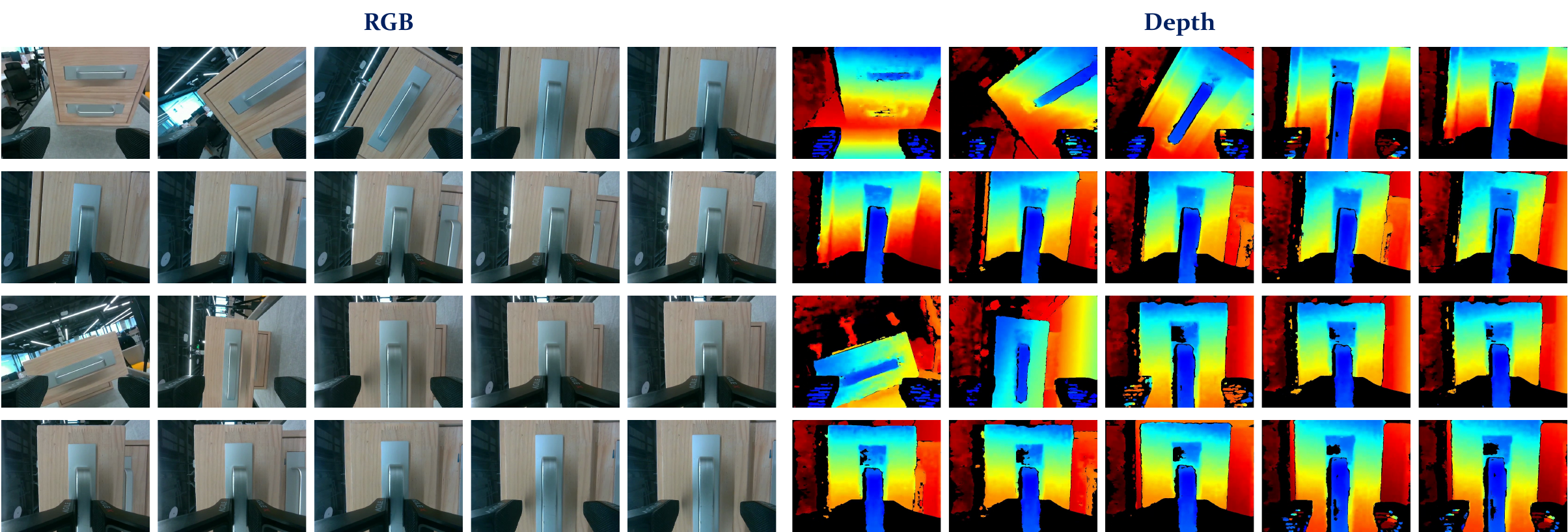}
  \par\vspace{3pt}{\small\textbf{(b) View 2}}
  \caption{\textbf{Multi-View RGB-D Rollout of Drawer Opening and Closing.}
  Two recorded viewpoints show the real-robot execution.
  Each panel places RGB frames on the left and depth frames on the right;
  within each modality, time advances from left to right and then top to bottom.}
  \label{fig:real-rollout-task1}
\end{figure}

\begin{figure}[p]
  \centering
  \includegraphics[width=\linewidth]{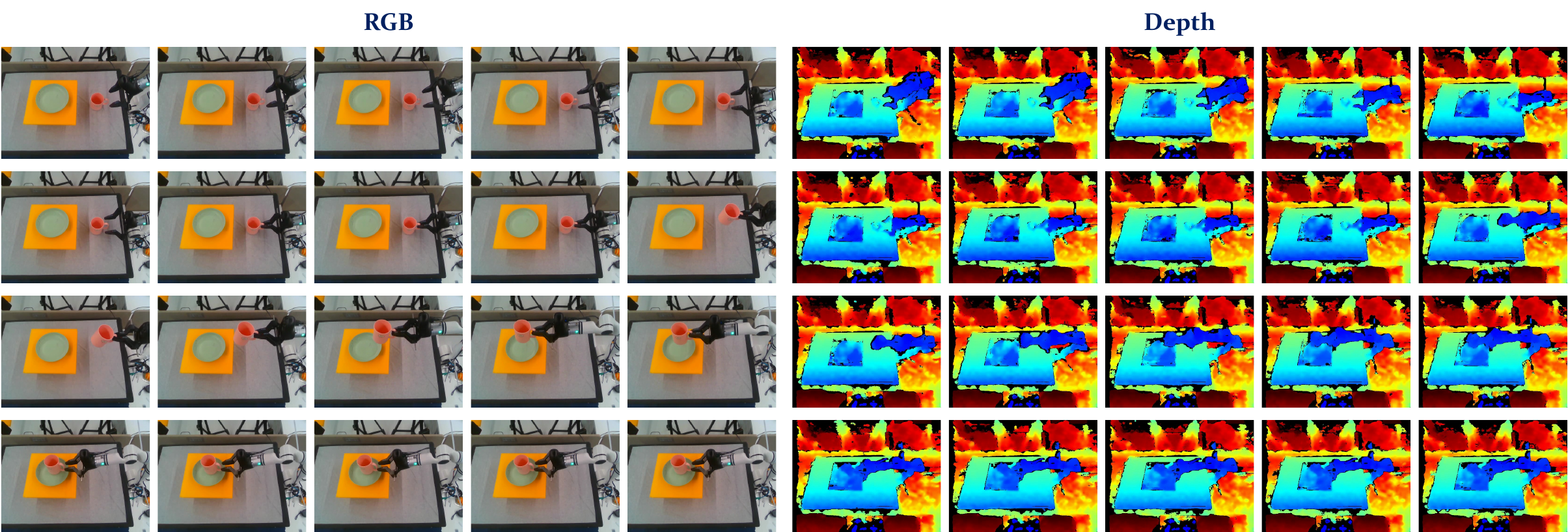}
  \par\vspace{3pt}{\small\textbf{(a) View 1}}\par\vspace{10pt}
  \includegraphics[width=\linewidth]{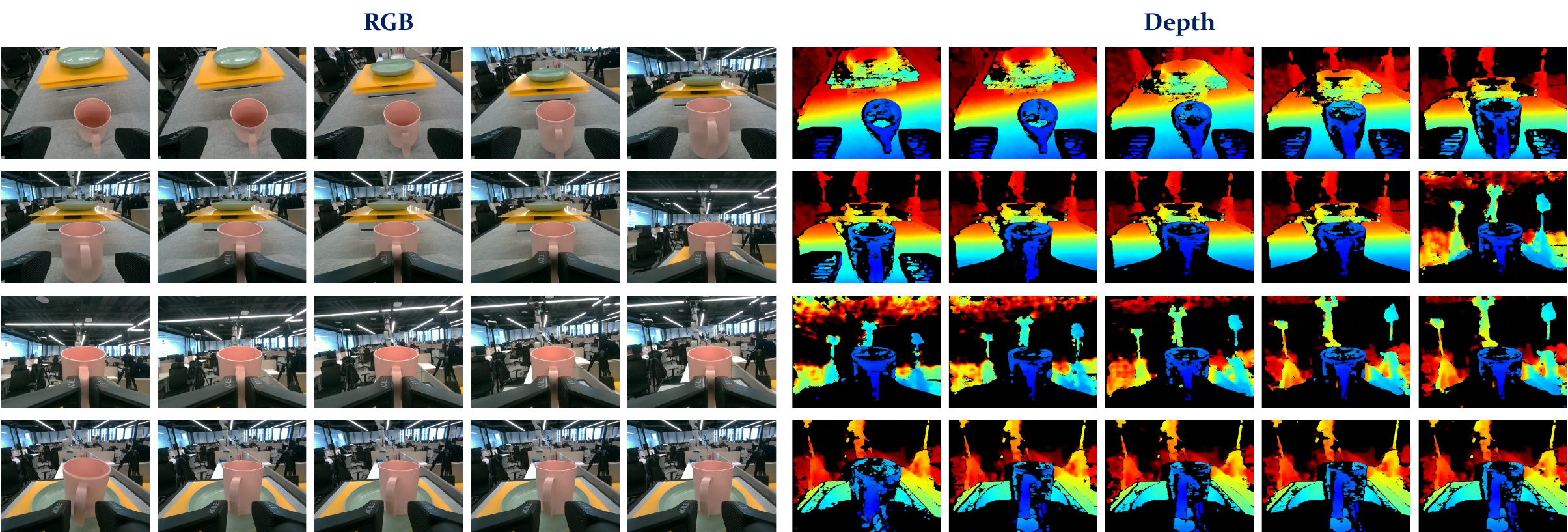}
  \par\vspace{3pt}{\small\textbf{(b) View 2}}
  \caption{\textbf{Multi-View RGB-D Rollout of Cup Picking and Placing.}
  Two recorded viewpoints show the real-robot execution.
  Each panel places RGB frames on the left and depth frames on the right;
  within each modality, time advances from left to right and then top to bottom.}
  \label{fig:real-rollout-task2}
\end{figure}

\clearpage

\subsection{Ablation Results, Protocols, and Statistical Analysis}
\label{app:ablation-evaluation}
\textbf{Organization and reference configuration.}
Tables~\ref{tab:ablation-geometry}--\ref{tab:ablation-temporal} record the configurations
and numerical results underlying Fig.~\ref{fig:ablation-study}(a)--(c), respectively.
Ex.16 is the full configuration in Appendices B--D and serves as the
common reference. Ex.1--Ex.3 form a two-factor
input experiment with Ex.16. Ex.4--Ex.15 each change the indicated design.
Capacity and loss-weight differences are identified with the relevant variants.
Entries report mean $\pm$ sample standard deviation across seeds 0, 1,
and 2. Bold marks the best displayed mean per metric, including ties.
Seed-level and task-level results appear in App.~\ref{app:ablation-detailed-results}.

Figure~\ref{fig:ablation-study} plots the three-seed means; filled markers
identify Ex.16. Connected points denote independent interventions,
not cumulative ablations.

\textbf{Task means and information access.}
L1, Cos, and SR are shared by the main-text ablation plots and the
three appendix tables. Let
$\mathcal T_m$ be the fixed set of tasks evaluated by metric $m$, and
$s_{e,r,t}^{m}$ the score of experiment $e$, seed $r$, and task $t$.
The reported point estimate is
\begin{equation}
 \bar s_e^m=\frac{1}{3}\sum_{r=0}^{2}
 \left(\frac{1}{|\mathcal T_m|}\sum_{t\in\mathcal T_m}s_{e,r,t}^m\right).
 \label{eq:ablation-task-mean}
\end{equation}
The three configuration tables report the standard deviation of the
seed-level task means. Tables~\ref{tab:ablation-seeds}--\ref{tab:ablation-control-tasks}
provide seed scores and task-level breakdowns. Each configuration uses
the same equally weighted task set.

For L1/Cos, $\mathcal T_m$ contains the LingBot-Depth RGB-D completion task
and the two held-out robot forecasting tasks, Drawer and Cup. Completion first
averages RGB/depth equally, retaining the scene-balanced reduction in
Appendix~\ref{app:spatial-completion}. Each forecasting task first averages
endpoints $h\in\{1,8,16,24,32\}$ on identical starting windows and then
averages equally over episodes. Its principal modality is RGB, consistent
with Table~\ref{tab:temporal-prediction}. All Ex. configurations use the
same frozen reference encoders, non-affine normalization, and training-only
readout calibration. The ablation forecasting track uses measured proprioception only at initialization,
then predicted configurations and velocities. In contrast,
Table~\ref{tab:temporal-prediction} conditions on recorded proprioception
at every step. Its scores are not included in these task means, and the
two tables are not presented as a paired comparison of conditioning modes.

For SR, $\mathcal T_{\rm SR}$ contains all 50 Meta-World tasks and the two
real manipulation tasks. Within each Meta-World task, actor-only and
latent-MPC success are averaged equally before the 52-task mean is formed;
each real task contributes its visual-goal MPC success once. The original
task-specific success criteria apply, with three seeds and thirty real
trials per full task and seed. Only full-task success enters this aggregate; atomic
task diagnostics do not receive additional task weights.
DMC episode return is not a success rate and is excluded.
Table~\ref{tab:ablation-control-tasks} retains controller-specific and
real-task scores: the aggregate assigns 50/52 of its weight to Meta-World
and 2/52 to real tasks, so it is primarily a simulated-control summary.

\textbf{Training and measurement controls.}
The comparisons hold pretrained initialization, data order, splits,
augmentation, target validity, optimizer settings, and task/control budgets
fixed, together with unchanged module dimensions. Loss interventions persist
through all applicable training and adaptation objectives. Each Ex. uses
the final Stage-B representation. Equal-budget completion adaptation with
a frozen representation measures the utility of that representation together
with its fitted readout. Common-reference calibration uses observed training
features only and is frozen before prediction scoring. Changes to state
construction also apply to its EMA copy; input-availability interventions
retain the stated teacher target interface.

The tables report sample standard deviations across three training seeds,
not confidence intervals over scenes or trials. Architectural interventions
can change active capacity despite retaining unused parameters. The
interpretation below therefore distinguishes component removal from
capacity-matched evidence and does not infer significance from ranking alone.

\subsubsection{Geometry-Grounded State Construction}
\label{app:ablation-geometry}
\begin{table}[!htbp]
  \centering
  \caption{\textbf{Geometry-Grounded State Construction.}}
  \label{tab:ablation-geometry}
  \footnotesize
  \setlength{\tabcolsep}{3pt}
  \renewcommand{\arraystretch}{1.16}
  \setlength{\aboverulesep}{0pt}
  \setlength{\belowrulesep}{0pt}
  \begin{tabular}{>{\centering\arraybackslash}p{\dimexpr0.09\linewidth-2\tabcolsep\relax}>{\raggedright\arraybackslash}p{\dimexpr0.49\linewidth-2\tabcolsep\relax}>{\centering\arraybackslash}p{\dimexpr0.14\linewidth-2\tabcolsep\relax}>{\centering\arraybackslash}p{\dimexpr0.14\linewidth-2\tabcolsep\relax}>{\centering\arraybackslash}p{\dimexpr0.14\linewidth-2\tabcolsep\relax}}
    \toprule
    \rowcolor{tableblue}
    ID & Experimental Setting & L1$\downarrow$ & Cos$\uparrow$ & SR$\uparrow$ \\
    \midrule
    Ex.1 & Replace online depth content with RGB and disable metric XYZ positional encoding. & 0.408\,{\scriptsize\color{stdorange}\scalebox{0.75}{$\pm$}\,0.021} & 0.832\,{\scriptsize\color{stdorange}\scalebox{0.75}{$\pm$}\,0.017} & 76.8\,{\scriptsize\color{stdorange}\scalebox{0.75}{$\pm$}\,2.9} \\
    Ex.2 & Retain depth content; replace metric XYZ positional encoding with an unknown-geometry embedding. & 0.369\,{\scriptsize\color{stdorange}\scalebox{0.75}{$\pm$}\,0.017} & 0.860\,{\scriptsize\color{stdorange}\scalebox{0.75}{$\pm$}\,0.014} & 80.1\,{\scriptsize\color{stdorange}\scalebox{0.75}{$\pm$}\,2.7} \\
    Ex.3 & Retain metric XYZ positional encoding; replace online depth content with RGB. & 0.359\,{\scriptsize\color{stdorange}\scalebox{0.75}{$\pm$}\,0.017} & 0.869\,{\scriptsize\color{stdorange}\scalebox{0.75}{$\pm$}\,0.014} & 81.3\,{\scriptsize\color{stdorange}\scalebox{0.75}{$\pm$}\,2.7} \\
    \midrule
    \rowcolor{tableblue}
    \textbf{Ex.16} & Full model: retain both depth content and metric XYZ positional encoding. & \textbf{0.340}\,{\scriptsize\color{stdorange}\scalebox{0.75}{$\pm$}\,0.013} & \textbf{0.880}\,{\scriptsize\color{stdorange}\scalebox{0.75}{$\pm$}\,0.011} & \textbf{83.4}\,{\scriptsize\color{stdorange}\scalebox{0.75}{$\pm$}\,2.9} \\
    \bottomrule
  \end{tabular}
  \par\vspace{3pt}{\footnotesize\raggedright All four retain two active backbones, the full RGB-D teacher targets, and identical supervision. Ex.3 still requires depth for metric lifting.\par}
\end{table}

\textbf{Ex.1--Ex.3 and Ex.16: content versus metric coordinates.}
The depth-content factor denotes whether the second online content encoder
receives measured depth. When disabled (Ex.1/Ex.3), it receives the same
RGB image as the first branch, with RGB normalization, independent
weights, and unchanged token count. This preserves an active second
backbone rather than replacing its features with a constant.
The metric-coordinate factor denotes visible-depth XYZ positional encoding in both
streams. When disabled (Ex.1/Ex.2), it is replaced by the same learned
unknown-geometry embedding at all positions. Ex.1 disables both factors,
Ex.2 retains only depth content, Ex.3 retains only metric coordinates,
and Ex.16 retains both. Ex.3 still uses depth for lifting and is not an
RGB-only inference method.

All four experiments retain the original full RGB-D teacher input
interface and depth target supervision. The input substitution is an
online-wrapper intervention; teacher parameters still follow EMA.
Masks, validity metadata, target sampling, physical labels, and known query
intrinsics are identical. Thus the comparison isolates online input
content at fixed supervision, not the benefit of depth-free pretraining.
The depth-content contrasts are Ex.2--Ex.1 and Ex.16--Ex.3; the
coordinate contrasts are Ex.3--Ex.1 and Ex.16--Ex.2. For each metric, the interaction
is $s_{16}-s_2-s_3+s_1$; its favorable sign differs between error and
similarity/success metrics. Full-model superiority alone does not
establish a super-additive interaction.

\textbf{Observed geometry effects.}
Relative to Ex.1, depth content alone (Ex.2) reduces L1 by 0.039 and
increases SR by 3.3 percentage points; coordinates alone (Ex.3) reduce L1
by 0.049 and increase SR by 4.5 points. Ex.16 improves both further, with
L1 0.340 and SR 83.4\%. With coordinates already present, depth content
reduces L1 by 0.019 and raises SR by 2.1 points; with depth content already
present, coordinates reduce L1 by 0.029 and raise SR by 3.3 points.
At the reported means, the combined gains are
smaller than the sums of the individual gains; the results support
complementary contributions without establishing super-additivity.

\subsubsection{Selective Sharing and Spatial Readout}
\label{app:ablation-sharing}
\begin{table}[!htbp]
  \centering
  \caption{\textbf{Selective Sharing and Spatial Readout.}}
  \label{tab:ablation-sharing}
  \footnotesize
  \setlength{\tabcolsep}{3pt}
  \renewcommand{\arraystretch}{1.16}
  \setlength{\aboverulesep}{0pt}
  \setlength{\belowrulesep}{0pt}
  \begin{tabular}{>{\centering\arraybackslash}p{\dimexpr0.09\linewidth-2\tabcolsep\relax}>{\raggedright\arraybackslash}p{\dimexpr0.49\linewidth-2\tabcolsep\relax}>{\centering\arraybackslash}p{\dimexpr0.14\linewidth-2\tabcolsep\relax}>{\centering\arraybackslash}p{\dimexpr0.14\linewidth-2\tabcolsep\relax}>{\centering\arraybackslash}p{\dimexpr0.14\linewidth-2\tabcolsep\relax}}
    \toprule
    \rowcolor{tableblue}
    ID & Experimental Setting & L1$\downarrow$ & Cos$\uparrow$ & SR$\uparrow$ \\
    \midrule
    Ex.4 & Disable the private cross-attention residual in the modality decoders; retain private projections and factorization losses. & 0.390\,{\scriptsize\color{stdorange}\scalebox{0.75}{$\pm$}\,0.019} & 0.852\,{\scriptsize\color{stdorange}\scalebox{0.75}{$\pm$}\,0.016} & 83.2\,{\scriptsize\color{stdorange}\scalebox{0.75}{$\pm$}\,2.7} \\
    Ex.5 & Add unit-weight alignment of corresponding private features to the existing shared-feature alignment. & 0.376\,{\scriptsize\color{stdorange}\scalebox{0.75}{$\pm$}\,0.018} & 0.855\,{\scriptsize\color{stdorange}\scalebox{0.75}{$\pm$}\,0.016} & 81.5\,{\scriptsize\color{stdorange}\scalebox{0.75}{$\pm$}\,2.9} \\
    Ex.6 & Replace projected aggregation keys/values by repeated valid-token means; retain learned queries and state size. & 0.440\,{\scriptsize\color{stdorange}\scalebox{0.75}{$\pm$}\,0.029} & 0.799\,{\scriptsize\color{stdorange}\scalebox{0.75}{$\pm$}\,0.025} & 77.0\,{\scriptsize\color{stdorange}\scalebox{0.75}{$\pm$}\,1.9} \\
    Ex.7 & Replace unit-ray query inputs with normalized pixel vectors $(u/W,v/H,1)$; retain the query MLP. & \textbf{0.340}\,{\scriptsize\color{stdorange}\scalebox{0.75}{$\pm$}\,0.013} & \textbf{0.880}\,{\scriptsize\color{stdorange}\scalebox{0.75}{$\pm$}\,0.011} & \textbf{83.5}\,{\scriptsize\color{stdorange}\scalebox{0.75}{$\pm$}\,2.7} \\
    Ex.8 & Remove completion supervision throughout training/adaptation; retain dense JEPA and temporal objectives. & 0.361\,{\scriptsize\color{stdorange}\scalebox{0.75}{$\pm$}\,0.018} & 0.870\,{\scriptsize\color{stdorange}\scalebox{0.75}{$\pm$}\,0.014} & 82.8\,{\scriptsize\color{stdorange}\scalebox{0.75}{$\pm$}\,2.9} \\
    \midrule
    \rowcolor{tableblue}
    \textbf{Ex.16} & Full model: private decoding, shared-only alignment, learned aggregation, ray queries, and completion supervision. & \textbf{0.340}\,{\scriptsize\color{stdorange}\scalebox{0.75}{$\pm$}\,0.013} & \textbf{0.880}\,{\scriptsize\color{stdorange}\scalebox{0.75}{$\pm$}\,0.011} & 83.4\,{\scriptsize\color{stdorange}\scalebox{0.75}{$\pm$}\,2.9} \\
    \bottomrule
  \end{tabular}
  \par\vspace{3pt}{\footnotesize\raggedright Each variant changes only the listed design relative to Ex.16; capacity and loss-weight qualifications are explained below.\par}
\end{table}

\textbf{Ex.4--Ex.8: one interface decision at a time.}
Ex.4 disables the private cross-attention residual in each modality
decoder, retaining its query path, private projections, and factorization
losses. This removes both private observation access and active private
attention capacity, so its effect cannot be attributed to private information
alone.
Ex.5 retains the shared alignment and adds a unit-weight cosine alignment
of corresponding private features (full-space alignment); Ex.16 aligns shared features
only (shared-only alignment). Ex.5 also increases total alignment weight;
its comparison therefore does not isolate alignment scope from strength.
Ex.6 replaces the aggregator's shared-feature keys and
values, after their projections, by their valid-token means repeated to
the original memory length. Learned queries, attention/FFN modules,
state size, and temporal interface remain unchanged. Thus Ex.6 removes
local information available to query attention without a large new
projection into the complete state; Ex.16 uses the original aggregator.
This contrast attributes the change to local-content access in this
aggregator. Ex.4 separately tests the private decoding path, and Ex.5 tests
the combined change in alignment scope and total weight.

Ex.7 replaces unit-ray inputs by normalized pixel vectors
$(u/W,v/H,1)$ while retaining time, Fourier input width, and the query
MLP. Fixed-intrinsics pixel and ray coordinates are deterministically
related, so this tests an encoding bias, not access to extra scene
information. Ex.8 sets only the completion-loss coefficient to zero
throughout adaptation and replay, retaining dense JEPA and temporal
objectives. All remaining settings match Ex.16.

\textbf{Observed sharing and readout effects.}
Mean aggregation (Ex.6) yields the largest degradation in this group,
with L1 0.440, Cos 0.799, and SR 77.0\%. Removing private decoding (Ex.4)
primarily affects visual prediction (L1 0.390), with SR remaining at
83.2\%. Ex.7 and Ex.16 tie in L1/Cos at the reported precision, while
Ex.7 has a slightly higher SR (83.5\% versus 83.4\%). Thus these aggregate
results place ray and pixel queries at comparable feature accuracy under
this camera protocol. Completion supervision and shared-only alignment
also yield better feature scores in their respective listed comparisons.
Together with Ex.4, these outcomes distinguish visual decoding from the
state pathway used by control.

\subsubsection{Temporal Prediction and Physical Grounding}
\label{app:ablation-temporal}
\begin{table}[!htbp]
  \centering
  \caption{\textbf{Temporal Prediction and Physical Grounding.}}
  \label{tab:ablation-temporal}
  \footnotesize
  \setlength{\tabcolsep}{3pt}
  \renewcommand{\arraystretch}{1.16}
  \setlength{\aboverulesep}{0pt}
  \setlength{\belowrulesep}{0pt}
  \begin{tabular}{>{\centering\arraybackslash}p{\dimexpr0.09\linewidth-2\tabcolsep\relax}>{\raggedright\arraybackslash}p{\dimexpr0.49\linewidth-2\tabcolsep\relax}>{\centering\arraybackslash}p{\dimexpr0.14\linewidth-2\tabcolsep\relax}>{\centering\arraybackslash}p{\dimexpr0.14\linewidth-2\tabcolsep\relax}>{\centering\arraybackslash}p{\dimexpr0.14\linewidth-2\tabcolsep\relax}}
    \toprule
    \rowcolor{tableblue}
    ID & Experimental Setting & L1$\downarrow$ & Cos$\uparrow$ & SR$\uparrow$ \\
    \midrule
    Ex.9 & Remove future-state matching; retain future-feature matching and physical supervision. & 0.397\,{\scriptsize\color{stdorange}\scalebox{0.75}{$\pm$}\,0.024} & 0.827\,{\scriptsize\color{stdorange}\scalebox{0.75}{$\pm$}\,0.021} & 79.5\,{\scriptsize\color{stdorange}\scalebox{0.75}{$\pm$}\,2.8} \\
    Ex.10 & Remove future-feature matching; retain future-state matching and physical supervision. & 0.392\,{\scriptsize\color{stdorange}\scalebox{0.75}{$\pm$}\,0.019} & 0.833\,{\scriptsize\color{stdorange}\scalebox{0.75}{$\pm$}\,0.017} & 82.2\,{\scriptsize\color{stdorange}\scalebox{0.75}{$\pm$}\,2.8} \\
    Ex.11 & Remove object-trajectory supervision; retain the object head and integration interface. & 0.344\,{\scriptsize\color{stdorange}\scalebox{0.75}{$\pm$}\,0.014} & 0.877\,{\scriptsize\color{stdorange}\scalebox{0.75}{$\pm$}\,0.012} & 83.2\,{\scriptsize\color{stdorange}\scalebox{0.75}{$\pm$}\,2.9} \\
    Ex.12 & Remove FK consistency regularization; retain independent joint/TCP heads and their direct labels. & 0.359\,{\scriptsize\color{stdorange}\scalebox{0.75}{$\pm$}\,0.018} & 0.864\,{\scriptsize\color{stdorange}\scalebox{0.75}{$\pm$}\,0.016} & 82.5\,{\scriptsize\color{stdorange}\scalebox{0.75}{$\pm$}\,2.8} \\
    Ex.13 & Apply FK gradients to joint/TCP heads only via a detached state; keep direct supervision undetached. & 0.350\,{\scriptsize\color{stdorange}\scalebox{0.75}{$\pm$}\,0.016} & 0.871\,{\scriptsize\color{stdorange}\scalebox{0.75}{$\pm$}\,0.014} & 83.1\,{\scriptsize\color{stdorange}\scalebox{0.75}{$\pm$}\,2.9} \\
    Ex.14 & Recursively update private memory from predicted states without future observations; add a memory updater. & 0.351\,{\scriptsize\color{stdorange}\scalebox{0.75}{$\pm$}\,0.017} & 0.871\,{\scriptsize\color{stdorange}\scalebox{0.75}{$\pm$}\,0.015} & 83.1\,{\scriptsize\color{stdorange}\scalebox{0.75}{$\pm$}\,2.9} \\
    Ex.15 & Use separate spatial and temporal decoder copies with the same per-branch architecture and initialization. & \textbf{0.333}\,{\scriptsize\color{stdorange}\scalebox{0.75}{$\pm$}\,0.014} & \textbf{0.885}\,{\scriptsize\color{stdorange}\scalebox{0.75}{$\pm$}\,0.011} & \textbf{83.4}\,{\scriptsize\color{stdorange}\scalebox{0.75}{$\pm$}\,2.6} \\
    \midrule
    \rowcolor{tableblue}
    \textbf{Ex.16} & Full model: both temporal targets, object supervision, FK gradients to state and heads, fixed private memory, and shared decoders. & 0.340\,{\scriptsize\color{stdorange}\scalebox{0.75}{$\pm$}\,0.013} & 0.880\,{\scriptsize\color{stdorange}\scalebox{0.75}{$\pm$}\,0.011} & \textbf{83.4}\,{\scriptsize\color{stdorange}\scalebox{0.75}{$\pm$}\,2.9} \\
    \bottomrule
  \end{tabular}
  \par\vspace{3pt}{\footnotesize\raggedright Joint and velocity supervision remain active. Ex.14 and Ex.15 add parameters; these are not matched-total-capacity comparisons.\par}
\end{table}

\textbf{Ex.9--Ex.13: predictive targets and physical gradients.}
Ex.9 removes future-state matching; Ex.10 removes future-feature matching.
Both retain the other temporal objective, state construction, and
physical supervision. Ex.11 removes only object-trajectory supervision,
retaining the object head and integration interface. Ex.12 removes only
FK regularization, preserving independent joint/TCP heads and their
direct labels. Ex.13 retains FK but limits its gradients to those heads:
compute the FK loss through an additional head evaluation on
$\operatorname{sg}(\hat S)$, reusing head parameters. Direct supervised
losses still use the ordinary, undetached state. Comparing Ex.12, Ex.13,
and Ex.16 separates absent consistency, output-head regularization, and
consistency gradients entering the latent transition and representation.
Joint and velocity supervision remain active in all variants; these
endpoint heads supply recursive proprioceptive inputs. The object accumulator serves
trajectory supervision; it never supplies future input to the transition.

\textbf{Ex.14: fixed versus updated private memory.}
Ex.14 recursively updates private memory without future observations.
Each modality uses one pre-LN residual cross-attention/FFN block $U_m$:
private queries have width 128, state keys/values project from 384 to 128,
attention has two heads, and the FFN width is 512. There is no token
self-attention or dropout. The attention and FFN output projections are
zero-initialized, giving an initial identity map. Starting from $P_0$,
Ex.14 uses $P_{h+1}=U_m(P_h,\hat S_{h+1})$ for decoding and subsequent
updates. No private memory enters the transition. The comparison with
Ex.16 includes the updater's added capacity and tests this particular
memory mechanism, rather than establishing the optimality of fixed memory.

\textbf{Ex.15: decoder reuse.}
Ex.15 uses separate spatial and temporal copies of $h_\eta,g_m^\ell$,
initialized from the same Stage-A decoders. Each branch retains Ex.16's
width and depth, increasing total decoder parameters. This comparison
holds per-branch architecture fixed; it does not match the combined
parameter budget of the shared and untied designs.

\textbf{Observed temporal and physical effects.}
Removing future-state matching (Ex.9) lowers SR to 79.5\%; removing
future-feature matching (Ex.10) worsens L1 to 0.392. Untied decoders
(Ex.15) obtain the best feature scores in this group (L1 0.333, Cos 0.885)
and tie Ex.16 in displayed SR (83.4\%), while adding decoder parameters.
Shared decoders provide one spatial/temporal readout interface, while the
separate copies trade additional parameters for better feature scores.

Relative to no FK (Ex.12), Ex.16 raises aggregate SR from 82.5\% to
83.4\%, a 0.9-point difference between the displayed means.
Restricting FK gradients to the heads (Ex.13) gives 83.1\%, compared
with 83.4\% when those gradients also reach the state. These results
support a modest control benefit of the regularizer at fixed direct
supervision. In Table~\ref{tab:ablation-control-tasks}, removing FK reduces
Drawer/Cup full-task success from 81.1/78.9\% to 77.8/74.4\%.
Removing object supervision (Ex.11) leaves Drawer success at 81.1\%
and reduces Cup success to 76.7\%, with aggregate SR 83.2\%.
These comparisons quantify the control effect of the auxiliary training
paths with joint and velocity supervision retained. The reported uncertainty
uses training seeds as replicates, as in the other control tables.

\subsubsection{Seed-Level and Task-Level Results}
\label{app:ablation-detailed-results}
Table~\ref{tab:ablation-seeds} reports each seed's aggregate scores before
rounding across seeds. Table~\ref{tab:ablation-prediction-tasks} separates
RGB-D completion from Drawer and Cup forecasting: completion averages
RGB and depth equally, and each forecasting entry averages horizons
$1,8,16,24,32$ within that task. Table~\ref{tab:ablation-control-tasks} separates
Meta-World policy learning (PL) and latent MPC from the six real tasks.
Real-task rates use 30 trials per seed; full-task trials are independent
of atomic-task trials. Only the two Full columns enter aggregate SR,
with 50 task weights for Meta-World and one for each real full task.
Task-level entries are mean $\pm$ sample standard deviation across three
seeds. Bold compares experiments within each displayed column, including
ties at the reported precision; the seed table compares within each seed.
\begin{table}[!htbp]
\centering
\caption{\textbf{Ablation \textit{L1}, \textit{Cos}, and \textit{SR} by Training Seed.}}
\label{tab:ablation-seeds}
\footnotesize
\setlength{\tabcolsep}{2pt}
\renewcommand{\arraystretch}{1.12}
\setlength{\aboverulesep}{0pt}
\setlength{\belowrulesep}{0pt}
\begin{tabular}{>{\centering\arraybackslash}p{\dimexpr0.10000000\linewidth-2\tabcolsep\relax}>{\centering\arraybackslash}p{\dimexpr0.10000000\linewidth-2\tabcolsep\relax}>{\centering\arraybackslash}p{\dimexpr0.10000000\linewidth-2\tabcolsep\relax}>{\centering\arraybackslash}p{\dimexpr0.10000000\linewidth-2\tabcolsep\relax}>{\centering\arraybackslash}p{\dimexpr0.10000000\linewidth-2\tabcolsep\relax}>{\centering\arraybackslash}p{\dimexpr0.10000000\linewidth-2\tabcolsep\relax}>{\centering\arraybackslash}p{\dimexpr0.10000000\linewidth-2\tabcolsep\relax}>{\centering\arraybackslash}p{\dimexpr0.10000000\linewidth-2\tabcolsep\relax}>{\centering\arraybackslash}p{\dimexpr0.10000000\linewidth-2\tabcolsep\relax}>{\centering\arraybackslash}p{\dimexpr0.10000000\linewidth-2\tabcolsep\relax}}
\toprule
\rowcolor{tableblue}
 & \multicolumn{3}{c}{Seed 0} & \multicolumn{3}{c}{Seed 1} & \multicolumn{3}{c}{Seed 2} \\
\noalign{{\color{tableblue}\hrule height \cmidrulewidth}\kern-\cmidrulewidth}
\cmidrule(lr){2-4}\cmidrule(lr){5-7}\cmidrule(lr){8-10}
\rowcolor{tableblue}
\multirow{-2}{*}{ID} & L1$\downarrow$ & Cos$\uparrow$ & SR$\uparrow$ & L1$\downarrow$ & Cos$\uparrow$ & SR$\uparrow$ & L1$\downarrow$ & Cos$\uparrow$ & SR$\uparrow$ \\
\midrule
Ex.1 & 0.407 & 0.834 & 74.0 & 0.388 & 0.848 & 79.7 & 0.430 & 0.814 & 76.7 \\
Ex.2 & 0.367 & 0.861 & 77.2 & 0.353 & 0.874 & 82.7 & 0.387 & 0.845 & 80.3 \\
Ex.3 & 0.358 & 0.870 & 78.5 & 0.342 & 0.882 & 84.0 & 0.377 & 0.854 & 81.2 \\
Ex.4 & 0.389 & 0.853 & 80.6 & 0.372 & 0.867 & 86.0 & 0.409 & 0.835 & 82.9 \\
Ex.5 & 0.374 & 0.856 & 78.8 & 0.358 & 0.870 & 84.5 & 0.395 & 0.838 & 81.3 \\
Ex.6 & 0.439 & 0.800 & 75.1 & 0.411 & 0.824 & 78.9 & 0.469 & 0.773 & 77.1 \\
Ex.7 & 0.339 & 0.882 & \textbf{80.9} & 0.327 & 0.890 & \textbf{86.3} & 0.353 & 0.869 & 83.2 \\
Ex.8 & 0.359 & 0.872 & 80.0 & 0.345 & 0.883 & 85.8 & 0.380 & 0.854 & 82.7 \\
Ex.9 & 0.397 & 0.828 & 76.7 & 0.374 & 0.847 & 82.3 & 0.421 & 0.805 & 79.4 \\
Ex.10 & 0.391 & 0.835 & 79.4 & 0.373 & 0.850 & 85.0 & 0.411 & 0.816 & 82.1 \\
Ex.11 & 0.343 & 0.879 & 80.3 & 0.330 & 0.888 & 86.0 & 0.358 & 0.865 & 83.3 \\
Ex.12 & 0.358 & 0.866 & 79.7 & 0.341 & 0.879 & 85.4 & 0.378 & 0.848 & 82.4 \\
Ex.13 & 0.349 & 0.873 & 80.3 & 0.335 & 0.884 & 86.0 & 0.367 & 0.857 & 83.0 \\
Ex.14 & 0.349 & 0.873 & 80.4 & 0.334 & 0.885 & 86.2 & 0.369 & 0.854 & 82.8 \\
Ex.15 & \textbf{0.332} & \textbf{0.887} & 80.7 & \textbf{0.320} & \textbf{0.896} & 85.9 & \textbf{0.348} & \textbf{0.873} & \textbf{83.7} \\
\midrule
\rowcolor{tableblue}
\textbf{Ex.16} & 0.339 & 0.881 & 80.6 & 0.327 & 0.890 & \textbf{86.3} & 0.353 & 0.869 & 83.3 \\
\bottomrule
\end{tabular}
\end{table}

\begin{table}[!htbp]
\centering
\caption{\textbf{Task-Level Ablation \textit{L1} and \textit{Cos}.}}
\label{tab:ablation-prediction-tasks}
\footnotesize
\setlength{\tabcolsep}{2pt}
\renewcommand{\arraystretch}{1.12}
\setlength{\aboverulesep}{0pt}
\setlength{\belowrulesep}{0pt}
\begin{tabular}{>{\centering\arraybackslash}p{\dimexpr0.08000000\linewidth-2\tabcolsep\relax}>{\centering\arraybackslash}p{\dimexpr0.15333333\linewidth-2\tabcolsep\relax}>{\centering\arraybackslash}p{\dimexpr0.15333333\linewidth-2\tabcolsep\relax}>{\centering\arraybackslash}p{\dimexpr0.15333333\linewidth-2\tabcolsep\relax}>{\centering\arraybackslash}p{\dimexpr0.15333333\linewidth-2\tabcolsep\relax}>{\centering\arraybackslash}p{\dimexpr0.15333333\linewidth-2\tabcolsep\relax}>{\centering\arraybackslash}p{\dimexpr0.15333333\linewidth-2\tabcolsep\relax}}
\toprule
\rowcolor{tableblue}
 & \multicolumn{2}{c}{RGB-D Completion} & \multicolumn{2}{c}{Drawer Forecasting} & \multicolumn{2}{c}{Cup Forecasting} \\
\noalign{{\color{tableblue}\hrule height \cmidrulewidth}\kern-\cmidrulewidth}
\cmidrule(lr){2-3}\cmidrule(lr){4-5}\cmidrule(lr){6-7}
\rowcolor{tableblue}
\multirow{-2}{*}{ID} & L1$\downarrow$ & Cos$\uparrow$ & L1$\downarrow$ & Cos$\uparrow$ & L1$\downarrow$ & Cos$\uparrow$ \\
\midrule
Ex.1 & 0.308\,{\scriptsize\color{stdorange}\scalebox{0.75}{$\pm$}\,0.008} & 0.931\,{\scriptsize\color{stdorange}\scalebox{0.75}{$\pm$}\,0.005} & 0.427\,{\scriptsize\color{stdorange}\scalebox{0.75}{$\pm$}\,0.024} & 0.815\,{\scriptsize\color{stdorange}\scalebox{0.75}{$\pm$}\,0.019} & 0.490\,{\scriptsize\color{stdorange}\scalebox{0.75}{$\pm$}\,0.030} & 0.750\,{\scriptsize\color{stdorange}\scalebox{0.75}{$\pm$}\,0.028} \\
Ex.2 & 0.263\,{\scriptsize\color{stdorange}\scalebox{0.75}{$\pm$}\,0.006} & 0.947\,{\scriptsize\color{stdorange}\scalebox{0.75}{$\pm$}\,0.004} & 0.387\,{\scriptsize\color{stdorange}\scalebox{0.75}{$\pm$}\,0.020} & 0.847\,{\scriptsize\color{stdorange}\scalebox{0.75}{$\pm$}\,0.016} & 0.456\,{\scriptsize\color{stdorange}\scalebox{0.75}{$\pm$}\,0.026} & 0.785\,{\scriptsize\color{stdorange}\scalebox{0.75}{$\pm$}\,0.023} \\
Ex.3 & 0.271\,{\scriptsize\color{stdorange}\scalebox{0.75}{$\pm$}\,0.008} & 0.945\,{\scriptsize\color{stdorange}\scalebox{0.75}{$\pm$}\,0.006} & 0.373\,{\scriptsize\color{stdorange}\scalebox{0.75}{$\pm$}\,0.020} & 0.857\,{\scriptsize\color{stdorange}\scalebox{0.75}{$\pm$}\,0.016} & 0.433\,{\scriptsize\color{stdorange}\scalebox{0.75}{$\pm$}\,0.025} & 0.805\,{\scriptsize\color{stdorange}\scalebox{0.75}{$\pm$}\,0.022} \\
Ex.4 & 0.314\,{\scriptsize\color{stdorange}\scalebox{0.75}{$\pm$}\,0.008} & 0.928\,{\scriptsize\color{stdorange}\scalebox{0.75}{$\pm$}\,0.006} & 0.400\,{\scriptsize\color{stdorange}\scalebox{0.75}{$\pm$}\,0.021} & 0.842\,{\scriptsize\color{stdorange}\scalebox{0.75}{$\pm$}\,0.017} & 0.456\,{\scriptsize\color{stdorange}\scalebox{0.75}{$\pm$}\,0.027} & 0.786\,{\scriptsize\color{stdorange}\scalebox{0.75}{$\pm$}\,0.025} \\
Ex.5 & 0.274\,{\scriptsize\color{stdorange}\scalebox{0.75}{$\pm$}\,0.006} & 0.943\,{\scriptsize\color{stdorange}\scalebox{0.75}{$\pm$}\,0.005} & 0.399\,{\scriptsize\color{stdorange}\scalebox{0.75}{$\pm$}\,0.023} & 0.836\,{\scriptsize\color{stdorange}\scalebox{0.75}{$\pm$}\,0.019} & 0.455\,{\scriptsize\color{stdorange}\scalebox{0.75}{$\pm$}\,0.026} & 0.785\,{\scriptsize\color{stdorange}\scalebox{0.75}{$\pm$}\,0.023} \\
Ex.6 & 0.320\,{\scriptsize\color{stdorange}\scalebox{0.75}{$\pm$}\,0.012} & 0.925\,{\scriptsize\color{stdorange}\scalebox{0.75}{$\pm$}\,0.008} & 0.466\,{\scriptsize\color{stdorange}\scalebox{0.75}{$\pm$}\,0.036} & 0.780\,{\scriptsize\color{stdorange}\scalebox{0.75}{$\pm$}\,0.031} & 0.533\,{\scriptsize\color{stdorange}\scalebox{0.75}{$\pm$}\,0.039} & 0.692\,{\scriptsize\color{stdorange}\scalebox{0.75}{$\pm$}\,0.037} \\
Ex.7 & 0.247\,{\scriptsize\color{stdorange}\scalebox{0.75}{$\pm$}\,0.004} & 0.951\,{\scriptsize\color{stdorange}\scalebox{0.75}{$\pm$}\,0.002} & 0.358\,{\scriptsize\color{stdorange}\scalebox{0.75}{$\pm$}\,0.016} & 0.869\,{\scriptsize\color{stdorange}\scalebox{0.75}{$\pm$}\,0.011} & 0.414\,{\scriptsize\color{stdorange}\scalebox{0.75}{$\pm$}\,0.022} & 0.822\,{\scriptsize\color{stdorange}\scalebox{0.75}{$\pm$}\,0.018} \\
Ex.8 & 0.293\,{\scriptsize\color{stdorange}\scalebox{0.75}{$\pm$}\,0.010} & 0.936\,{\scriptsize\color{stdorange}\scalebox{0.75}{$\pm$}\,0.006} & 0.366\,{\scriptsize\color{stdorange}\scalebox{0.75}{$\pm$}\,0.020} & 0.861\,{\scriptsize\color{stdorange}\scalebox{0.75}{$\pm$}\,0.016} & 0.424\,{\scriptsize\color{stdorange}\scalebox{0.75}{$\pm$}\,0.024} & 0.812\,{\scriptsize\color{stdorange}\scalebox{0.75}{$\pm$}\,0.021} \\
Ex.9 & 0.249\,{\scriptsize\color{stdorange}\scalebox{0.75}{$\pm$}\,0.006} & 0.950\,{\scriptsize\color{stdorange}\scalebox{0.75}{$\pm$}\,0.004} & 0.440\,{\scriptsize\color{stdorange}\scalebox{0.75}{$\pm$}\,0.032} & 0.799\,{\scriptsize\color{stdorange}\scalebox{0.75}{$\pm$}\,0.028} & 0.503\,{\scriptsize\color{stdorange}\scalebox{0.75}{$\pm$}\,0.035} & 0.731\,{\scriptsize\color{stdorange}\scalebox{0.75}{$\pm$}\,0.032} \\
Ex.10 & 0.247\,{\scriptsize\color{stdorange}\scalebox{0.75}{$\pm$}\,0.004} & 0.951\,{\scriptsize\color{stdorange}\scalebox{0.75}{$\pm$}\,0.002} & 0.435\,{\scriptsize\color{stdorange}\scalebox{0.75}{$\pm$}\,0.025} & 0.804\,{\scriptsize\color{stdorange}\scalebox{0.75}{$\pm$}\,0.022} & 0.492\,{\scriptsize\color{stdorange}\scalebox{0.75}{$\pm$}\,0.030} & 0.746\,{\scriptsize\color{stdorange}\scalebox{0.75}{$\pm$}\,0.028} \\
Ex.11 & 0.247\,{\scriptsize\color{stdorange}\scalebox{0.75}{$\pm$}\,0.005} & 0.951\,{\scriptsize\color{stdorange}\scalebox{0.75}{$\pm$}\,0.003} & 0.360\,{\scriptsize\color{stdorange}\scalebox{0.75}{$\pm$}\,0.018} & 0.867\,{\scriptsize\color{stdorange}\scalebox{0.75}{$\pm$}\,0.012} & 0.423\,{\scriptsize\color{stdorange}\scalebox{0.75}{$\pm$}\,0.023} & 0.813\,{\scriptsize\color{stdorange}\scalebox{0.75}{$\pm$}\,0.020} \\
Ex.12 & 0.248\,{\scriptsize\color{stdorange}\scalebox{0.75}{$\pm$}\,0.005} & 0.951\,{\scriptsize\color{stdorange}\scalebox{0.75}{$\pm$}\,0.003} & 0.382\,{\scriptsize\color{stdorange}\scalebox{0.75}{$\pm$}\,0.024} & 0.849\,{\scriptsize\color{stdorange}\scalebox{0.75}{$\pm$}\,0.019} & 0.447\,{\scriptsize\color{stdorange}\scalebox{0.75}{$\pm$}\,0.029} & 0.793\,{\scriptsize\color{stdorange}\scalebox{0.75}{$\pm$}\,0.026} \\
Ex.13 & 0.247\,{\scriptsize\color{stdorange}\scalebox{0.75}{$\pm$}\,0.004} & 0.951\,{\scriptsize\color{stdorange}\scalebox{0.75}{$\pm$}\,0.002} & 0.371\,{\scriptsize\color{stdorange}\scalebox{0.75}{$\pm$}\,0.020} & 0.857\,{\scriptsize\color{stdorange}\scalebox{0.75}{$\pm$}\,0.016} & 0.432\,{\scriptsize\color{stdorange}\scalebox{0.75}{$\pm$}\,0.027} & 0.806\,{\scriptsize\color{stdorange}\scalebox{0.75}{$\pm$}\,0.024} \\
Ex.14 & 0.247\,{\scriptsize\color{stdorange}\scalebox{0.75}{$\pm$}\,0.005} & 0.951\,{\scriptsize\color{stdorange}\scalebox{0.75}{$\pm$}\,0.003} & 0.380\,{\scriptsize\color{stdorange}\scalebox{0.75}{$\pm$}\,0.024} & 0.849\,{\scriptsize\color{stdorange}\scalebox{0.75}{$\pm$}\,0.020} & 0.424\,{\scriptsize\color{stdorange}\scalebox{0.75}{$\pm$}\,0.026} & 0.812\,{\scriptsize\color{stdorange}\scalebox{0.75}{$\pm$}\,0.023} \\
Ex.15 & \textbf{0.241}\,{\scriptsize\color{stdorange}\scalebox{0.75}{$\pm$}\,0.005} & \textbf{0.954}\,{\scriptsize\color{stdorange}\scalebox{0.75}{$\pm$}\,0.003} & \textbf{0.353}\,{\scriptsize\color{stdorange}\scalebox{0.75}{$\pm$}\,0.018} & \textbf{0.872}\,{\scriptsize\color{stdorange}\scalebox{0.75}{$\pm$}\,0.012} & \textbf{0.405}\,{\scriptsize\color{stdorange}\scalebox{0.75}{$\pm$}\,0.022} & \textbf{0.829}\,{\scriptsize\color{stdorange}\scalebox{0.75}{$\pm$}\,0.019} \\
\midrule
\rowcolor{tableblue}
\textbf{Ex.16} & 0.247\,{\scriptsize\color{stdorange}\scalebox{0.75}{$\pm$}\,0.005} & 0.951\,{\scriptsize\color{stdorange}\scalebox{0.75}{$\pm$}\,0.003} & 0.358\,{\scriptsize\color{stdorange}\scalebox{0.75}{$\pm$}\,0.016} & 0.868\,{\scriptsize\color{stdorange}\scalebox{0.75}{$\pm$}\,0.011} & 0.414\,{\scriptsize\color{stdorange}\scalebox{0.75}{$\pm$}\,0.022} & 0.822\,{\scriptsize\color{stdorange}\scalebox{0.75}{$\pm$}\,0.019} \\
\bottomrule
\end{tabular}
\end{table}

\begin{table}[!htbp]
\centering
\caption{\textbf{Controller- and Task-Level Ablation \textit{Success Rates} (\%).}}
\label{tab:ablation-control-tasks}
\footnotesize
\setlength{\tabcolsep}{2pt}
\renewcommand{\arraystretch}{1.12}
\setlength{\aboverulesep}{0pt}
\setlength{\belowrulesep}{0pt}
\begin{tabular}{>{\centering\arraybackslash}p{\dimexpr0.08000000\linewidth-2\tabcolsep\relax}>{\centering\arraybackslash}p{\dimexpr0.11500000\linewidth-2\tabcolsep\relax}>{\centering\arraybackslash}p{\dimexpr0.11500000\linewidth-2\tabcolsep\relax}>{\centering\arraybackslash}p{\dimexpr0.11500000\linewidth-2\tabcolsep\relax}>{\centering\arraybackslash}p{\dimexpr0.11500000\linewidth-2\tabcolsep\relax}>{\centering\arraybackslash}p{\dimexpr0.11500000\linewidth-2\tabcolsep\relax}>{\centering\arraybackslash}p{\dimexpr0.11500000\linewidth-2\tabcolsep\relax}>{\centering\arraybackslash}p{\dimexpr0.11500000\linewidth-2\tabcolsep\relax}>{\centering\arraybackslash}p{\dimexpr0.11500000\linewidth-2\tabcolsep\relax}}
\toprule
\rowcolor{tableblue}
 & \multicolumn{2}{c}{Meta-World} & \multicolumn{3}{c}{Drawer} & \multicolumn{3}{c}{Cup} \\
\noalign{{\color{tableblue}\hrule height \cmidrulewidth}\kern-\cmidrulewidth}
\cmidrule(lr){2-3}\cmidrule(lr){4-6}\cmidrule(lr){7-9}
\rowcolor{tableblue}
\multirow{-2}{*}{ID} & PL$\uparrow$ & MPC$\uparrow$ & Pull$\uparrow$ & Close$\uparrow$ & Full$\uparrow$ & Pick$\uparrow$ & Place$\uparrow$ & Full$\uparrow$ \\
\midrule
Ex.1 & 70.9\,{\scriptsize\color{stdorange}\scalebox{0.75}{$\pm$}\,3.2} & 83.8\,{\scriptsize\color{stdorange}\scalebox{0.75}{$\pm$}\,2.4} & 82.2\,{\scriptsize\color{stdorange}\scalebox{0.75}{$\pm$}\,3.8} & 85.6\,{\scriptsize\color{stdorange}\scalebox{0.75}{$\pm$}\,1.9} & 65.6\,{\scriptsize\color{stdorange}\scalebox{0.75}{$\pm$}\,5.1} & 80.0\,{\scriptsize\color{stdorange}\scalebox{0.75}{$\pm$}\,3.3} & 81.1\,{\scriptsize\color{stdorange}\scalebox{0.75}{$\pm$}\,1.9} & 61.1\,{\scriptsize\color{stdorange}\scalebox{0.75}{$\pm$}\,5.1} \\
Ex.2 & 73.6\,{\scriptsize\color{stdorange}\scalebox{0.75}{$\pm$}\,2.9} & 87.1\,{\scriptsize\color{stdorange}\scalebox{0.75}{$\pm$}\,2.5} & 90.0\,{\scriptsize\color{stdorange}\scalebox{0.75}{$\pm$}\,3.3} & 91.1\,{\scriptsize\color{stdorange}\scalebox{0.75}{$\pm$}\,1.9} & 74.4\,{\scriptsize\color{stdorange}\scalebox{0.75}{$\pm$}\,5.1} & 86.7\,{\scriptsize\color{stdorange}\scalebox{0.75}{$\pm$}\,3.3} & 90.0\,{\scriptsize\color{stdorange}\scalebox{0.75}{$\pm$}\,3.3} & 71.1\,{\scriptsize\color{stdorange}\scalebox{0.75}{$\pm$}\,5.1} \\
Ex.3 & 74.9\,{\scriptsize\color{stdorange}\scalebox{0.75}{$\pm$}\,2.6} & 88.1\,{\scriptsize\color{stdorange}\scalebox{0.75}{$\pm$}\,2.7} & 91.1\,{\scriptsize\color{stdorange}\scalebox{0.75}{$\pm$}\,3.8} & 93.3\,{\scriptsize\color{stdorange}\scalebox{0.75}{$\pm$}\,3.3} & 77.8\,{\scriptsize\color{stdorange}\scalebox{0.75}{$\pm$}\,5.1} & 88.9\,{\scriptsize\color{stdorange}\scalebox{0.75}{$\pm$}\,3.8} & 91.1\,{\scriptsize\color{stdorange}\scalebox{0.75}{$\pm$}\,1.9} & 74.4\,{\scriptsize\color{stdorange}\scalebox{0.75}{$\pm$}\,5.1} \\
Ex.4 & 76.9\,{\scriptsize\color{stdorange}\scalebox{0.75}{$\pm$}\,2.4} & 89.7\,{\scriptsize\color{stdorange}\scalebox{0.75}{$\pm$}\,2.7} & \textbf{95.6}\,{\scriptsize\color{stdorange}\scalebox{0.75}{$\pm$}\,1.9} & 94.4\,{\scriptsize\color{stdorange}\scalebox{0.75}{$\pm$}\,1.9} & 80.0\,{\scriptsize\color{stdorange}\scalebox{0.75}{$\pm$}\,6.7} & 93.3\,{\scriptsize\color{stdorange}\scalebox{0.75}{$\pm$}\,3.3} & \textbf{93.3}\,{\scriptsize\color{stdorange}\scalebox{0.75}{$\pm$}\,0.0} & 77.8\,{\scriptsize\color{stdorange}\scalebox{0.75}{$\pm$}\,5.1} \\
Ex.5 & 75.3\,{\scriptsize\color{stdorange}\scalebox{0.75}{$\pm$}\,2.9} & 88.4\,{\scriptsize\color{stdorange}\scalebox{0.75}{$\pm$}\,2.7} & 91.1\,{\scriptsize\color{stdorange}\scalebox{0.75}{$\pm$}\,1.9} & 92.2\,{\scriptsize\color{stdorange}\scalebox{0.75}{$\pm$}\,1.9} & 75.6\,{\scriptsize\color{stdorange}\scalebox{0.75}{$\pm$}\,5.1} & 90.0\,{\scriptsize\color{stdorange}\scalebox{0.75}{$\pm$}\,3.3} & 91.1\,{\scriptsize\color{stdorange}\scalebox{0.75}{$\pm$}\,1.9} & 73.3\,{\scriptsize\color{stdorange}\scalebox{0.75}{$\pm$}\,6.7} \\
Ex.6 & 71.2\,{\scriptsize\color{stdorange}\scalebox{0.75}{$\pm$}\,1.8} & 83.9\,{\scriptsize\color{stdorange}\scalebox{0.75}{$\pm$}\,1.8} & 83.3\,{\scriptsize\color{stdorange}\scalebox{0.75}{$\pm$}\,3.3} & 86.7\,{\scriptsize\color{stdorange}\scalebox{0.75}{$\pm$}\,3.3} & 66.7\,{\scriptsize\color{stdorange}\scalebox{0.75}{$\pm$}\,3.3} & 80.0\,{\scriptsize\color{stdorange}\scalebox{0.75}{$\pm$}\,3.3} & 84.4\,{\scriptsize\color{stdorange}\scalebox{0.75}{$\pm$}\,1.9} & 61.1\,{\scriptsize\color{stdorange}\scalebox{0.75}{$\pm$}\,5.1} \\
Ex.7 & \textbf{77.1}\,{\scriptsize\color{stdorange}\scalebox{0.75}{$\pm$}\,2.6} & \textbf{90.1}\,{\scriptsize\color{stdorange}\scalebox{0.75}{$\pm$}\,2.8} & \textbf{95.6}\,{\scriptsize\color{stdorange}\scalebox{0.75}{$\pm$}\,1.9} & \textbf{95.6}\,{\scriptsize\color{stdorange}\scalebox{0.75}{$\pm$}\,1.9} & \textbf{81.1}\,{\scriptsize\color{stdorange}\scalebox{0.75}{$\pm$}\,1.9} & \textbf{94.4}\,{\scriptsize\color{stdorange}\scalebox{0.75}{$\pm$}\,1.9} & \textbf{93.3}\,{\scriptsize\color{stdorange}\scalebox{0.75}{$\pm$}\,3.3} & 78.9\,{\scriptsize\color{stdorange}\scalebox{0.75}{$\pm$}\,6.9} \\
Ex.8 & 76.4\,{\scriptsize\color{stdorange}\scalebox{0.75}{$\pm$}\,2.6} & 89.7\,{\scriptsize\color{stdorange}\scalebox{0.75}{$\pm$}\,3.0} & 93.3\,{\scriptsize\color{stdorange}\scalebox{0.75}{$\pm$}\,3.3} & 94.4\,{\scriptsize\color{stdorange}\scalebox{0.75}{$\pm$}\,1.9} & 78.9\,{\scriptsize\color{stdorange}\scalebox{0.75}{$\pm$}\,6.9} & 93.3\,{\scriptsize\color{stdorange}\scalebox{0.75}{$\pm$}\,3.3} & \textbf{93.3}\,{\scriptsize\color{stdorange}\scalebox{0.75}{$\pm$}\,3.3} & 76.7\,{\scriptsize\color{stdorange}\scalebox{0.75}{$\pm$}\,3.3} \\
Ex.9 & 72.9\,{\scriptsize\color{stdorange}\scalebox{0.75}{$\pm$}\,2.6} & 86.8\,{\scriptsize\color{stdorange}\scalebox{0.75}{$\pm$}\,2.8} & 90.0\,{\scriptsize\color{stdorange}\scalebox{0.75}{$\pm$}\,3.3} & 91.1\,{\scriptsize\color{stdorange}\scalebox{0.75}{$\pm$}\,1.9} & 71.1\,{\scriptsize\color{stdorange}\scalebox{0.75}{$\pm$}\,5.1} & 86.7\,{\scriptsize\color{stdorange}\scalebox{0.75}{$\pm$}\,3.3} & 90.0\,{\scriptsize\color{stdorange}\scalebox{0.75}{$\pm$}\,3.3} & 67.8\,{\scriptsize\color{stdorange}\scalebox{0.75}{$\pm$}\,5.1} \\
Ex.10 & 75.9\,{\scriptsize\color{stdorange}\scalebox{0.75}{$\pm$}\,2.6} & 88.9\,{\scriptsize\color{stdorange}\scalebox{0.75}{$\pm$}\,2.8} & 93.3\,{\scriptsize\color{stdorange}\scalebox{0.75}{$\pm$}\,3.3} & 94.4\,{\scriptsize\color{stdorange}\scalebox{0.75}{$\pm$}\,1.9} & 77.8\,{\scriptsize\color{stdorange}\scalebox{0.75}{$\pm$}\,5.1} & 92.2\,{\scriptsize\color{stdorange}\scalebox{0.75}{$\pm$}\,1.9} & 92.2\,{\scriptsize\color{stdorange}\scalebox{0.75}{$\pm$}\,3.8} & 74.4\,{\scriptsize\color{stdorange}\scalebox{0.75}{$\pm$}\,5.1} \\
Ex.11 & 76.9\,{\scriptsize\color{stdorange}\scalebox{0.75}{$\pm$}\,2.9} & 89.8\,{\scriptsize\color{stdorange}\scalebox{0.75}{$\pm$}\,2.7} & \textbf{95.6}\,{\scriptsize\color{stdorange}\scalebox{0.75}{$\pm$}\,1.9} & \textbf{95.6}\,{\scriptsize\color{stdorange}\scalebox{0.75}{$\pm$}\,1.9} & \textbf{81.1}\,{\scriptsize\color{stdorange}\scalebox{0.75}{$\pm$}\,5.1} & 93.3\,{\scriptsize\color{stdorange}\scalebox{0.75}{$\pm$}\,3.3} & 92.2\,{\scriptsize\color{stdorange}\scalebox{0.75}{$\pm$}\,1.9} & 76.7\,{\scriptsize\color{stdorange}\scalebox{0.75}{$\pm$}\,3.3} \\
Ex.12 & 76.3\,{\scriptsize\color{stdorange}\scalebox{0.75}{$\pm$}\,2.7} & 89.2\,{\scriptsize\color{stdorange}\scalebox{0.75}{$\pm$}\,2.8} & 93.3\,{\scriptsize\color{stdorange}\scalebox{0.75}{$\pm$}\,3.3} & 93.3\,{\scriptsize\color{stdorange}\scalebox{0.75}{$\pm$}\,0.0} & 77.8\,{\scriptsize\color{stdorange}\scalebox{0.75}{$\pm$}\,5.1} & 92.2\,{\scriptsize\color{stdorange}\scalebox{0.75}{$\pm$}\,1.9} & \textbf{93.3}\,{\scriptsize\color{stdorange}\scalebox{0.75}{$\pm$}\,3.3} & 74.4\,{\scriptsize\color{stdorange}\scalebox{0.75}{$\pm$}\,5.1} \\
Ex.13 & 76.8\,{\scriptsize\color{stdorange}\scalebox{0.75}{$\pm$}\,2.7} & 89.7\,{\scriptsize\color{stdorange}\scalebox{0.75}{$\pm$}\,2.9} & \textbf{95.6}\,{\scriptsize\color{stdorange}\scalebox{0.75}{$\pm$}\,1.9} & \textbf{95.6}\,{\scriptsize\color{stdorange}\scalebox{0.75}{$\pm$}\,1.9} & 80.0\,{\scriptsize\color{stdorange}\scalebox{0.75}{$\pm$}\,3.3} & \textbf{94.4}\,{\scriptsize\color{stdorange}\scalebox{0.75}{$\pm$}\,1.9} & \textbf{93.3}\,{\scriptsize\color{stdorange}\scalebox{0.75}{$\pm$}\,3.3} & 77.8\,{\scriptsize\color{stdorange}\scalebox{0.75}{$\pm$}\,5.1} \\
Ex.14 & 76.9\,{\scriptsize\color{stdorange}\scalebox{0.75}{$\pm$}\,3.1} & 89.7\,{\scriptsize\color{stdorange}\scalebox{0.75}{$\pm$}\,2.6} & \textbf{95.6}\,{\scriptsize\color{stdorange}\scalebox{0.75}{$\pm$}\,1.9} & 94.4\,{\scriptsize\color{stdorange}\scalebox{0.75}{$\pm$}\,3.8} & 78.9\,{\scriptsize\color{stdorange}\scalebox{0.75}{$\pm$}\,6.9} & \textbf{94.4}\,{\scriptsize\color{stdorange}\scalebox{0.75}{$\pm$}\,1.9} & \textbf{93.3}\,{\scriptsize\color{stdorange}\scalebox{0.75}{$\pm$}\,3.3} & 77.8\,{\scriptsize\color{stdorange}\scalebox{0.75}{$\pm$}\,5.1} \\
Ex.15 & \textbf{77.1}\,{\scriptsize\color{stdorange}\scalebox{0.75}{$\pm$}\,2.4} & 90.0\,{\scriptsize\color{stdorange}\scalebox{0.75}{$\pm$}\,2.8} & \textbf{95.6}\,{\scriptsize\color{stdorange}\scalebox{0.75}{$\pm$}\,1.9} & \textbf{95.6}\,{\scriptsize\color{stdorange}\scalebox{0.75}{$\pm$}\,1.9} & \textbf{81.1}\,{\scriptsize\color{stdorange}\scalebox{0.75}{$\pm$}\,1.9} & \textbf{94.4}\,{\scriptsize\color{stdorange}\scalebox{0.75}{$\pm$}\,1.9} & \textbf{93.3}\,{\scriptsize\color{stdorange}\scalebox{0.75}{$\pm$}\,3.3} & \textbf{80.0}\,{\scriptsize\color{stdorange}\scalebox{0.75}{$\pm$}\,3.3} \\
\midrule
\rowcolor{tableblue}
\textbf{Ex.16} & \textbf{77.1}\,{\scriptsize\color{stdorange}\scalebox{0.75}{$\pm$}\,2.8} & \textbf{90.1}\,{\scriptsize\color{stdorange}\scalebox{0.75}{$\pm$}\,2.8} & \textbf{95.6}\,{\scriptsize\color{stdorange}\scalebox{0.75}{$\pm$}\,1.9} & \textbf{95.6}\,{\scriptsize\color{stdorange}\scalebox{0.75}{$\pm$}\,1.9} & \textbf{81.1}\,{\scriptsize\color{stdorange}\scalebox{0.75}{$\pm$}\,5.1} & \textbf{94.4}\,{\scriptsize\color{stdorange}\scalebox{0.75}{$\pm$}\,1.9} & \textbf{93.3}\,{\scriptsize\color{stdorange}\scalebox{0.75}{$\pm$}\,3.3} & 78.9\,{\scriptsize\color{stdorange}\scalebox{0.75}{$\pm$}\,3.8} \\
\bottomrule
\end{tabular}
\end{table}

\clearpage

\section{Limitations}
\label{app:limitations}

\textbf{Camera and task scope.}
The robot protocol uses a calibrated fixed external camera and
embodiment-specific training. Real-world evaluation covers drawer
opening/closing and cup pick-and-place in matched indoor workspaces.
Moving-camera prediction, zero-shot transfer across embodiments, and
broader object and workspace changes remain outside the demonstrated scope.

\textbf{Representation and prediction scope.}
NYUv2 and VOC12 probes measure the adapted RGB encoder, while RGB-D
completion evaluates the shared state with its decoding paths. The
shared/private projections encourage a division of predictive roles,
not an identifiable semantic decomposition. Temporal comparisons with
recorded proprioception measure conditioned visual forecasting;
planning and ablation rollouts use predicted proprioception. Whole-image
feature accuracy does not by itself establish action sensitivity in
moving regions or accurate physical trajectories.

\textbf{Memory and component interpretation.}
A single initial frame may omit occluded state or motion history.
Fixed initial private memory is a compact decoding choice whose adequacy
under large displacement and newly visible surfaces requires further study.
The reported pixel-query and untied-decoder variants do not support an
accuracy advantage for ray queries or decoder sharing. FK is a soft
kinematic regularizer; neither it nor a valid SE(3) parameterization
guarantees correct contact dynamics or collision-free control.

\end{document}